\documentclass[runningheads]{llncs}

\usepackage{eccv}

\usepackage{eccvabbrv}

\usepackage{graphicx}
\usepackage{booktabs}
\usepackage{subcaption}
\usepackage{multirow}
\usepackage{amsmath}
\usepackage{amssymb}
\usepackage{algorithm}
\usepackage{wrapfig}
\usepackage{subcaption}
\usepackage{tikz}
\usetikzlibrary{calc}
\usepackage[noend]{algpseudocode}

\usepackage{comment}

\usepackage[accsupp]{axessibility}  

\usepackage{hyperref}

\usepackage{orcidlink}

\begin{document}

\title{Evaluating the Interpretability of Sparse Autoencoders with Concept Annotations}


\titlerunning{Evaluating the Interpretability of SAEs with Concept Annotations} 

\author{Jonas Klotz\inst{1,2}\orcidlink{0009-0003-3297-2365} \and
Cassio F. Dantas\inst{3,5}\orcidlink{0000-0002-1934-0625} \and
Pallavi Jain\inst{4,5}\orcidlink{0000-0002-1731-8993}\and
Diego Marcos\inst{4,5}\orcidlink{0000-0001-5607-4445}\and
Beg\"{u}m Demir \inst{1,2}\orcidlink{0000-0003-2175-7072}}

\authorrunning{J.~Klotz et al.}


\institute{
The Berlin Institute for the Foundations of Learning and Data (BIFOLD) \and
Technische Universität Berlin, Germany \email{\{j.klotz,demir\}@tu-berlin.de}\and
INRAE \email{cassio.fraga-dantas@inrae.fr} \and
Inria, EVERGREEN \email{\{pallavi.jain, diego.marcos\}@inria.fr}
\and UMR TETIS, Univ Montpellier, France}

\maketitle

\begin{abstract}
Sparse autoencoders (SAEs) are increasingly used to extract interpretable concepts from vision and vision language models, yet existing evaluation methods largely rely on proxy metrics or qualitative inspection rather than measuring semantic correspondence. We present a human-grounded evaluation framework that quantifies alignment between SAE latents and human-annotated concepts, without requiring user studies, and validate this matching through targeted attribute perturbations. To enable this intervention-style evaluation in vision, we construct synCUB and synCOCO, synthetic benchmarks of paired images that differ in exactly one attribute. We introduce Fully-Binary Matching Pursuit (FBMP), a coalition-based matching procedure that supports many-to-one mappings between SAE latents and annotated concepts, and consistently outperforms one-to-one baselines. For functional validation, we propose a Targeted Attribute Perturbation Alignment Score (TAPAScore), which tests whether matched concepts respond selectively and in the expected direction under targeted image-level attribute perturbations.
Under sanity checks, our matching and TAPAScore are the only evaluated metrics that reliably distinguish trained SAEs from untrained ones. Across SAEs trained on CLIP and DINOv2 embeddings, we find that increased overcompleteness can reduce perturbation alignment, indicating a reduction in interpretability. Our evaluation framework suggests that moderate dictionary sizes provide the best trade-off, yielding the most interpretable SAEs. Code and datasets are available at \url{https://github.com/JonasKlotz/sae-concept-eval}.
\end{abstract}

\section{Introduction}
\begin{figure}
    \centering
            \vspace{-2.5em}

    \includegraphics[width=1\linewidth, trim=0cm .8cm 0cm 0.4cm]{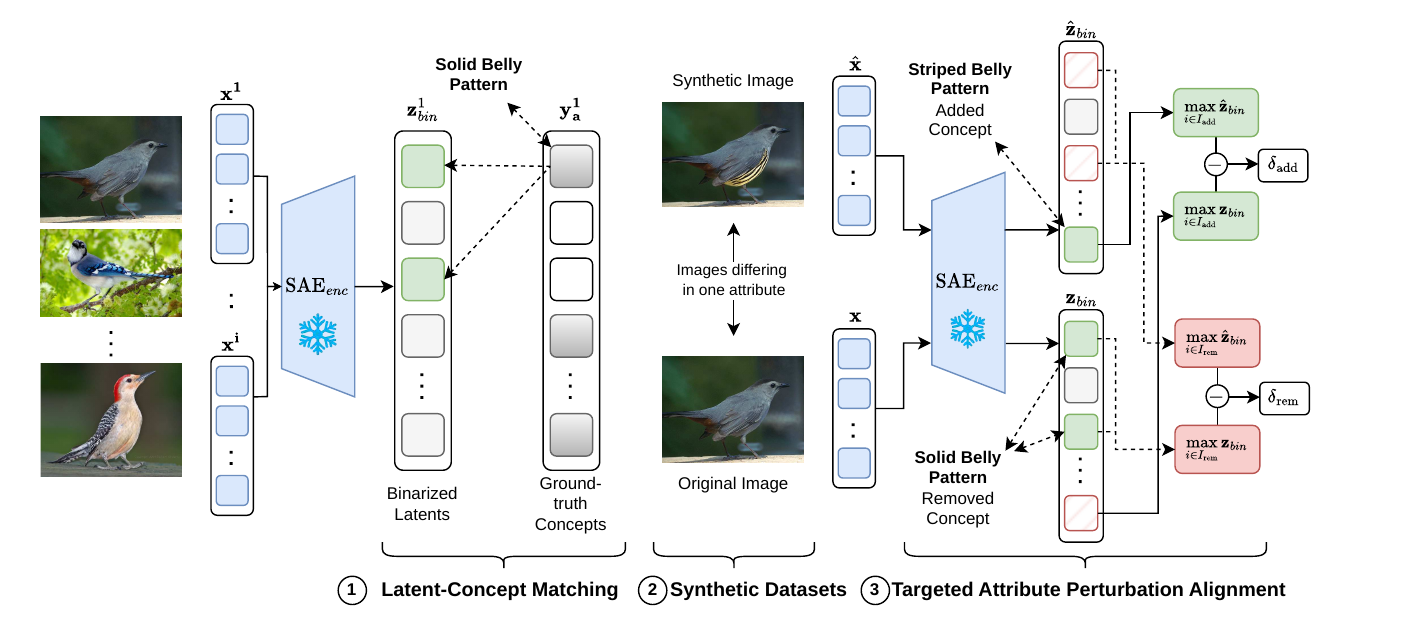}
    \vspace{-1.2em}  
    \caption{\textbf{Human-grounded evaluation of SAE interpretability.} The framework comprises three components: \textbf{(1)} latent-concept matching, where binarized SAE latents are aligned to ground truth attribute vectors; \textbf{(2)} synthetic datasets, where paired images are constructed to differ in exactly one concept (e.g., belly pattern changed from solid to striped); and \textbf{(3)} targeted attribute perturbation alignment, where the induced latent change is measured via the signed responses $\delta_{\text{add}}$ and $\delta_{\text{rem}}$. TAPAScore combines these components to quantify the perturbation alignment.}
    \label{fig:full_method}
        \vspace{-2.5em}

\end{figure}

Large vision and vision-language models such as DINOv2~\cite{oquab2024dinov} and CLIP~\cite{radford2021learning} rely on high-dimensional internal representations whose structure is difficult to interpret. Understanding what individual latent dimensions represent is therefore critical for trust, controllability, and scientific analysis of these models~\cite{saeed2023explainable}. One promising approach is the use of sparse autoencoders (SAEs) to inspect their internal representations, a technique first developed for large language models~\cite{bricken2023towards, huben2024sparse}. By enforcing sparsity, SAEs decompose high-dimensional activations into more localized and disentangled components that can be aligned with human-interpretable concepts. Following their success in language-model interpretability, SAEs are being applied to vision encoders to identify sparse latents associated with object parts, textures, and attributes~\cite{fel2023holistic, bhalla2024interpreting, lim2024sparse, rao2024discover}.

However, SAEs are known to exhibit systematic failure modes, such as feature splitting, where a single concept is fragmented across multiple latents~\cite{bricken2023towards}; feature absorption, where a general feature develops blind spots covered by specialized latents~\cite{chanin2024absorption}; and feature composition, where co-occurring concepts merge into a single latent~\cite{wattenberg2024relational}. SAE features are furthermore often organized hierarchically, with both fine-grained and higher-level abstractions distributed across latents~\cite{bussmann2025learning}. Unless the SAE captures inductive biases that reflect the semantic structure present in the data, there is no guarantee that learned features will align with the concepts that human observers identify as meaningful~\cite{fel2025archetypal}. Consequently, we claim that the alignment between SAE features and human-understandable concepts cannot be assumed from architectural design or reconstruction quality alone, but must be evaluated explicitly.
Such explicit evaluation requires metrics that capture semantic correspondence. Existing SAE metrics, developed primarily in the language domain, fall into structural and functional families~\cite{shu2025survey}. Whereas structural metrics, which assess representational properties such as sparsity or reconstruction fidelity, transfer naturally from language to vision, functional evaluation requires controlled interventions that isolate a single semantic change. This condition is particularly difficult to satisfy in vision, where attribute changes in images rarely occur in isolation. As a result, SAE evaluation for vision models remains dominated by qualitative examples and structural proxies, which measure representational organization rather than whether features correspond to semantic concepts.

To address this limitation, we operationalize interpretability as the degree to which learned SAE features align with human-understandable concepts present in the data, as in prior work on representation interpretability~\cite{bau2017network, kim2018interpretability}. Human evaluation is considered the gold standard of interpretability assessment~\cite{doshi2017towards}, and human-annotated concepts/attributes provide a practical measure of this standard, as they precisely reflect the factors that human observers consistently identify as meaningful. 
Figure~\ref{fig:full_method} summarizes our proposed evaluation framework. We first quantify semantic correspondence by matching binarized SAE activations to binary ground truth attribute vectors. We then construct controlled image pairs that differ in exactly one attribute, enabling intervention-style tests where the target change is precisely known. Finally, we introduce TAPAScore, which measures whether the latents previously matched to the perturbed attribute respond in the expected direction.
In summary, \textbf{our contributions are}:
\begin{itemize}
    \item We propose matching metrics between SAE latents and human-annotated attributes, including a coalition-based formulation, fully-binary matching pursuit (FBMP), that supports many-to-one mappings. We find that existing metrics fail basic sanity checks that our proposed metrics pass (Fig.~\ref{fig:matching_failure_mode}) and that FBMP consistently outperforms one-to-one matchings (Figs.~\ref{fig:matching_and_tapas_cub}, ~\ref{fig:matching_and_tapas_coco}, \textit{top}).

    \item  We construct two synthetic datasets, synCUB and synCOCO, consisting of paired images that differ in exactly one attribute or object label, enabling intervention-style evaluation of SAEs in vision.

    \item We propose a targeted attribute perturbation alignment score (TAPAScore) that tests whether matched latent sets respond selectively and in the right direction under targeted attribute perturbations. We show that increased overcompleteness can degrade perturbation alignment (Figs.~\ref{fig:matching_and_tapas_cub}, ~\ref{fig:matching_and_tapas_coco}, \textit{bottom}), and that matching score and TAPAScore are positively correlated (Fig.~\ref{fig:delta_matching_score_vs_tapas}).
\end{itemize}
Beyond the framework itself, our evaluation yields consistent empirical guidance for practitioners: increasing overcompleteness improves statistical matching but tends to degrade perturbation alignment; moderate dictionary sizes offer the best trade-off. 
Based on these findings, we recommend FBMP $F_{0.5}$ matching paired with TAPAScore as the default evaluation protocol.

\section{Background and Related Work}
\noindent\textbf{Sparse Autoencoders} (SAEs) \cite{bricken2023towards, huben2024sparse} are a method to solve sparse dictionary learning, where signals are represented as sparse linear combinations of basis elements (atoms) from an overcomplete dictionary \cite{rubinstein2010dictionaries, tovsic2011dictionary}. More recently, SAEs have been used to identify interpretable concepts in neural network activations. This is motivated by evidence that individual neurons are frequently polysemantic and encode multiple unrelated concepts, suggesting that deep networks store information in superposed representations \cite{ elhage2022toy}. SAEs disentangle these representations by learning a sparse and overcomplete representation in which the number of learned features exceeds the dimensionality of the activation space.
SAEs implement dictionary learning through a neural network consisting of an encoder $\mathbf{W}_{\text{enc}} \in \mathbb{R}^{D \times L}$, decoder $\mathbf{W}_{\text{dec}} \in \mathbb{R}^{L \times D}$, and shared bias $\mathbf{b} \in \mathbb{R}^D$. Given an embedding $\mathbf{x} \in \mathbb{R}^D$ from a pretrained model, the SAE computes sparse activations $\mathbf{z} \in \mathbb{R}^L$ and the corresponding reconstruction $\hat{\mathbf{x}}$ as:
\begin{equation}
\mathbf{z} = \sigma\!\left(\mathbf{W}_{\text{enc}}^{\top}(\mathbf{x} - \mathbf{b})\right),
\qquad
\hat{\mathbf{x}} = \mathbf{W}_{\text{dec}}^{\top}\mathbf{z} + \mathbf{b}.
\end{equation}
The model parameters are learned by minimizing an objective $\mathcal{L}(\mathbf{x}) = \mathcal{R}(\mathbf{x}) + \lambda\mathcal{S}(\mathbf{x})$ with a reconstruction and a sparsity-inducing term.
SAE architectures primarily differ in how sparsity is enforced. Vanilla SAEs \cite{bricken2023towards} use ReLU activation with $L^1$ penalty and $L^2$ reconstruction. TopK SAEs \cite{gao2024scaling} and BatchTopK SAEs \cite{bussmann2024batchtopk} replace soft sparsity penalties with hard constraints, retaining only the $K$ largest activations per sample or batch. JumpReLU SAEs \cite{rajamanoharan2024jumping} instead learn a per-latent activation threshold under a direct $L^0$ penalty. Matryoshka SAEs \cite{bussmann2025learning} learn nested dictionaries with grouped activations.

\noindent\textbf{Evaluation Metrics in Language-Model Interpretability.}
Evaluation metrics for SAEs are commonly categorized into \emph{structural} and \emph{functional} metrics \cite{shu2025survey}. Structural metrics assess whether the learned representation of the SAE preserves properties of the original embedding space, including reconstruction error, recovery of known features, and ablation-based diagnostics \cite{gao2024scaling, templeton2024scaling, makelov2024towards}. The centered kernel nearest neighbor alignment (CKNNA) metric \cite{zaigrajew2025interpreting} quantifies how well the neighborhood structure of the original embedding space is preserved after transformation. While essential for diagnosing training behavior, these measures do not directly evaluate whether learned features correspond to meaningful semantic concepts.
Functional metrics assess whether SAE representations are semantically interpretable and practically useful. Feature Monosemanticity Score (FMS)~\cite{harle2025measuring} trains a predictor to relate latents to semantic attributes, while automated interpretability approaches~\cite{paulo2025automatically} score features by prompting an LLM to generate and test natural language explanations. A complementary direction focuses on intervention-based evaluation, testing whether manipulating identified features produces the intended semantic effect~\cite{saphra2024mechanistic, mueller2024quest, bhalla2024towards}, a paradigm widely applied in mechanistic analysis \cite{hernandez2024linearity} and activation steering \cite{rimsky2024steering, ghandeharioun2024s}. However, applying this paradigm to vision is non-trivial, as it often requires counterfactual data that isolates the influence of individual visual attributes.

\noindent\textbf{SAEs for Vision Models.}
While SAEs were initially developed and widely applied in natural language processing \cite{chanin2024absorption, shu2025survey}, recent work has transferred this methodology to vision models to recover sparse visual features corresponding to object parts, textures, attributes, or other interpretable concepts~\cite{fel2023holistic, bhalla2024interpreting, lim2024sparse, fel2025into}, enabling applications such as concept discovery, representation probing, and interpretability analysis~\cite{rao2024discover, vielhaben2025beyond, zaigrajew2025interpreting, olson2025analyzing}. Empirical studies further indicate that the internal representations of the CLIP vision encoder organize along interpretable concept directions recoverable through sparse feature learning \cite{rao2024discover, vielhaben2025beyond, bhalla2024interpreting}.
Evaluating vision SAEs, however, presents unique challenges: while structural metrics transfer across domains, functional evaluation must be adapted to visual semantics. Pach et al.~\cite{pach2025sparse} introduce the MonoSemanticity (MS) score, measuring how consistently a latent is activated by semantically similar inputs, computed as the activation-weighted similarity between images that strongly activate it. Complementary steerability metrics quantify how strongly manipulating a feature alters model output distributions and concept coverage~\cite{joseph2025steering}.

An approach towards ground-truth evaluation uses synthetic benchmarks with known generative factors. Fel et al.~\cite{fel2025archetypal} propose a Soft Identifiability Benchmark where images are constructed by collaging distinct objects, evaluating SAE features by whether each object class has a corresponding latent that activates when present. However, the synthetic scenes lack visual richness, which may limit the evaluation of larger SAEs. The SUB~\cite{bader2025sub} dataset offers more realistic concept-level evaluation via CUB-derived bird images with controlled concept substitutions, but is designed for concept bottleneck models rather than SAEs, leaving a gap for ground-truth evaluation on complex data.
\section{Human-Grounded Evaluation of SAE Concepts}
Evaluating whether SAE features correspond to meaningful concepts requires a concrete operationalization of interpretability. Following prior work~\cite{bau2017network, kim2018interpretability}, we define interpretability as the degree to which SAE features align with human-understandable concepts, and use human-annotated concepts as a scalable proxy for human judgment. This forms the basis of two complementary evaluation components: latent-concept matching, which quantifies statistical alignment between SAE latents and annotated concepts, and targeted attribute perturbation alignment, which tests whether matched latents respond selectively and in the correct direction to images differing in exactly one concept.

\subsection{Latent-Concept Matching}\label{sec:matching}
To measure statistical alignment between SAE latents and human-annotated concepts, we compare binary ground-truth attribute annotations with binarized SAE activations.  
Let $\mathbf{Y} \in \{0,1\}^{A \times N}$ denote the attribute annotation matrix for $N$ samples and $A$ attributes, and let $\mathbf{y}_a \in \{0,1\}^N$ be the row corresponding to attribute $a$.  
For a latent unit $l$, let $\mathbf{z}_l \in \mathbb{R}^N$ denote its activation vector across all samples, where $(\mathbf{z}_l)_i$ is the activation produced by the SAE for sample $i$.  
We define the binarized activation vector $\mathbf{z}^l_{\text{bin}} \in \{0,1\}^N$ such that $(\mathbf{z}^l_{\text{bin}})_i \!=\! 1$ if $(\mathbf{z}_l)_i \!>\! 0$ and $0$ otherwise.  
Alignment between latent unit $l$ and attribute $a$ is then evaluated by comparing $\mathbf{z}^l_{\text{bin}}$ and $\mathbf{y}_a$ using standard binary matching metrics.

\noindent\textbf{One-to-one matching.}
The SAE latent that best aligns with attribute $a$ under the $F_1$ score is given by:
\begin{equation} \label{eq:F1_matching}
l_a^\ast = \arg\max_l F_1(\mathbf{y}_a, \mathbf{z}^l_{\text{bin}}).
\end{equation}
One-to-one matching assumes that each attribute is represented by exactly one SAE latent. However, feature splitting~\cite{bricken2023towards} causes a unified concept to fragment across multiple specialized latents. For instance, `striped belly' may be split into one latent responding to stripe pattern and one to belly region, such that neither alone achieves high $F_1$ even though together they fully encode the attribute. This motivates a many-to-one matching formulation, described next.

\noindent\textbf{Many-to-one matching (Fully-Binary Matching Pursuit).}
Rather than selecting the single best-aligned latent, we seek a small subset (coalition) of latents whose combined activations better reconstruct a given attribute annotation. A naïve approach would select the top-$k$ latents by individual binary similarity score (e.g., $F_1$), but this tends to produce redundant coalitions rather than complementary ones. We therefore adopt a sequential selection procedure that greedily builds a more informative coalition.
The proposed approach is based on well-established greedy techniques from sparse reconstruction literature, namely Matching Pursuit \cite{mallat1993matching} and its variants \cite{pati1993orthogonal,wen2021binary}. At each step, the latent that best complements the current coalition is selected, and its contribution is removed from the residual before the next selection. This ensures that each newly added latent is complementary to the previously selected ones rather than redundant. Although best subset selection is NP-hard~\cite{natarajan1995sparse}, greedy approaches of this kind have proven effective in practice and optimal under certain conditions~\cite{tropp2004greed, tropp2007signal}.

We propose Fully-Binary Matching Pursuit (FBMP), a variant of Matching Pursuit tailored for the reconstruction of binary signals. While binary matching pursuit has been explored in prior work~\cite{wen2021binary,li2024binary}, existing formulations target binary coefficient vectors while leaving the input vector and candidate atoms continuous. In our setting, both the attribute annotation vectors and the latent activations are binary, which precludes the use of standard inner products and vector sums. We therefore adapt the matching pursuit procedure to operate entirely in the binary domain. 
The resulting approach, described in Algorithm~\ref{alg:FBMP}, is, to the best of our knowledge, novel. In the proposed procedure: 
(1) a binary similarity metric (e.g., $F_\beta$ score%
\footnote{We propose using $F_\beta$ score with some $\beta\leq1$ to favor precision over recall. Because multiple latents are to be combined, high recall is not crucial for a single latent.}) is used instead of standard inner-product-based correlation for best-atom selection; (2) logical OR ($\lor$) operations are used to sum the contributions of each atom in the reconstruction, instead of simple sums; (3) logical AND ($\land$) and NOT ($\lnot$) operations are used to update residuals instead of standard subtraction operations. A stopping criterion terminates the algorithm as soon as a newly-selected latent no longer improves the $F_1$ score of the current approximation. Although the maximum coalition size $k$ may be set to a large value, the algorithm typically returns a smaller subset (see Appendix~\ref{sup:fbmp}).

While FBMP is many-to-one \emph{per attribute call}, as each attribute selects a subset of latents, the overall matching is many-to-many at the set level: a latent can be selected by multiple attributes simultaneously (see Fig.~\ref{fig:matching_abstract}).
Binarizing the latent activations discards magnitude information. This is deliberate: latents whose magnitudes encode several distinct concepts~\cite{kopf2026capturing} are penalized precisely because we treat such magnitude-encoded multiplicity as insufficient disentanglement. FBMP nonetheless outperforms a magnitude-aware non-negative orthogonal matching pursuit \cite{bruckstein2008sparse} baseline in causal alignment at matched sparsity and without thresholding (Sec.~\ref{app:nnomp_results}).

\begin{algorithm}
\caption{Fully-Binary Matching Pursuit (FBMP) \label{alg:FBMP}}
\begin{algorithmic}[1]
\Require 
\parbox[t]{\linewidth}{%
Set of binarized latent activation vectors $\mathbf{z}^l_{\text{bin}} \!\in\! \{0,1\}^N$, with $l \in \{1, \dots, L\}$\\
Target attribute annotations $\mathbf{y} \in \{0,1\}^N$\\
Max iterations $k$ (max subset size), $\beta \leq 1$ (for concept selection criterion),\\
Binary similarity metric $\text{SIM}_{bin}$ (e.g., $F_\beta$).
}
\Ensure 
    Latents subset $\mathcal{S} \subset \{1, \dots,L\}$ that well approximate $\mathbf{y}$ 
\State $\mathbf{r}^0 \gets \mathbf{y}$ \Comment{Initialize residual}
\State $\hat{\mathbf{y}}^0 \gets \mathbf{0}$ \Comment{Initialize approximation}
\State $\mathcal{S} \gets \emptyset$ \Comment{Initialize empty subset of selected concepts}
\For{$\kappa = 0$ \textbf{to} $k-1$}
    \State $l^{\ast} \gets \arg\max_{l} \text{SIM}_{\text{bin}} (\mathbf{r}^\kappa, \mathbf{z}^l_{\text{bin}})$ \Comment{Select most-aligned concept}
    \State $\mathbf{r}^{\kappa+1} \gets \mathbf{r}^\kappa \land \lnot \mathbf{z}^{l^\ast}_{\text{bin}}$ \Comment{Update residual}
    \State $\hat{\mathbf{y}}^{\kappa+1} \gets \hat{\mathbf{y}}^{\kappa} \lor \mathbf{z}^{l^\ast}_{\text{bin}}$ \Comment{Update approximation}
    \If{$F_1(\mathbf{y}, \hat{\mathbf{y}}^{\kappa+1}) \leq F_1(\mathbf{y}, \hat{\mathbf{y}}^{\kappa})$} \Comment{Stopping criterion}
        \State \textbf{break}
    \EndIf
    \State $\mathcal{S} \gets \mathcal{S} \cup \{l^\ast\}$ \Comment{Add to subset}

\EndFor
\State \Return $\mathcal{S}$
\end{algorithmic}
\end{algorithm}

\noindent\textbf{Matching Score.} Given a set of concepts $\mathcal{S}_a$ matched to attribute $a$, the corresponding matching score is defined as the $F_1$ similarity between the coalition of selected latents and the ground-truth annotations $\textbf{y}_a$. The overall $\text{MATCHScore}$ is then obtained by averaging over attributes: 
\begin{equation}
\label{eq:matchscore_unnormalized}
    \text{MATCHScore}  = \frac{1}{A} \sum_{a=1}^A  F_1(\textbf{y}_a,\textstyle \bigwedge_{l\in\mathcal{S}_a}  \textbf{z}_\text{bin}^l)
\end{equation}
For one-to-one matching, we simply have $\mathcal{S}_a=\{l_a^*\}$ as defined in Eq.~\eqref{eq:F1_matching}. 
Finally, to enable fair comparisons between SAEs of different sizes, we introduce an \emph{adjusted} matching score. Larger dictionaries provide a larger pool of candidate latent units, which artificially inflates matching performance (Fig.~\ref{fig:app_matching_untrained}). To account for this effect, we subtract $F_{1,\text{rng}}$, the matching score achieved by an untrained SAE (random weights) using the same matching strategy, to compute the $\Delta\text{MATCHScore}$, which reflects improvement over a random baseline:
\begin{equation}
\label{eq:matchscore}
\Delta\text{MATCHScore} = \text{MATCHScore} - F_{1,\text{rng}}.
\end{equation}

\subsection{Evaluation via Targeted Attribute Perturbation Alignment}
A high latent-concept matching score indicates statistical alignment between the SAE's latent activations and the annotated attributes. However, correlation does not imply that a latent encodes the underlying semantic attribute. A latent may correlate with an attribute due to confounding structure, co-occurrence patterns, or background cues, without encoding the semantic factor itself. 
If a latent truly encodes an attribute, then changing that attribute while holding all other attributes constant should induce a structured and directionally consistent change in the latent activation. 
We use targeted perturbations to measure this change, by isolating a single semantic attribute while minimizing confounding variation. 
By constructing paired inputs $(\mathbf{x}, \hat{\mathbf{x}})$ that differ in exactly one annotated attribute, we can test whether latents matched to the perturbed attribute respond in the correct direction, increasing for additions and decreasing for removals.

Let $\mathbf{x}$ denote the original image embedding produced by a pretrained model $f$, and let $\hat{\mathbf{x}}$ denote the perturbed version that differs in exactly one attribute. 
Let $\mathbf{z}_\text{bin}, \hat{\mathbf{z}}_\text{bin} \in \mathbb{R}^L$ denote the binarized SAE latent representations of $\mathbf{x}$ and $\hat{\mathbf{x}}$, respectively, with dictionary size $L$.
Using the latent-concept matching defined previously, let $I_{\text{add}} \subset \{1,\dots,L\}$ denote the set of latents matched to the attribute being added, and $I_{\text{rem}}$ those matched to the attribute being removed (see Fig.~\ref{fig:full_method}). We aggregate over each matched set via the maximum (equivalently, a logical OR): the concept is considered active if at least one of its latents fires. The signed response for a single image pair is:
\begin{equation}
    \delta_{\text{add}} = \max_{i \in I_{\text{add}}} \hat{\mathbf{z}}^i_\text{bin} - \max_{i \in I_{\text{add}}} \mathbf{z}_\text{bin}^i,
    \quad \text{and} \quad
    \delta_{\text{rem}} = \max_{i \in I_{\text{rem}}} \hat{\mathbf{z}}^i_{bin} - \max_{i \in I_{\text{rem}}} \mathbf{z}^i_{bin}.
\end{equation}
Averaging over all $P$ image pairs in the dataset yields:
\begin{equation}
    \Delta_{\text{add}} = \frac{1}{P}\sum_{p} \delta_{\text{add}}^{(p)},
    \quad \text{and} \quad
    \Delta_{\text{rem}} = \frac{1}{P}\sum_{p} \delta_{\text{rem}}^{(p)}.
\end{equation}
Finally, we define the \textbf{\underline{T}}argeted \textbf{\underline{A}}ttribute \textbf{\underline{P}}erturbation \textbf{\underline{A}}lignment \textbf{\underline{S}}core (TAPAScore) as:

\begin{equation}
\label{eq:tapas}
    \text{TAPAScore} = \Delta_{\text{add}} - \Delta_{\text{rem}}.
\end{equation}
For datasets containing only removal perturbations (e.g., synCOCO), we set $\Delta_{\text{add}} = 0$. A positive TAPAScore indicates that matched latents respond selectively and in the correct direction under targeted attribute perturbations.
\subsection{Synthetic Dataset generation}
\label{subsec:dataset_generation}

Evaluating TAPAScore requires image pairs that differ in exactly one semantic attribute while keeping all other factors constant. Since no such benchmark exists for naturalistic images, we construct two synthetic datasets tailored to this requirement: synCUB, which targets fine-grained attribute perturbations in single-object bird images, and synCOCO, which targets object removal in complex multi-object scenes. To the best of our knowledge, this constitutes the first intervention-style benchmark for SAE evaluation in naturalistic images, as prior work has been largely restricted to simple controlled settings~\cite{fel2025archetypal}.
\begin{figure}
    \centering
        \vspace{-2.5em}
    \includegraphics[width=1\linewidth,  trim=0 0.6cm 0 0]{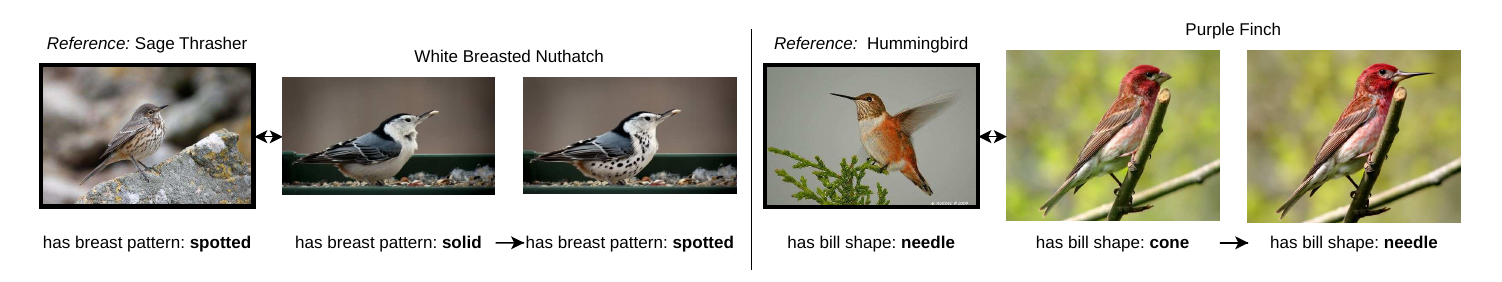}
    \caption{synCUB pairs: a reference image guides the target attribute (e.g., breast pattern: solid $\rightarrow$ spotted), while the base image preserves identity, pose, and background.}
    \label{fig:syncub_qualitative}
        \vspace{-2.5em}

\end{figure}

\noindent\textbf{Synthetic CUB (synCUB, attribute perturbations).}
CUB-200-2011~\cite{CUB_dataset} provides dense, per-instance attribute annotations across 312 attributes covering color, shape, and pattern across bird parts, making it a natural starting point for attribute-level intervention evaluation. We construct synCUB following the SUB benchmark~\cite{bader2025sub}, which introduces attribute variation by generating prototypical class examples with uncommon attributes. However, SUB pairs images across different class instances rather than the same bird before and after a single attribute change, making it unsuitable for image-level intervention. We therefore construct synCUB, where each pair $(\mathbf{x}, \hat{\mathbf{x}})$ differs in exactly one target attribute while all other annotated attributes remain unchanged (Fig.~\ref{fig:syncub_qualitative}), restricted to the 33-class subset and 45 attribute concepts of SUB. For each target attribute, a base image and a reference image exhibiting the desired state are passed to Flux2~\cite{flux-2-2025}, which edits the base image (preserving identity, pose, and background) while the reference guides the target attribute. We validate edits with an attribute predictor and manually curate failures (Appendix Sec.~\ref{sup:dataset_curation}).

\begin{figure}
    \centering
        \vspace{-2.5em}
    \includegraphics[width=1\linewidth, trim=0 0.6cm 0 0]{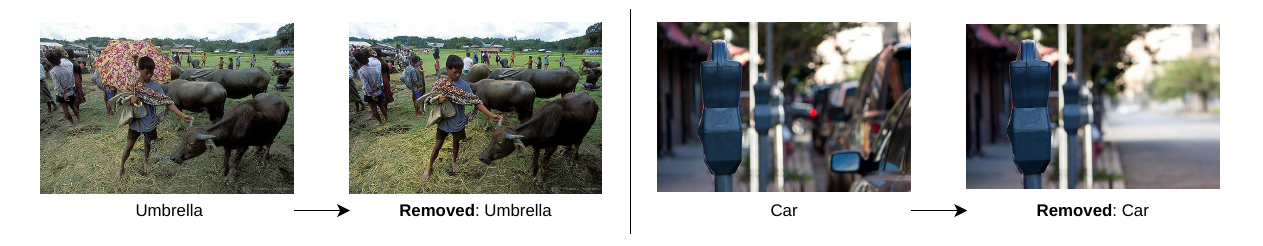}
    \caption{synCOCO pairs: the target object (e.g., umbrella, car) is removed while the remaining scene is preserved.}
    \label{fig:syncoco_qualitative}
        \vspace{-2.5em}
\end{figure}

\noindent\textbf{Synthetic COCO (synCOCO, object removal perturbations).}
While synCUB evaluates attribute-level alignment in a controlled single-object setting, real-world scenes are considerably more complex. MS-COCO~\cite{lin2014coco} provides a natural complement: its images contain multiple objects, diverse backgrounds, and rich compositional structure absent in CUB. As COCO has no attribute annotations, we treat object labels as high-level semantic concepts and construct synCOCO, where each pair $(\mathbf{x}, \hat{\mathbf{x}})$ differs in one object label, with all other objects preserved (Fig.~\ref{fig:syncoco_qualitative}). We select the target object by lowest instance count, breaking ties by largest area, and remove all its instances via Flux2~\cite{flux-2-2025} conditioned on the original image and a fixed removal prompt. We verify removal with the same automatic classifier check as synCUB, followed by manual validation of every retained pair. Full details for both datasets are in Appendix Sec.~\ref{sup:dataset}.
\section{Experimental Results}
\textbf{Experimental Settings.}
We evaluate SAE interpretability on two controlled benchmarks derived from CUB-200-2011~\cite{CUB_dataset} and MS-COCO~\cite{lin2014coco}. For both datasets, we construct synthetic intervention benchmarks (synCUB and synCOCO) as described in Sec.~\ref{subsec:dataset_generation}. Latent-concept matching is computed on the original CUB and COCO datasets, while perturbation alignment is evaluated on the synthetic benchmarks. 
For each dataset, we extract image embeddings from two pretrained vision models, CLIP (ViT-L-14 backbone) and DINOv2 (ViT-S-14 backbone), and train SAEs on these embeddings. We compare four SAE variants: JumpReLU, TopK, BatchTopK, and Matryoshka SAEs, across dictionary sizes \{128, 256, 512, 1024, 2048, 4096\}. Unless stated otherwise, all SAEs except JumpReLU use TopK sparsity with $K{=}32$, which is the average occurrence of attributes in CUB; a sweep over the sparsity $K$ in the Appendix (Sec.~\ref{app:sparsity_sweep}) confirms that moderate sparsity levels provide the best trade-off between matching and perturbation alignment.
Full training statistics, including reconstruction loss and sparsity metrics, are listed in the Appendix in Sec.~\ref{app:subsec:sae_training}. 
For state-of-the-art comparison, we compare functional proxy metrics commonly used in prior work, namely FMS~\cite{harle2025measuring}, MS~\cite{pach2025sparse}, and CKNNA~\cite{zaigrajew2025interpreting}. We evaluate monosemanticity-based scores (FMS \cite{harle2025measuring} and MS \cite{pach2025sparse}), aggregated over all latents. While monosemanticity-based scores were not designed as a metric for evaluating a full SAE, we treat the average over latents as a proxy: if an SAE contains more monosemantic latents, its average score should indicate higher interpretability. For our selection criteria, we evaluate latent-concept matching using classical $F_1$ (with $k{=}1$) and $F_\beta$-based matching pursuit (FBMP, with $k{=}3$) across $\beta \in \{0.25, 0.5, 1\}$; an analysis of metric sensitivity over $k$ is provided in the Appendix Sec.~\ref{app:subsec:tapas_over_k}. As a supervised upper bound, we additionally train per-attribute logistic regression probes on the raw embeddings and evaluate their thresholded outputs within the same pipeline (Figs.~\ref{fig:matching_and_tapas_cub} and~\ref{fig:matching_and_tapas_coco}).
%
\subsubsection{Sanity Check Analysis.}
\label{sec:sae_based_tests}
\begin{figure}[t]
    \centering
    \begin{subfigure}{0.49\textwidth}
        \centering
        \includegraphics[width=\linewidth]{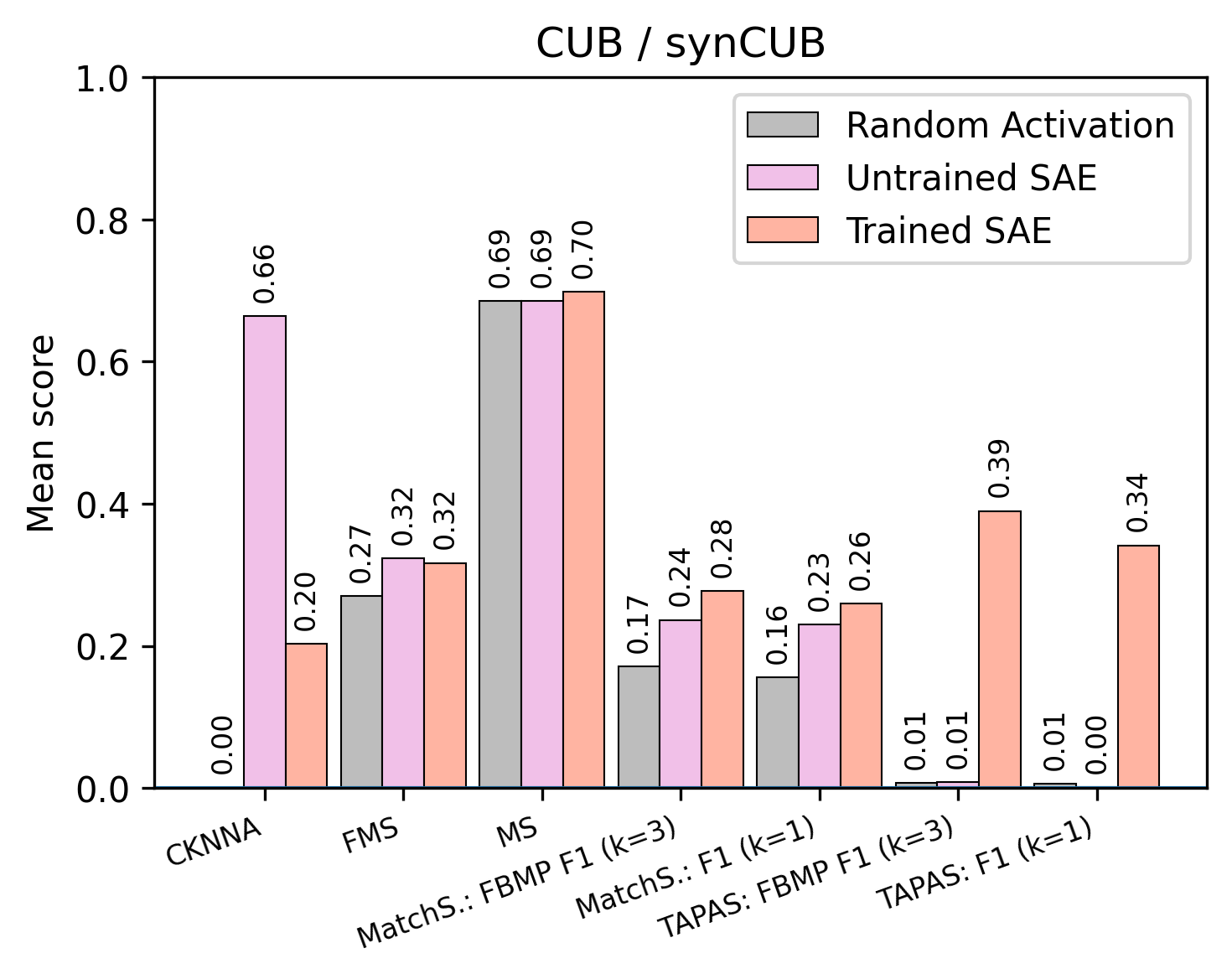}
        \label{fig:matching_comparison_dict256}
    \end{subfigure}
    \hfill
    \begin{subfigure}{0.49\textwidth}
        \centering
        \includegraphics[width=\linewidth]{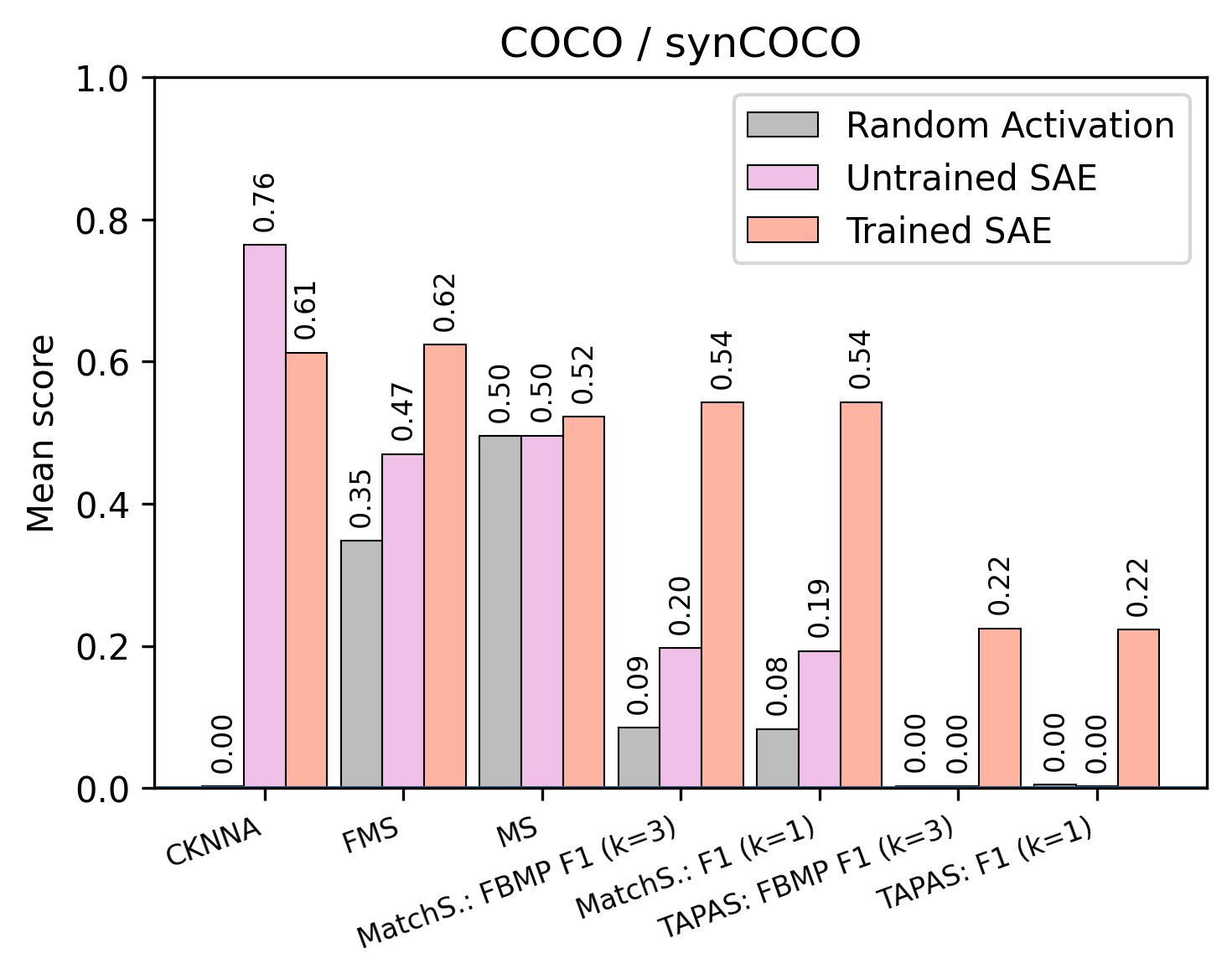}
        \label{fig:matching_comparison_dict2048}
    \end{subfigure}
\vspace{-2.3em}  
\caption{Metrics failure mode comparison aggregated over all dictionary sizes for CUB (left) and COCO (right) with CLIP, across three conditions: trained SAEs, an untrained TopK SAE, and random activations. TAPAScore (computed on the synthetic datasets) and MATCHScore with FBMP clearly drop under untrained and random conditions, while FMS and MS show little sensitivity, and CKNNA inflates for the untrained SAE.}
\label{fig:matching_failure_mode}
    \vspace{-2em}

\end{figure}
Evaluating interpretability metrics is inherently challenging due to the absence of reliable ground truth explanations \cite{hedstrom2023meta, klotz2025effectiveness}. Drawing on the disentanglement literature, where failure modes of common metrics are well-documented \cite{carbonneau2022measuring}, we test whether a metric can distinguish meaningful concepts from spurious or random correspondence. Concretely, we compare three conditions: trained SAEs (with scores aggregated over all dictionary sizes and SAE variants), an untrained TopK SAE, and random activations. 
We report results for CKNNA \cite{zaigrajew2025interpreting}, FMS \cite{harle2025measuring}, and MS \cite{pach2025sparse}, alongside our MATCHScore with two criteria (FBMP $F_1$ with $k{=}3$ and $F_1$ with $k{=}1$) and the TAPAScore for the respective matching. To compare with the untrained baseline, we evaluate the unnormalized matching score as described in Eq.~\ref{eq:matchscore_unnormalized}. The metrics and MATCHScores are calculated on the respective dataset (CUB or COCO), whereas TAPAS is calculated on the corresponding synthetic version. \raggedbottom

From Fig.~\ref{fig:matching_failure_mode}, one can observe that both matching variants exhibit a pronounced drop when replacing trained SAEs with either untrained weights or random activations. For both datasets, the MATCHScores for FBMP with selection criterion $F_1$ ($k{=}3$) and $F_1$ ($k{=}1$) show clear absolute decreases in matching $F_1$ score under the untrained and random conditions. In contrast, MS shows limited sensitivity to whether the representation is trained, untrained, or random across both datasets, while FMS shows this limitation only on CUB but not on COCO. CKNNA behaves inconsistently: while it shows a similar drop for random activations, it exhibits a strong absolute increase for the untrained SAE, indicating that the untrained SAE achieves higher CKNNA scores than the trained models. TAPAScore shows the strongest separation, dropping to near zero for both untrained and random conditions on both datasets.
Overall, only our matching variants and TAPAScore pass the intended sanity checks, clearly distinguishing trained SAEs from both untrained and random conditions. 
\subsection{Latent to Concept Matching}
\label{sec:matching_results}
  \begin{figure*}[t]
  \centering
    \begin{subfigure}[t]{0.24\textwidth}
      \centering
      \includegraphics[width=\linewidth]{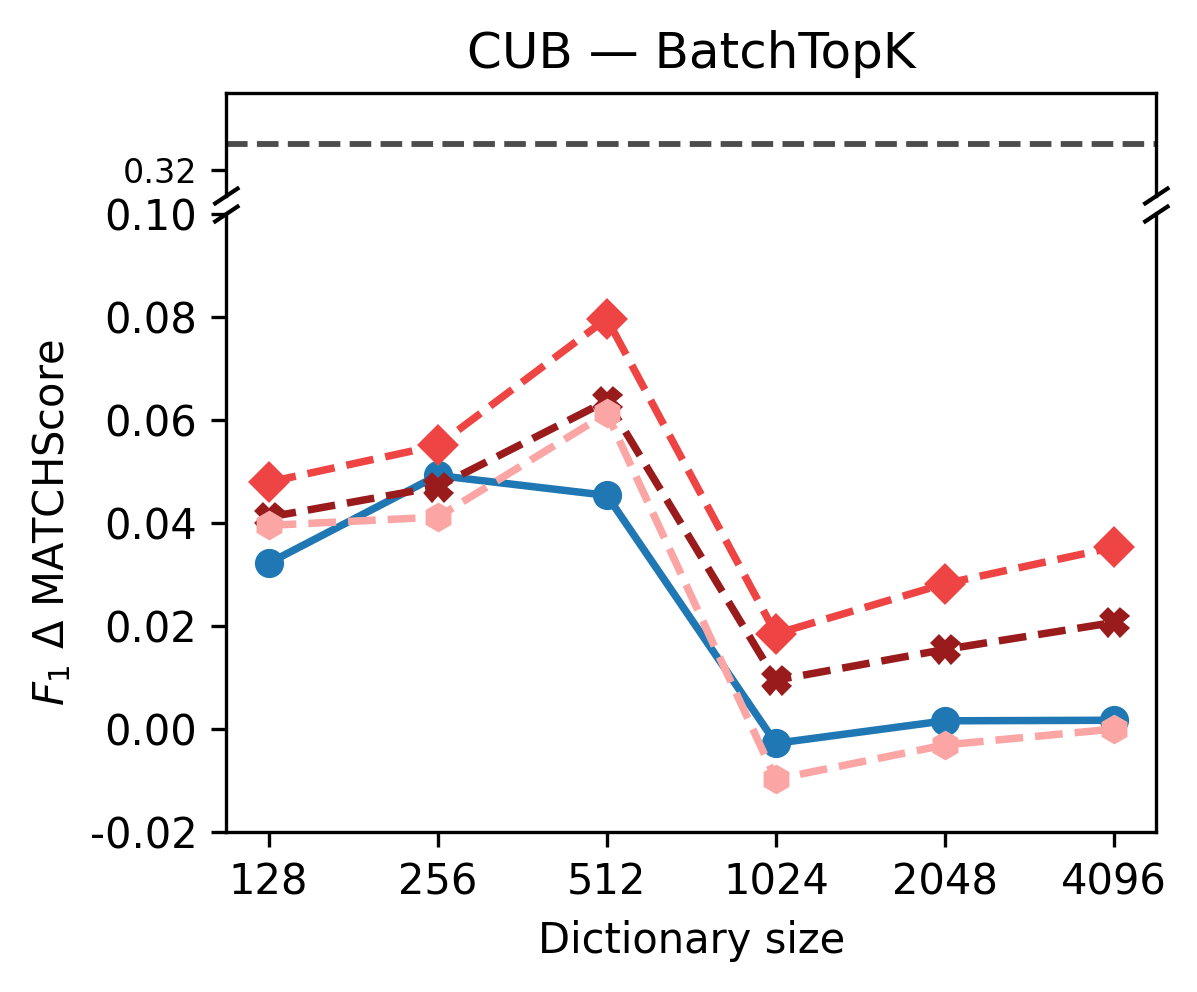}
    \end{subfigure}
    \hfill
    \begin{subfigure}[t]{0.24\textwidth}
      \centering
      \includegraphics[width=\linewidth]{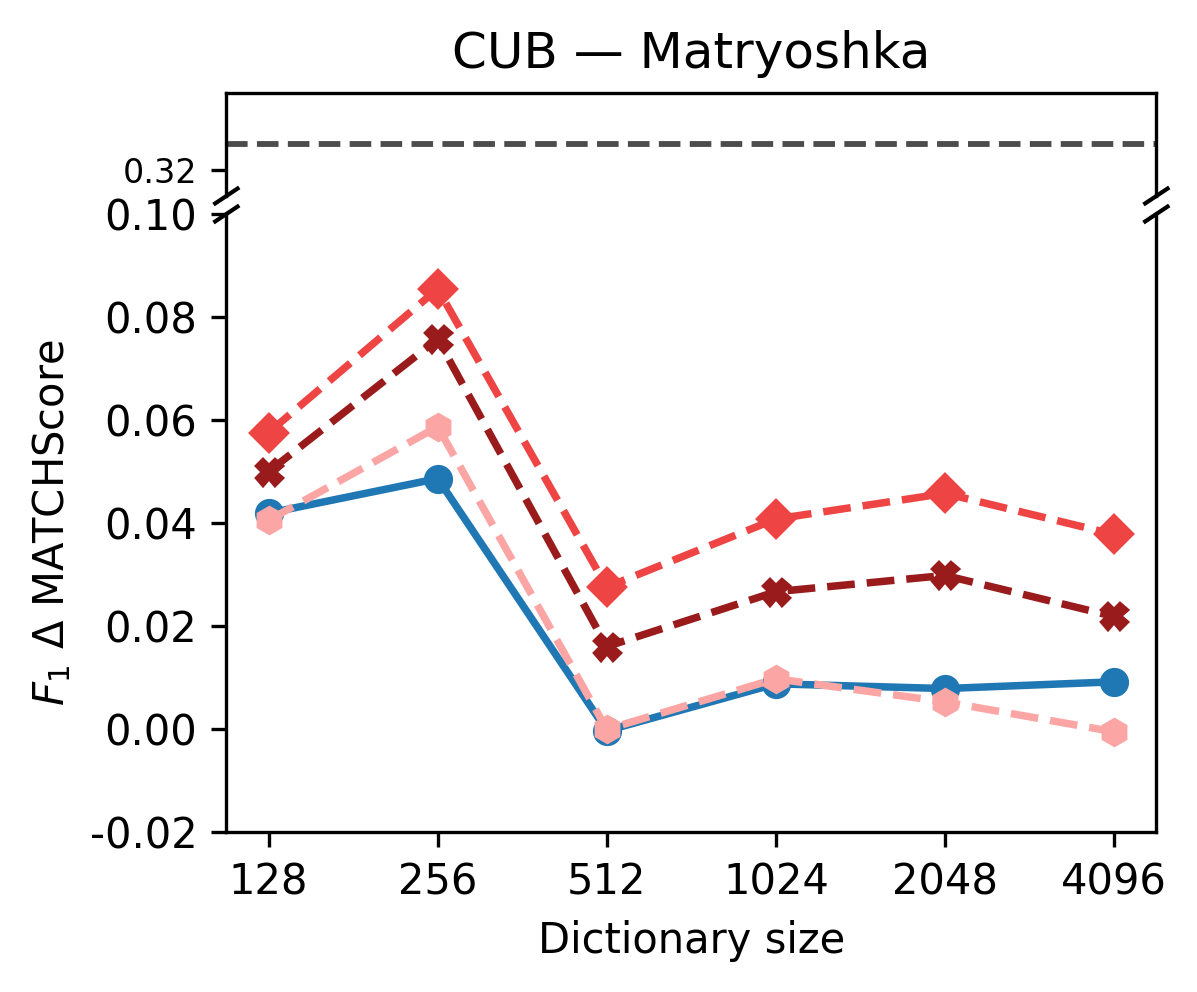}
    \end{subfigure}
    \hfill
    \begin{subfigure}[t]{0.24\textwidth}
      \centering
      \includegraphics[width=\linewidth]{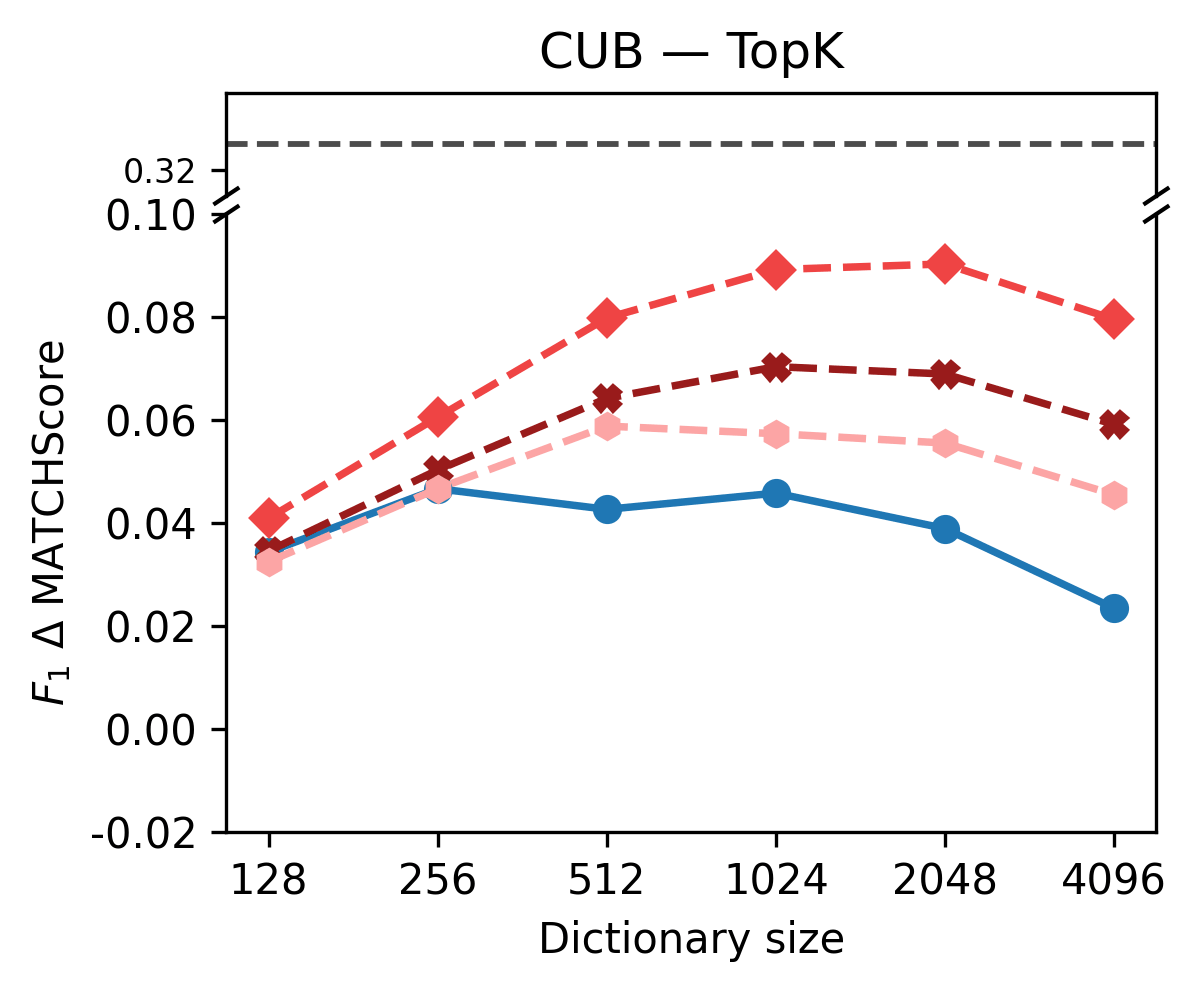}
    \end{subfigure}
    \hfill
    \begin{subfigure}[t]{0.24\textwidth}
      \centering
      \includegraphics[width=\linewidth]{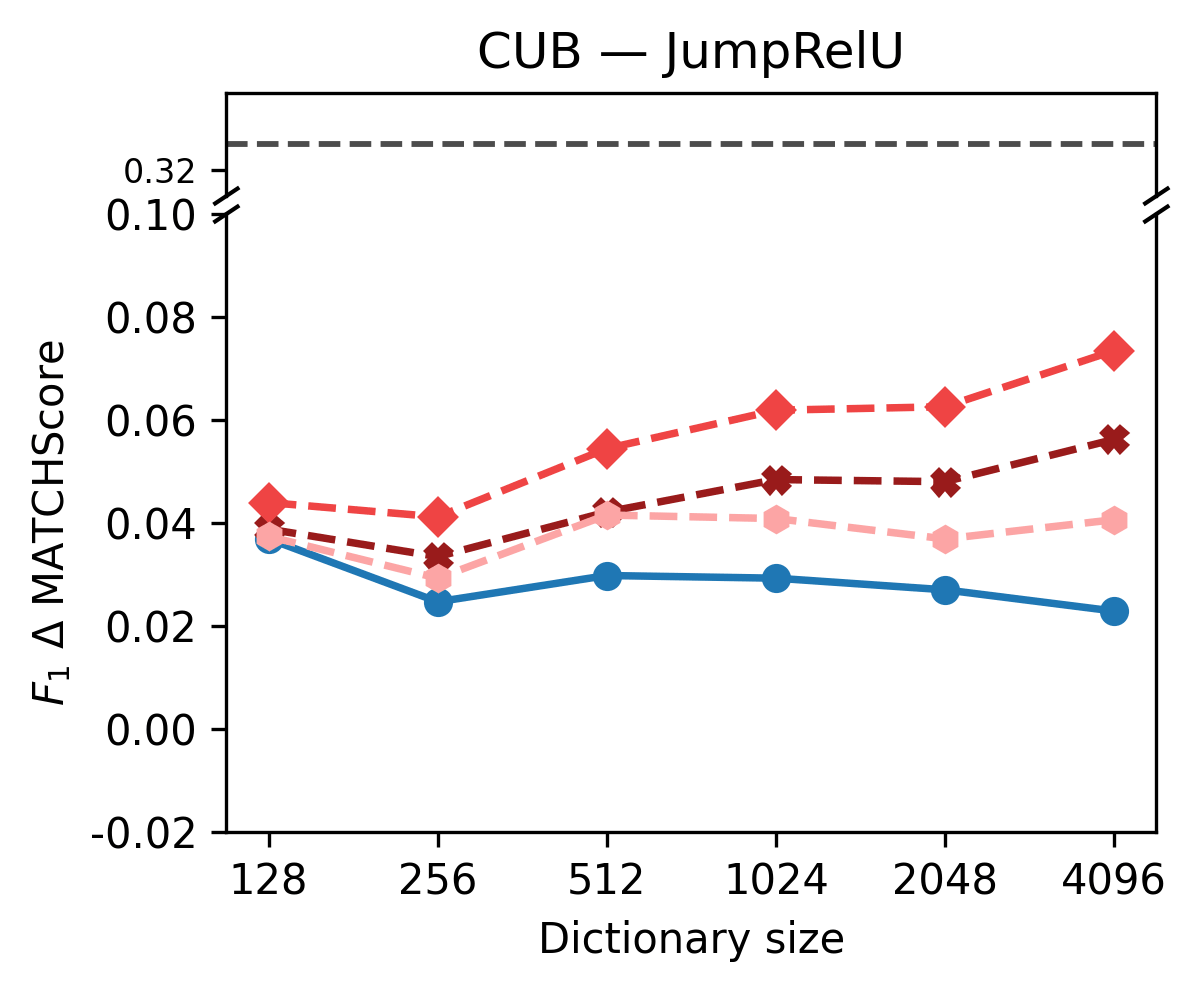}
    \end{subfigure}

    \vspace{-1mm}
  
    \begin{subfigure}[t]{0.24\textwidth}
      \centering
      \includegraphics[width=\linewidth]{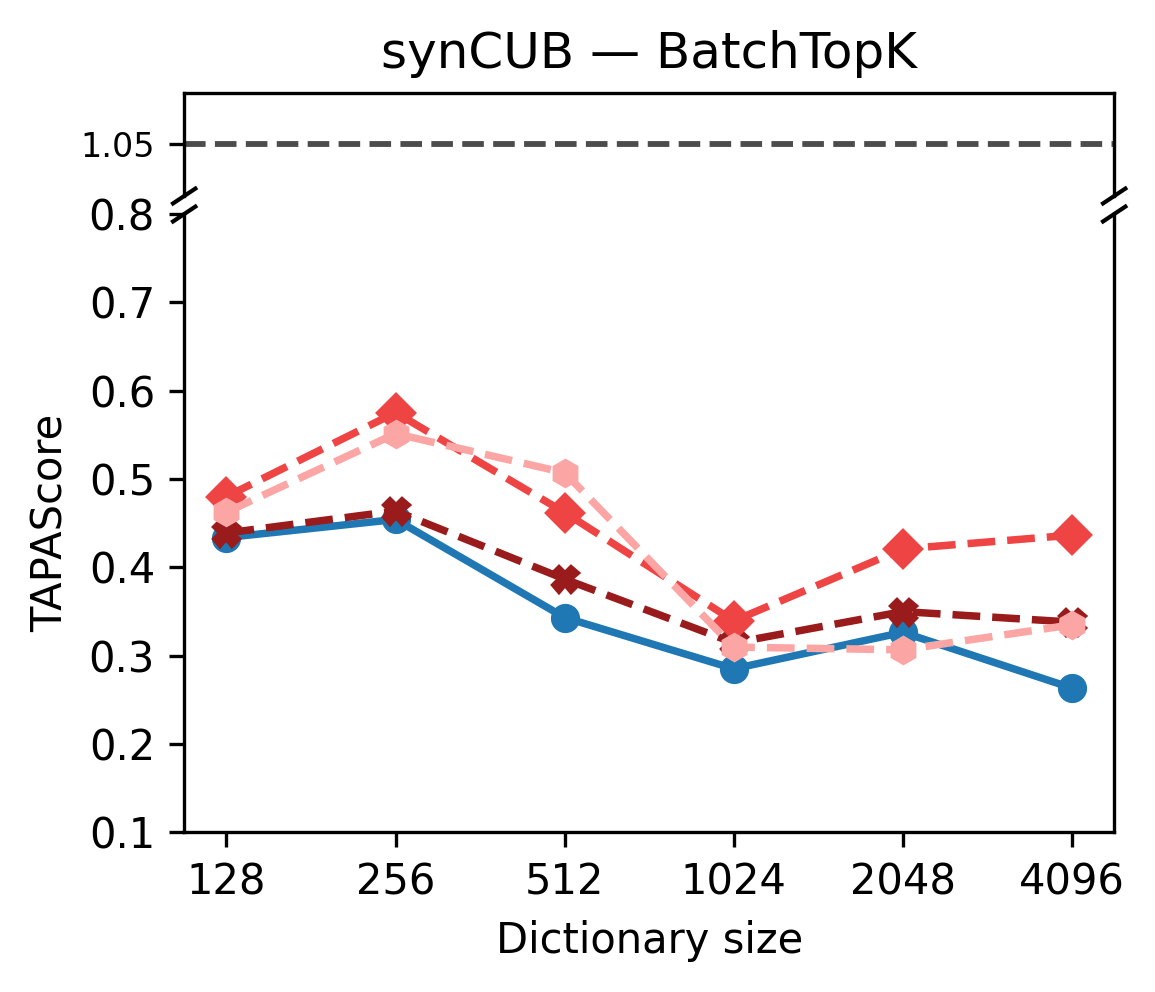}
    \end{subfigure}
    \hfill
    \begin{subfigure}[t]{0.24\textwidth}
      \centering
      \includegraphics[width=\linewidth]{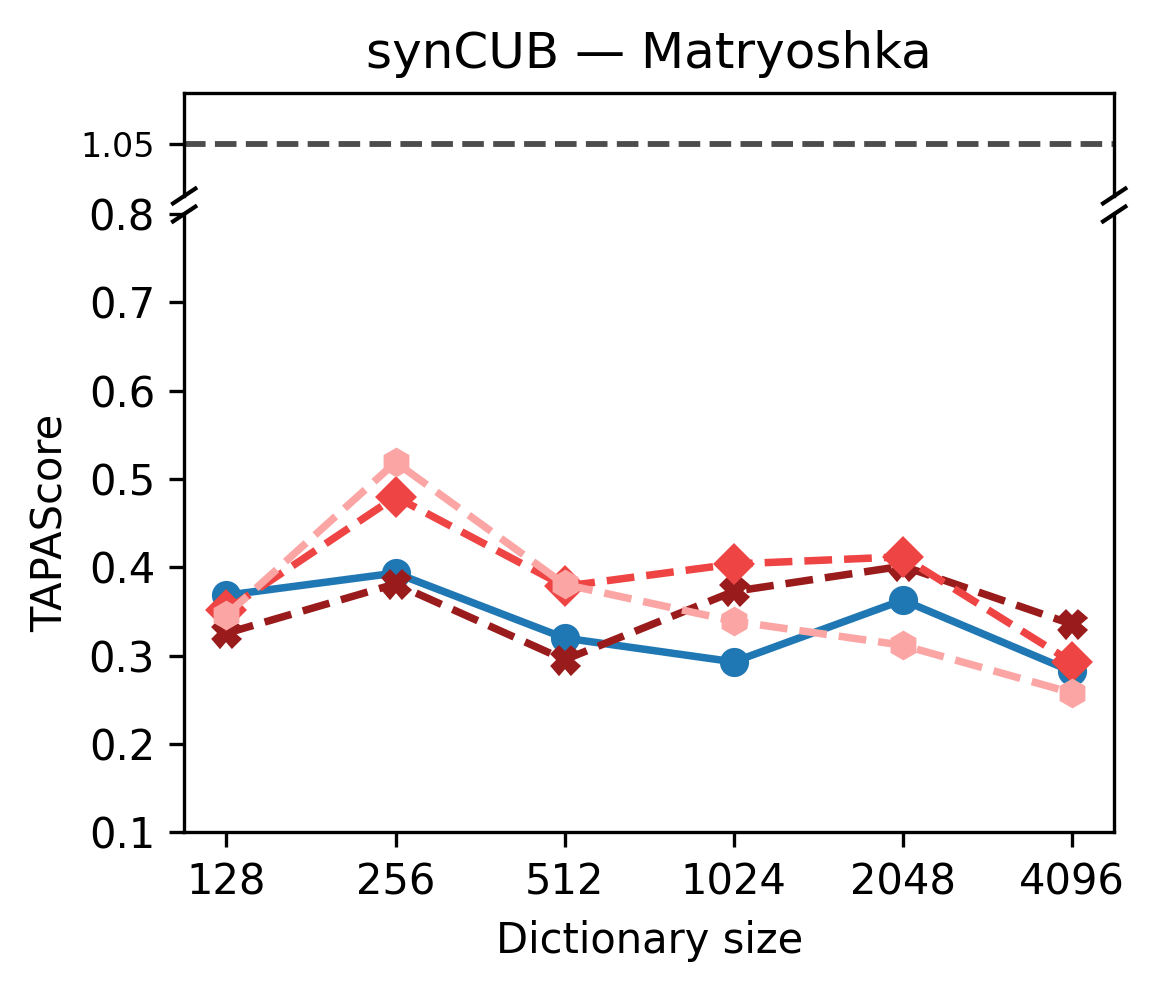}
    \end{subfigure}
    \hfill
    \begin{subfigure}[t]{0.24\textwidth}
      \centering
      \includegraphics[width=\linewidth]{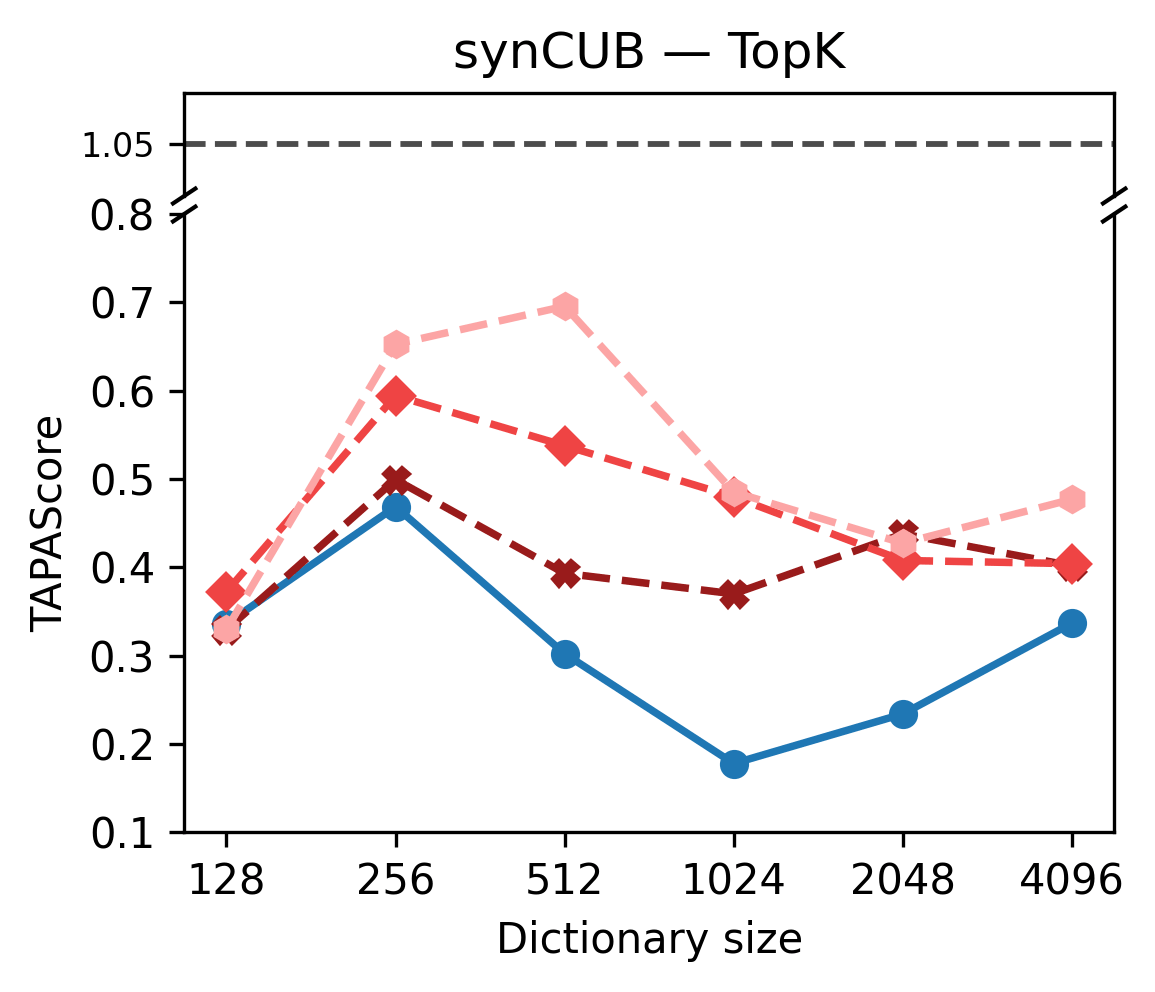}
    \end{subfigure}
    \hfill
    \begin{subfigure}[t]{0.24\textwidth}
      \centering
      \includegraphics[width=\linewidth]{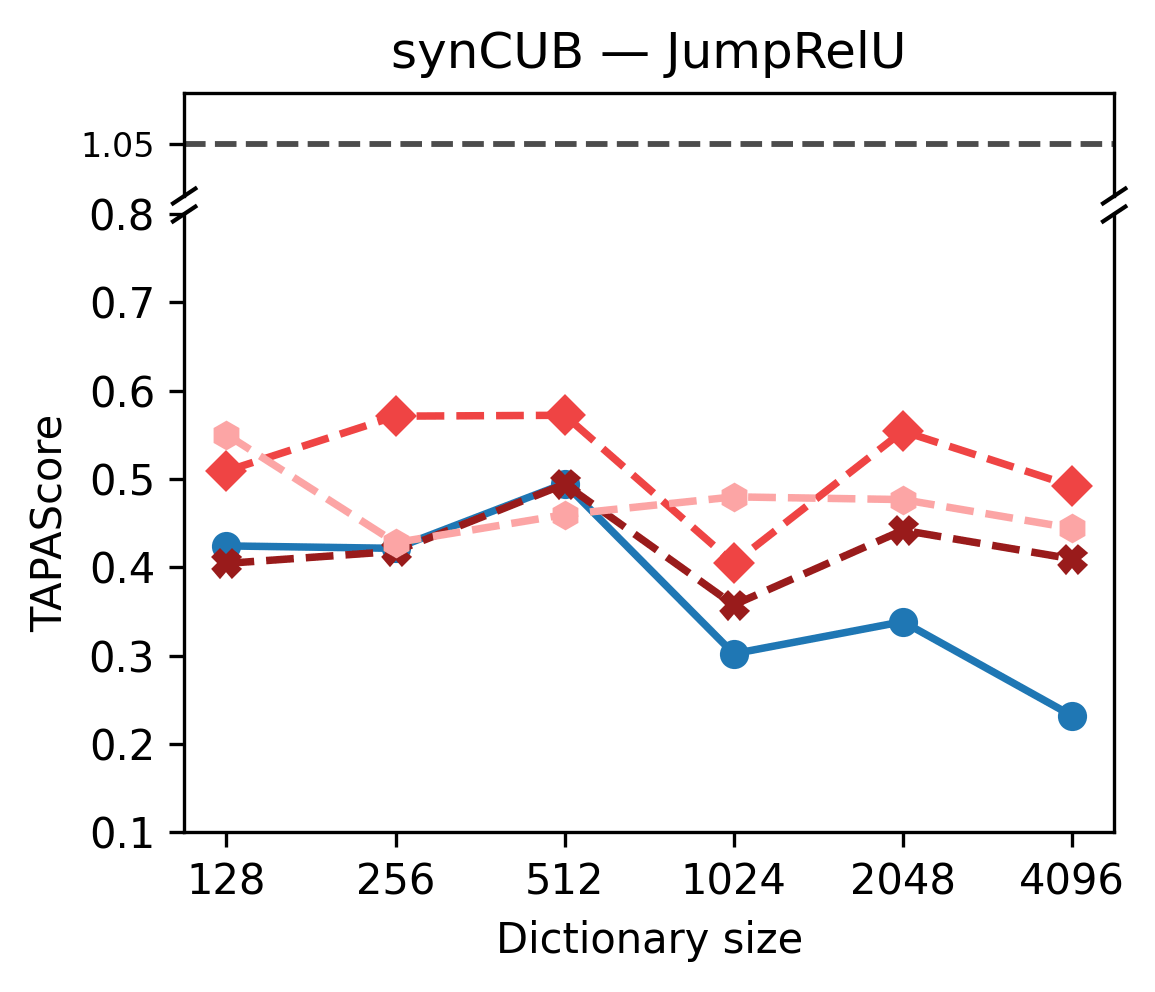}
    \end{subfigure}

    \vspace{-1mm}
  
    \includegraphics[width=\linewidth]{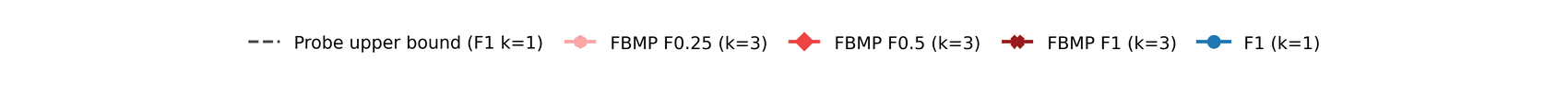}

  \vspace{-1.2em}
  \caption{$F_1$ $\Delta$MATCHScore on CUB (\textit{top row}) and TAPAScore on synCUB (\textit{bottom row}) as a function of dictionary size for SAEs trained on CLIP embeddings of CUB, across BatchTopK, Matryoshka, TopK and JumpReLU SAE variants (\textit{left to right}) and different matching criteria. The gray dashed line marks the supervised linear-probe upper bound.}
  \label{fig:matching_and_tapas_cub}
      \vspace{-2em}

  \end{figure*}

  \begin{figure*}[t]
  \centering
    \begin{subfigure}[t]{0.24\textwidth}
      \centering
      \includegraphics[width=\linewidth]{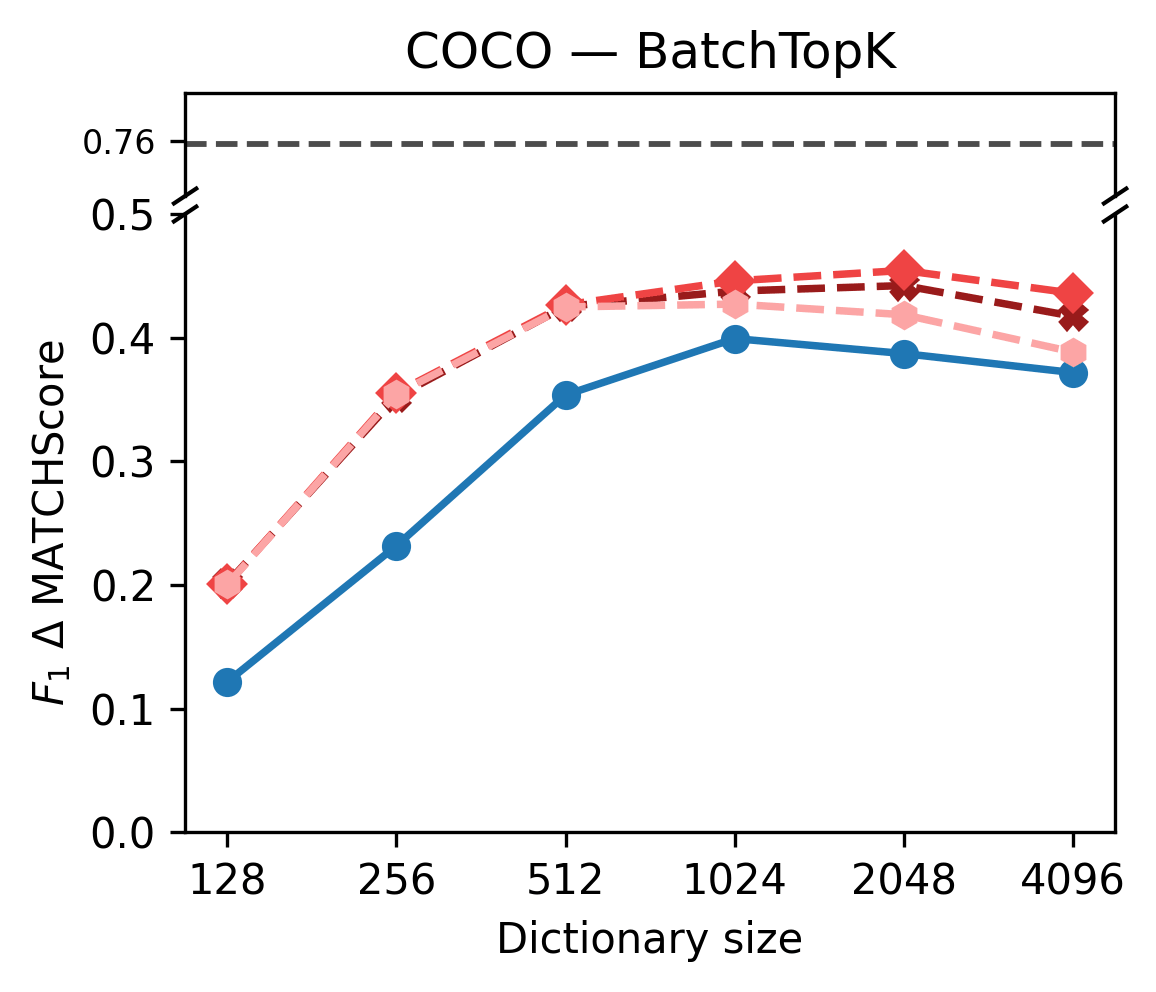}
    \end{subfigure}
    \hfill
    \begin{subfigure}[t]{0.24\textwidth}
      \centering
      \includegraphics[width=\linewidth]{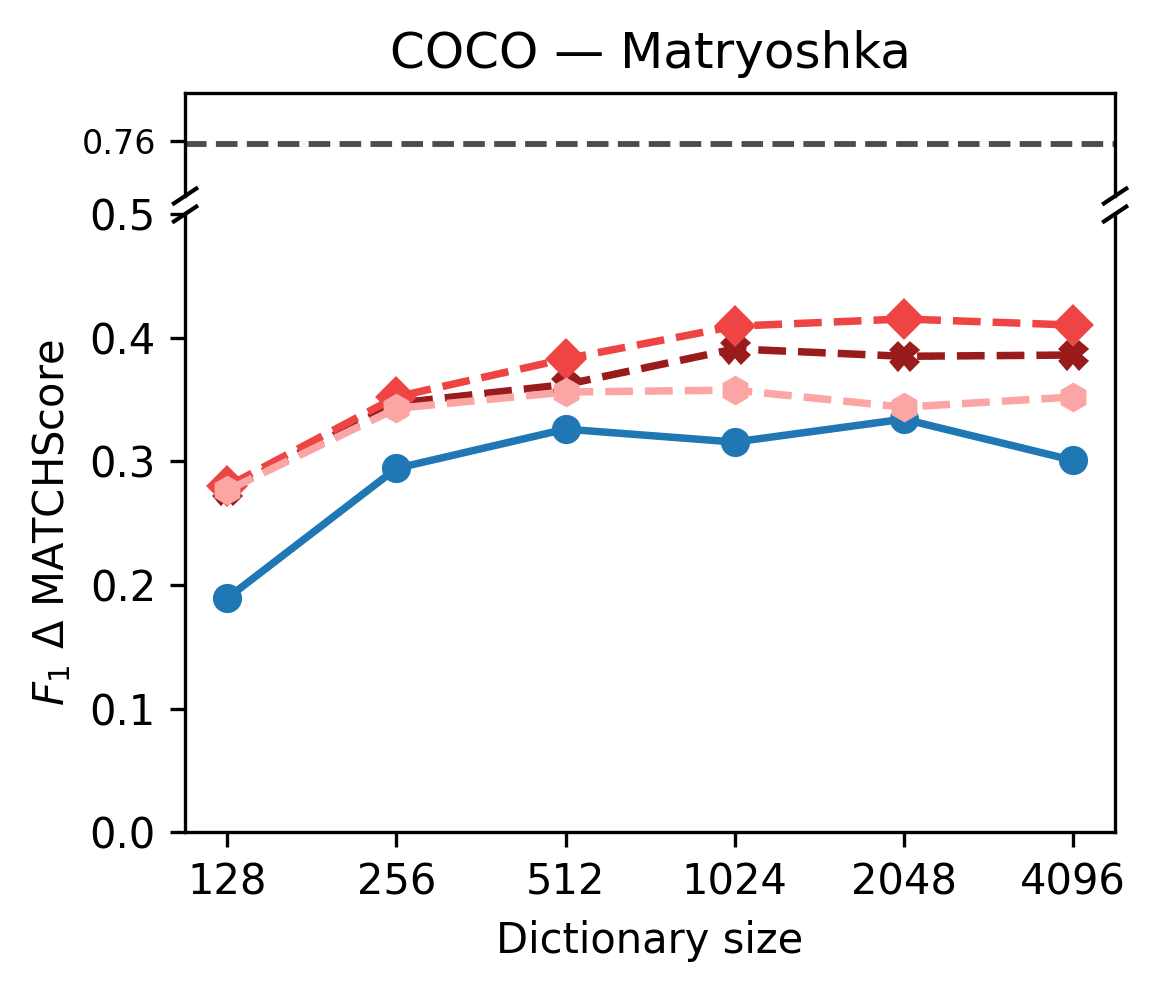}
    \end{subfigure}
    \hfill
    \begin{subfigure}[t]{0.24\textwidth}
      \centering
      \includegraphics[width=\linewidth]{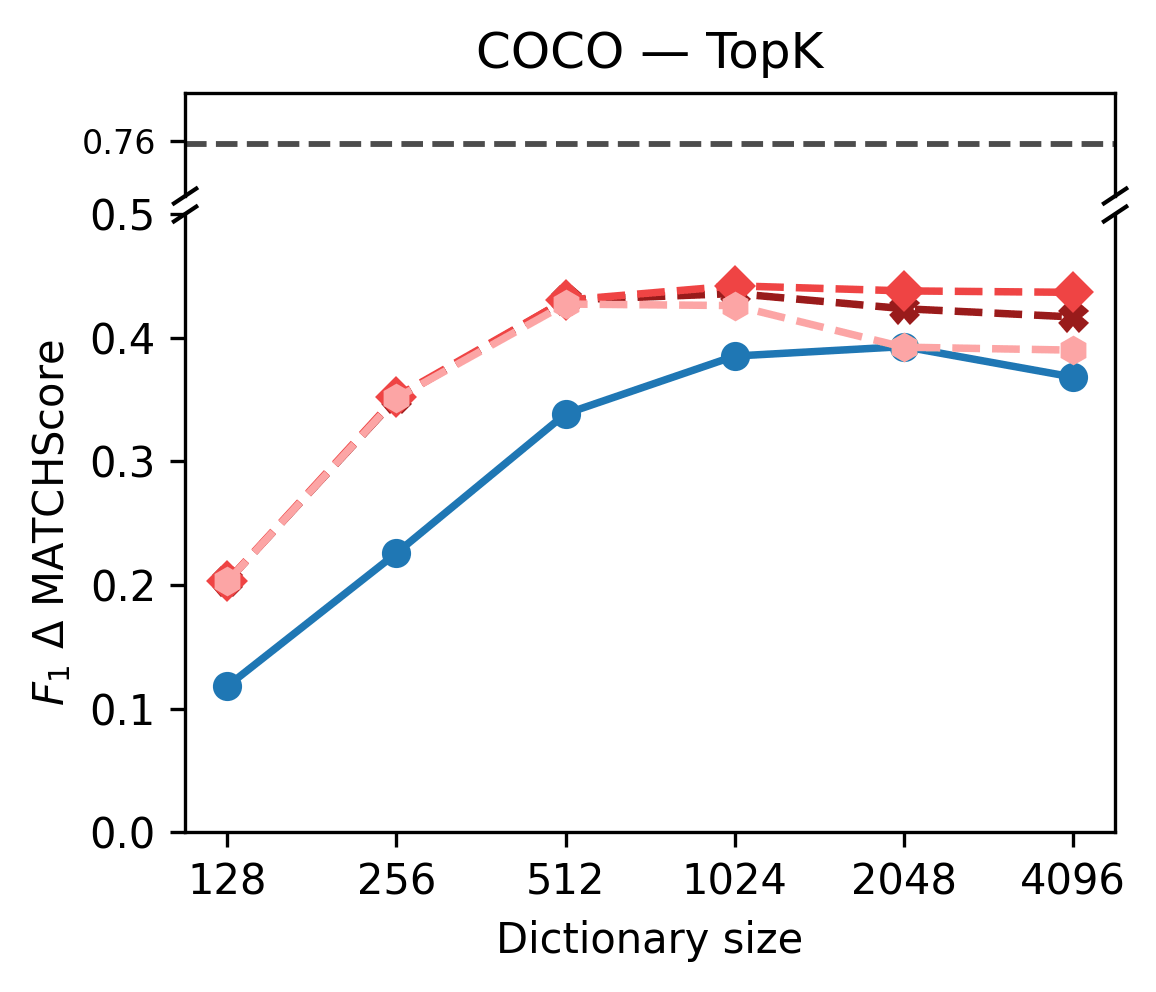}
    \end{subfigure}
    \hfill
    \begin{subfigure}[t]{0.24\textwidth}
      \centering
      \includegraphics[width=\linewidth]{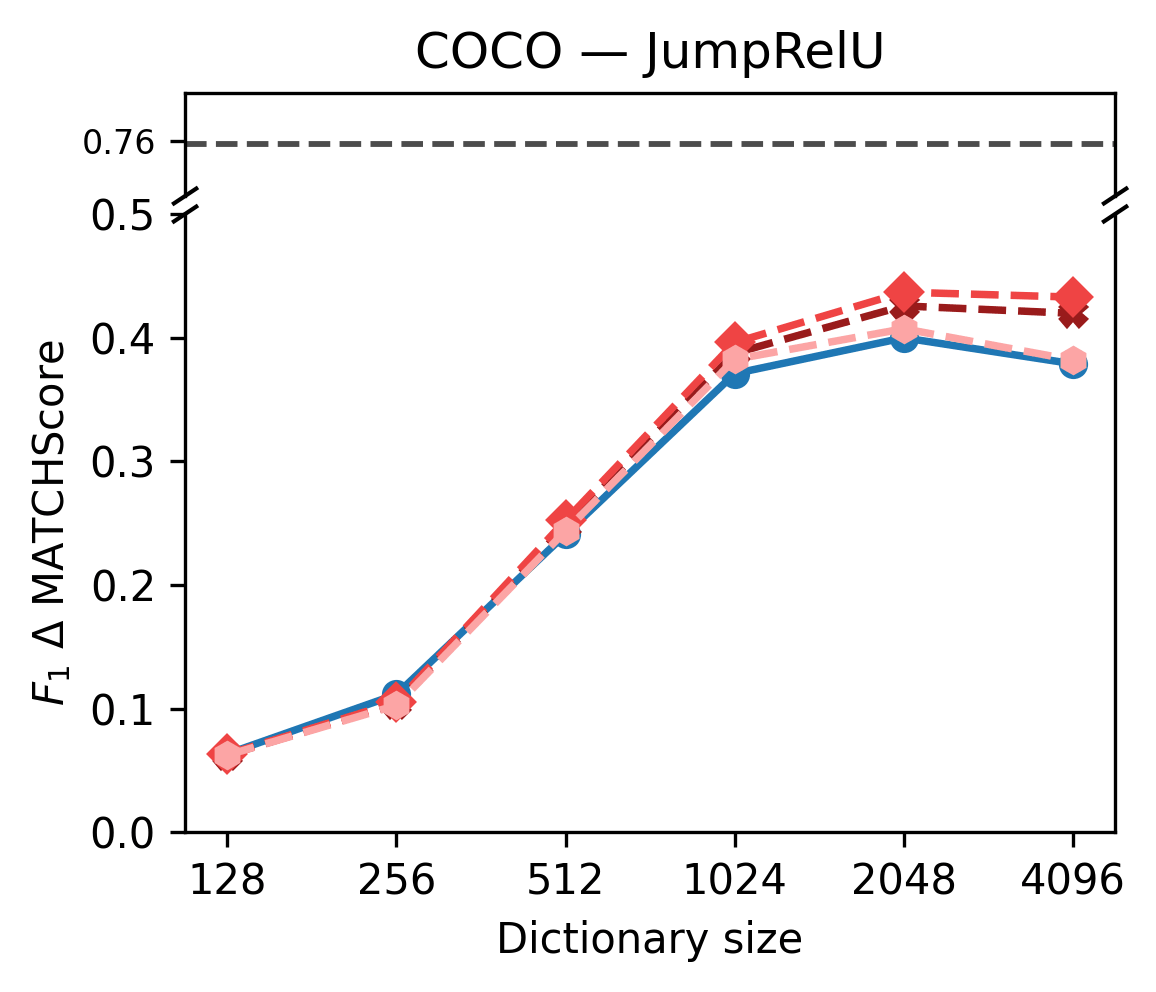}
    \end{subfigure}

    \vspace{-1mm}

    \begin{subfigure}[t]{0.24\textwidth}
      \centering
      \includegraphics[width=\linewidth]{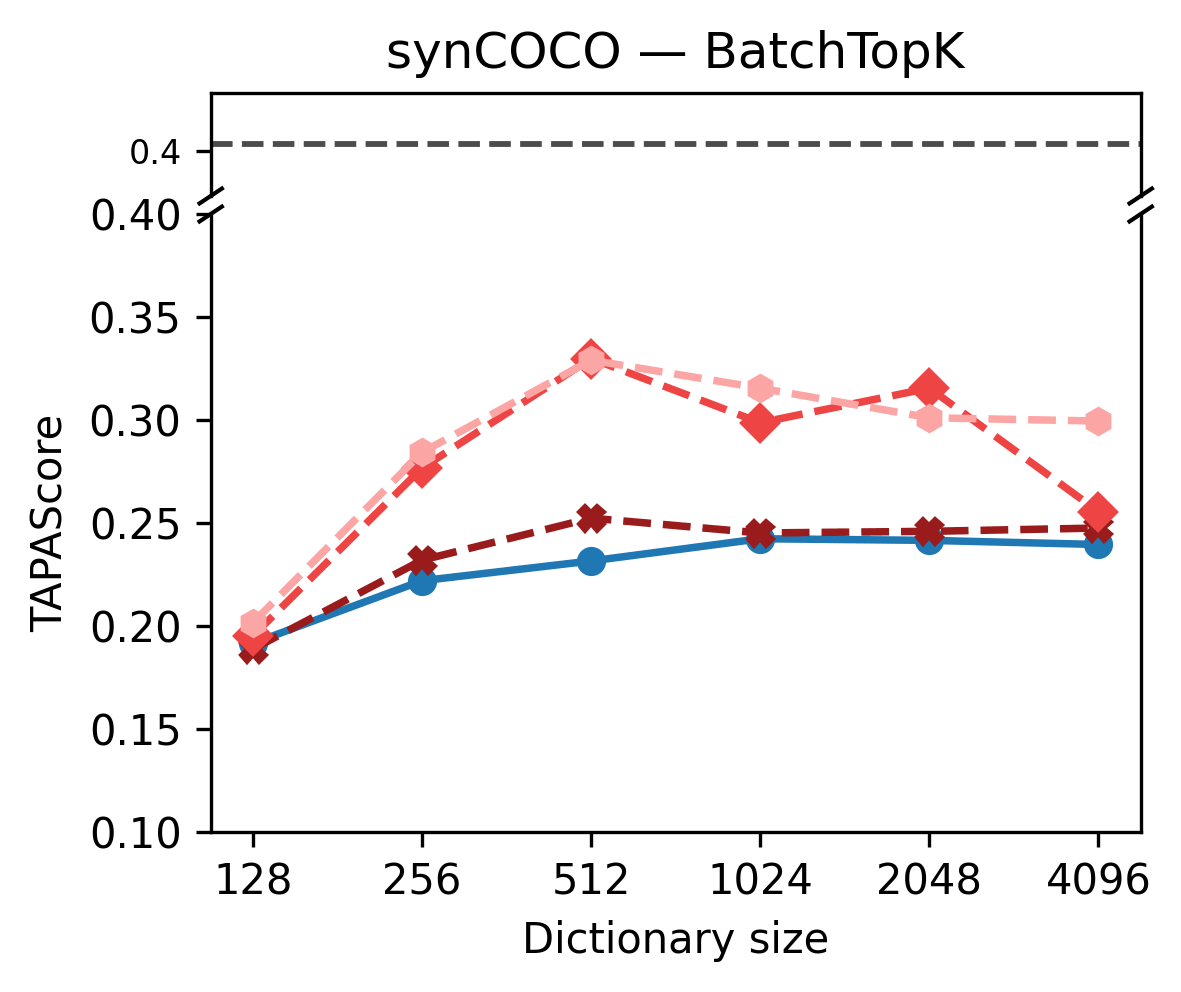}
    \end{subfigure}
    \hfill
    \begin{subfigure}[t]{0.24\textwidth}
      \centering
      \includegraphics[width=\linewidth]{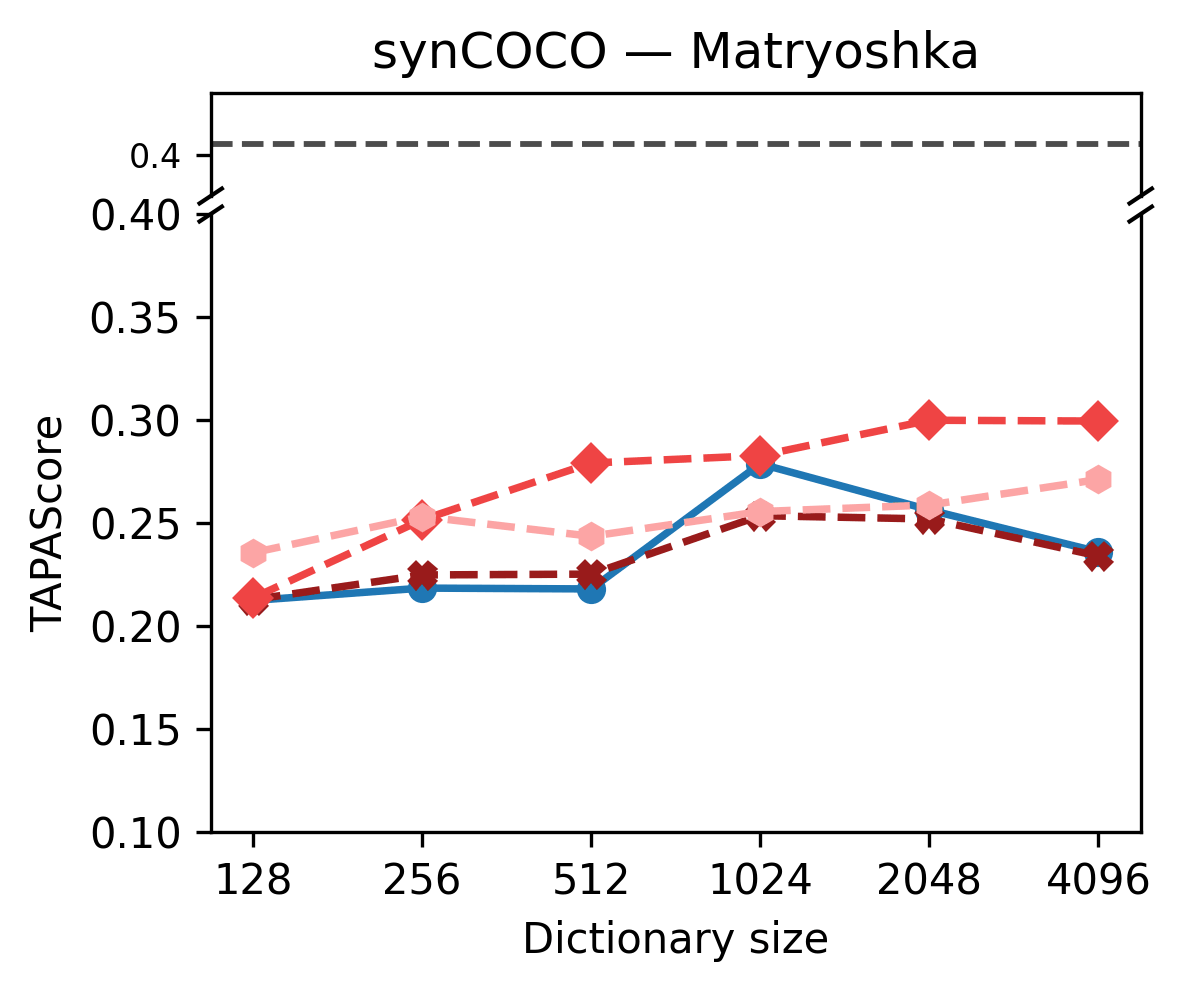}
    \end{subfigure}
    \hfill
    \begin{subfigure}[t]{0.24\textwidth}
      \centering
      \includegraphics[width=\linewidth]{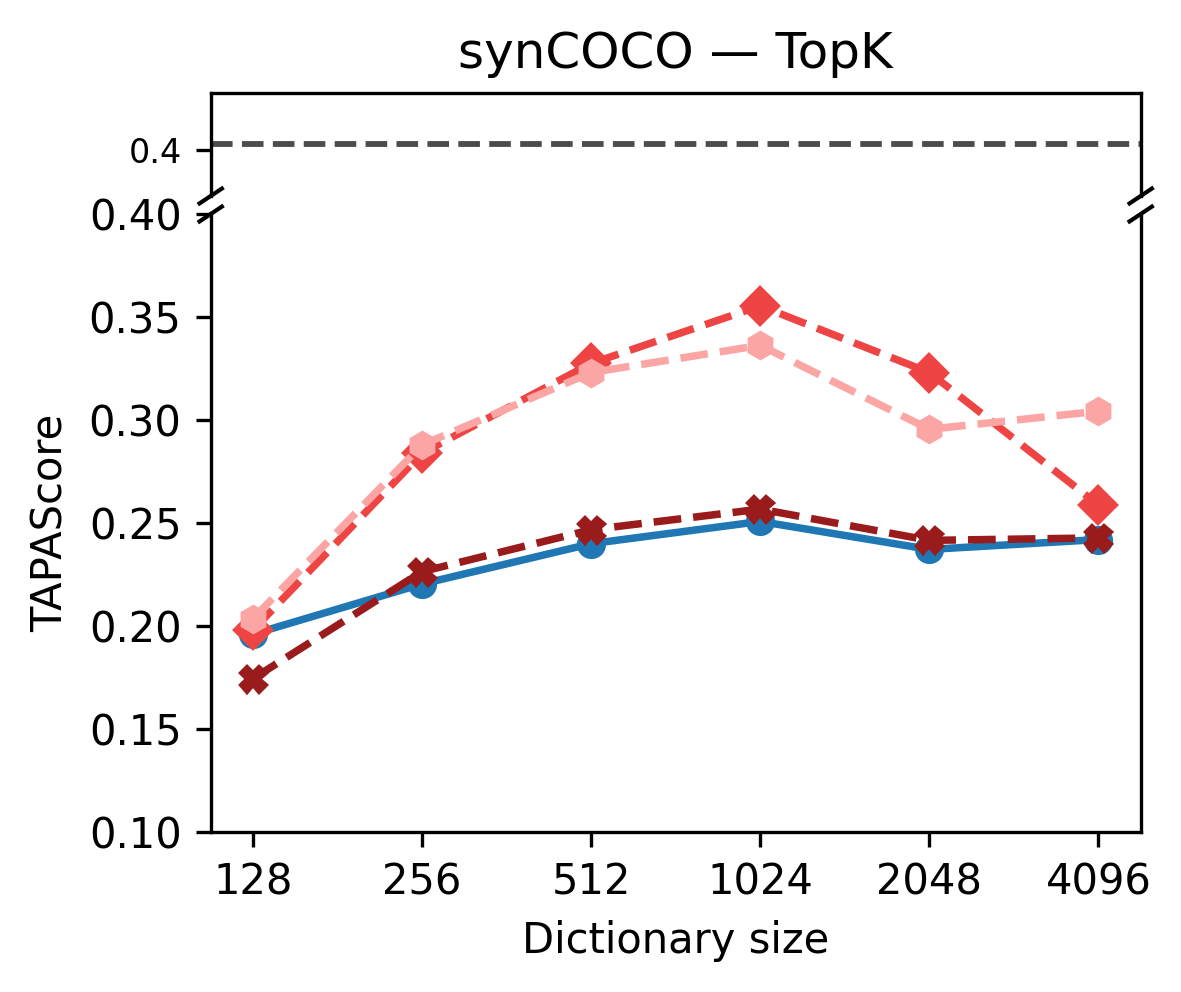}
    \end{subfigure}
    \hfill
    \begin{subfigure}[t]{0.24\textwidth}
      \centering
      \includegraphics[width=\linewidth]{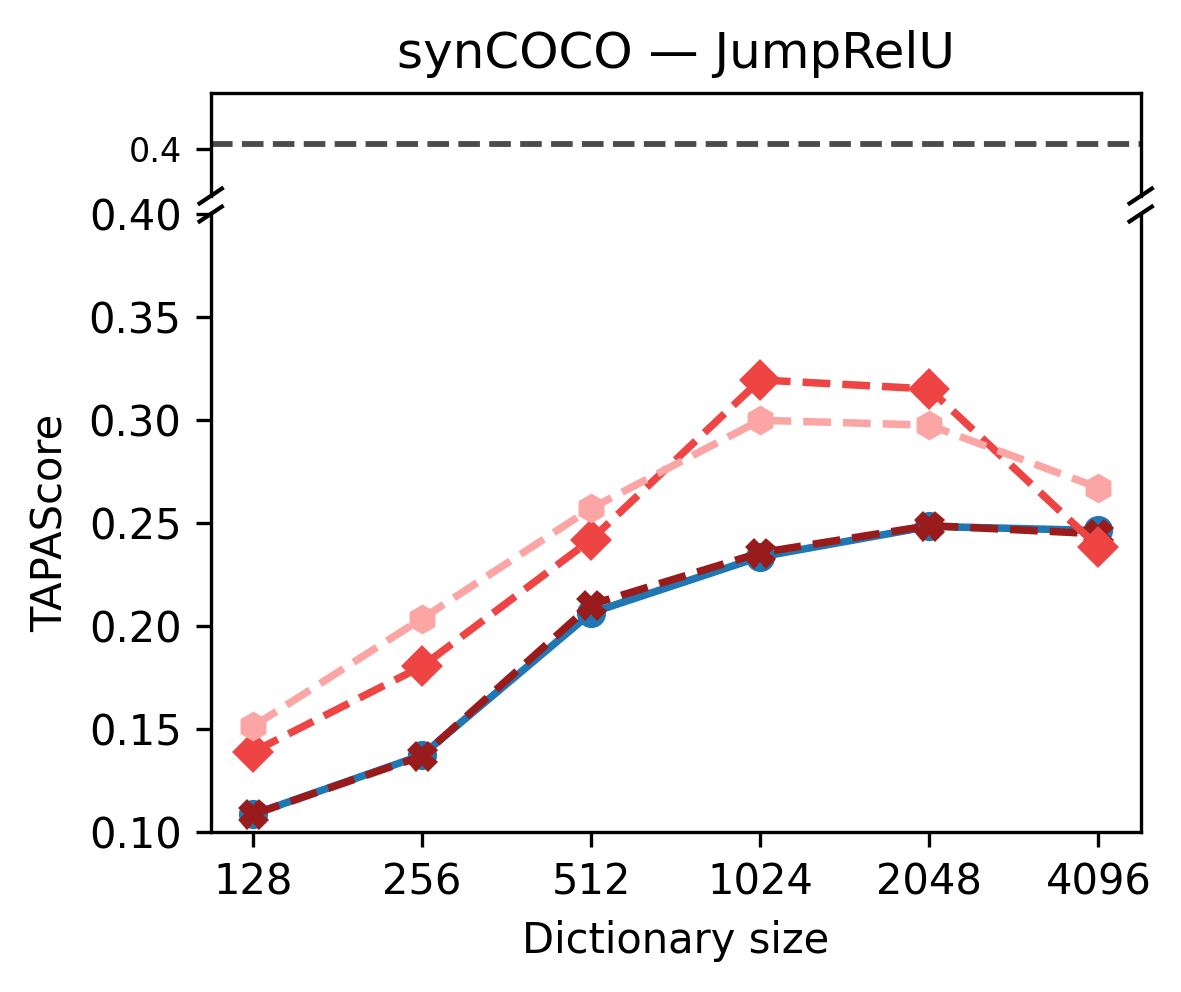}
    \end{subfigure}

    \vspace{-1mm}
  
    \includegraphics[width=\linewidth]{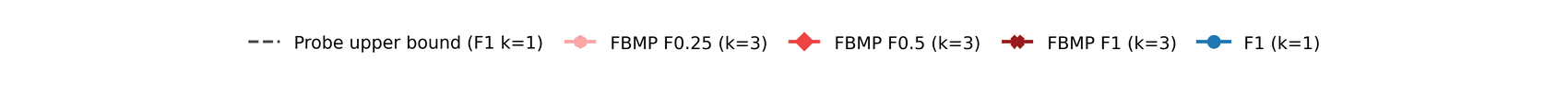}

  \vspace{-1.2em}
  \caption{$F_1$ $\Delta$MATCHScore on COCO (\textit{top row}) and TAPAScore on synCOCO (\textit{bottom row}) as a function of dictionary size for SAEs trained on CLIP embeddings of COCO, across BatchTopK, Matryoshka, TopK and JumpReLU SAE variants (\textit{left to right}) and different matching criteria. The gray dashed line marks the supervised linear-probe upper bound.}
  \label{fig:matching_and_tapas_coco}
    \vspace{-2em}
  \end{figure*}

We evaluate semantic alignment between SAE latents and human-annotated concepts across dictionary sizes and SAE variants for CLIP. The matching scores of untrained SAEs increase with dictionary size, as larger dictionaries offer more latents and thus a greater chance of spurious correlations, a trend we validate in the Appendix (Fig.~\ref{fig:app_matching_untrained}). To enable fair comparison across dictionary sizes, we therefore calculate the $\Delta$MATCHScore (Eq.~\ref{eq:matchscore}) throughout. The raw scores and results for different models are provided in the Appendix (Sec.~\ref{app:matching_results}).

Fig.~\ref{fig:matching_and_tapas_cub} (\textit{top row}) shows the $F_1 \Delta$ matching scores across dictionary sizes for BatchTopK, Matryoshka, TopK, and JumpReLU SAEs on CUB with a CLIP model. For each attribute, we find the best matching latent(s) under a given criterion, either one-to-one ($F_1,k{=}1$) or different FBMP variants with $k{=}3$ (FBMP $F_1$, FBMP $F_{0.5}$, FBMP $F_{0.25}$). The scores are aggregated over all attributes to obtain a mean $F_1$ matching score per dictionary size.
We observe that FBMP consistently outperforms its one-to-one counterpart across all SAE variants, consistent with the motivation that attributes may be represented by multiple complementary latents rather than a single unit. Among the FBMP variants, FBMP $F_{0.5}$ tends to achieve the highest scores. Matching scores are not monotonically increasing with dictionary size: BatchTopK and Matryoshka peak at dictionary sizes 512 and 256, respectively, after which the score decreases. TopK, in contrast, exhibits a more monotonic increase, peaking at 2048, and therefore outperforms the other variants at large dictionary sizes, though a drop is observed at 4096. JumpReLU shows a dip at dictionary size 256 followed by a steady increase up to 4096, while its one-to-one matching slowly decreases with dictionary size. All SAE variants remain well below the linear-probe upper bound, indicating that only part of the attribute information present in the embeddings is recovered as individual latents.

Fig.~\ref{fig:matching_and_tapas_coco} (\textit{top row}) reports the $F_1$ matching scores for BatchTopK, Matryoshka, TopK, and JumpReLU SAEs trained on COCO with CLIP. Similarly to the CUB results, FBMP consistently outperforms the one-to-one baseline across all SAE variants, and FBMP $F_{0.5}$ achieves the highest scores overall. Matching scores are notably higher than on CUB, which we attribute to the smaller and higher-level concept set whose categories are more semantically distinct from one another. Unlike in CUB, BatchTopK and Matryoshka do not exhibit a clear peak followed by a decline; instead, their performance remains stable or continues to improve at larger dictionary sizes, suggesting that the greater visual complexity of COCO requires more latents to fully encode its representations. JumpReLU follows the same pattern, increasing monotonically and saturating around dictionary size 2048, although the gap between FBMP and one-to-one matching is smaller than for the other variants.

\subsection{Functional Validation of Concept Alignment}
\label{sec:tapas_results}
While a high matching score indicates statistical alignment between SAE latents and annotated attributes, correlation alone does not imply causal encoding. To validate that matched latents genuinely encode the corresponding semantic attributes, we compute TAPAScore on our synthetic datasets. The SAEs are trained on CLIP embeddings of CUB and COCO. The results for DINOv2 are provided in the Appendix (Sec.~\ref{app:tapas_results}).
From Fig.~\ref{fig:matching_and_tapas_cub} (\textit{bottom row}), we can observe that for the BatchTopK and Matryoshka SAE, the TAPAScore values closely mirror the matching score, suggesting that statistical and causal alignment are consistent for these variants. In contrast, the TopK SAE exhibits a markedly different pattern: while matching scores increase with dictionary size, TAPAScore peaks early and then degrades sharply. This dissociation suggests that larger TopK dictionaries achieve better statistical alignment without encoding concepts that are causally valid. The JumpReLU SAE shows no such severe degradation: its FBMP TAPAScores fluctuate around a stable level across dictionary sizes, and only the one-to-one criterion declines beyond dictionary size 512.
\begin{wrapfigure}{r}{0.33\linewidth}
    \centering
    \vspace{-20pt}
    \includegraphics[width=\linewidth]{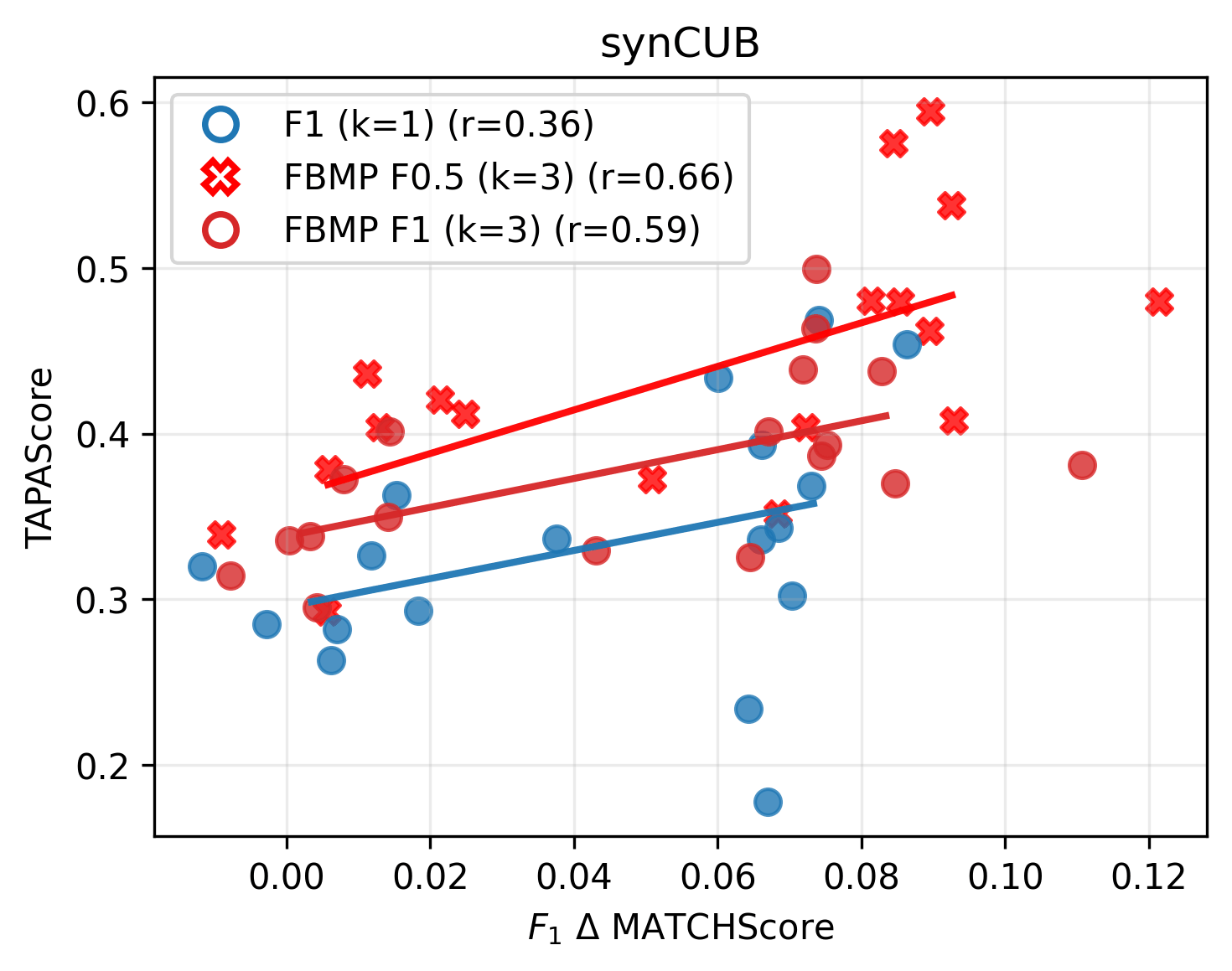}
    \vspace{-0.5em}
    \includegraphics[width=\linewidth]{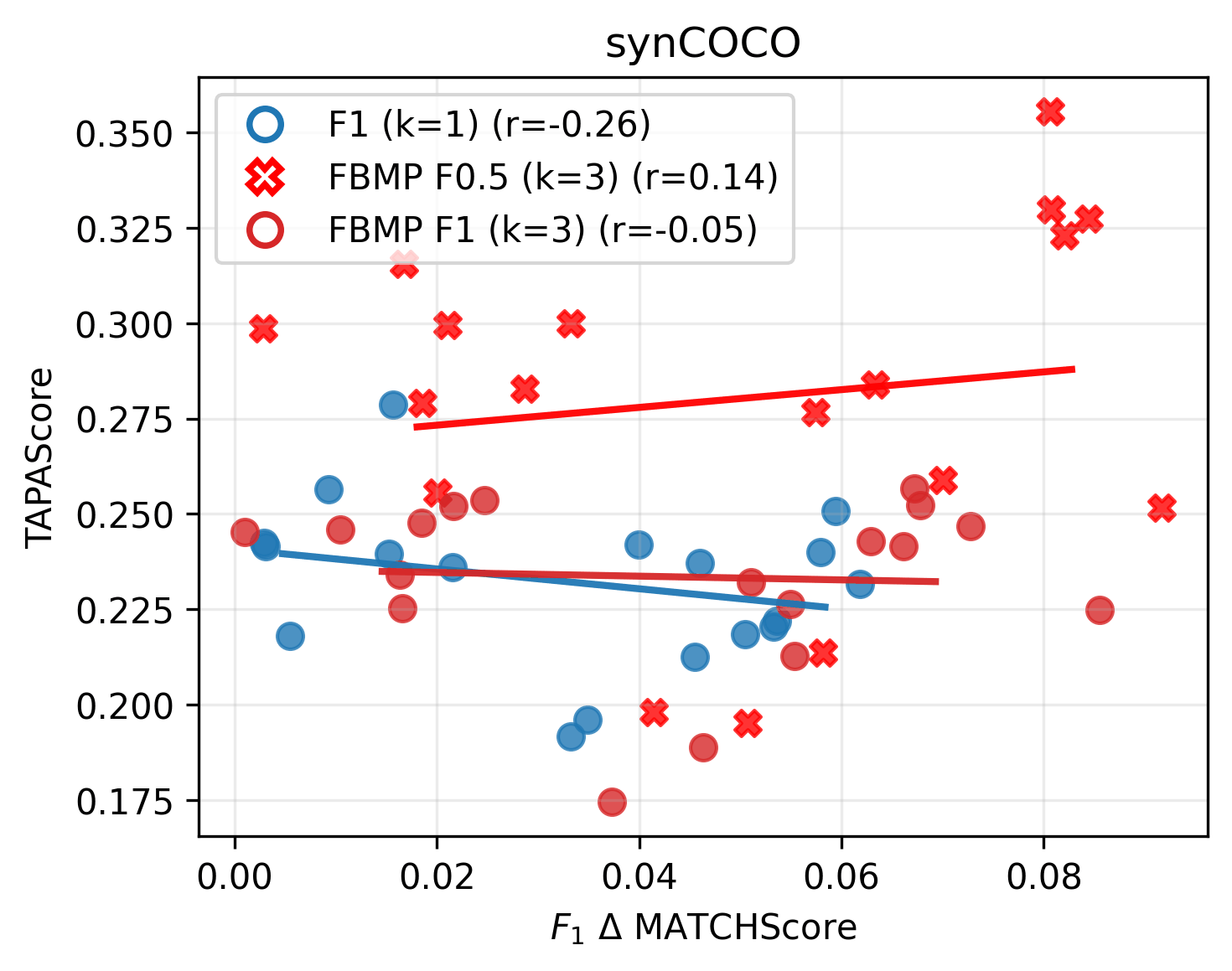}
    \caption{Correlation between matching score and TAPAScore across SAE configurations on synCUB (\textit{top}) and synCOCO (\textit{bottom}).}
        \vspace{-20pt}
    \label{fig:delta_matching_score_vs_tapas}
\end{wrapfigure}

From Fig.~\ref{fig:matching_and_tapas_coco} (\textit{bottom row}), we observe a larger divergence between matching and TAPAScore than in synCUB. $F_1$ $\Delta$MATCHScore grows with dictionary size for all SAE variants, saturating around sizes 512--1024. TAPAScore follows a similar trend up to this point, but then declines for larger dictionary sizes in BatchTopK, TopK, and JumpReLU, suggesting that overcompleteness degrades perturbation alignment despite improved matching quality; for JumpReLU this decline only sets in at dictionary size 4096 and is the least pronounced. Matryoshka behaves differently: FBMP $F_{0.5}$ ($k{=}3$) continues to improve beyond 1024, reaching its maximum at dictionary size 2048 before plateauing, suggesting that the nested structure of Matryoshka SAEs requires larger dictionaries to fully capture the representations. However, Matryoshka scores are overall lower than those of the other SAE types.
In the Appendix (Sec.~\ref{app:delta_stay}) we further verify that TAPAScore is not inflated by leakage: latents matched to untouched attributes remain stable under perturbation.

\noindent \textbf{Correlation analysis.}
To validate the patterns observed in Secs.~\ref{sec:matching_results} and~\ref{sec:tapas_results}, we examine whether latent-concept matching predicts targeted perturbation alignment. We compute the Pearson correlation between matching scores and TAPAScore across all SAE configurations (variant and dictionary size), with each point in Fig.~\ref{fig:delta_matching_score_vs_tapas} corresponding to one configuration.
On synCUB, we observe a consistent positive correlation across all criteria, with FBMP variants outperforming one-to-one matching, and FBMP $F_{0.5}$ achieving the highest correlation. 
On synCOCO, the correlations are weaker overall, reflecting the divergence between matching and TAPAScore at larger dictionary sizes, where matching keeps improving while TAPAScore declines. Here, FBMP variants yield near-zero correlations, while $F_1$ $k{=}1$ yields a negative correlation.

Taken together, the results reveal a consistent picture: statistical and causal alignment largely agree, with higher-matching configurations tending to exhibit stronger perturbation alignment, but the correspondence is not universal. TopK on CUB shows that high matching scores do not guarantee causality, and the COCO results show that overcompleteness can reduce perturbation alignment despite higher matching quality. TAPAScore therefore provides a necessary complement that statistical alignment alone cannot replace, and moderate dictionary sizes achieve the best trade-off between the two. We recommend FBMP $F_{0.5}$ with $k{=}3$ as the default matching criterion, paired with TAPAScore for causal validation.

\section{Conclusion}
We present a human-grounded framework for evaluating SAE interpretability in vision models, operationalizing interpretability as alignment between learned sparse latents and human-annotated semantic concepts. The framework comprises latent-concept matching criteria, including FBMP, a coalition-based formulation robust to common SAE failure modes; two synthetic intervention benchmarks (synCUB and synCOCO), which provide controlled single-attribute perturbations; and TAPAScore, a functional metric that tests whether matched latents respond selectively and in the correct direction under targeted perturbations. We find that increased overcompleteness can reduce perturbation alignment, indicating a loss of interpretability, and that moderate dictionary sizes provide the best trade-off between semantic coverage and functional selectivity.

The framework has a limitation that suggests a direction for future work. The matching quality is directly bounded by annotation quality and granularity, and since TAPAScore relies on matched latent sets, poor annotations lead to unreliable perturbation alignment estimates. Future work could therefore explore approaches that reduce the dependence on manually curated attribute annotations. Beyond evaluation, a natural extension is to use TAPAScore as a steering tool. Instead of measuring whether matched latents respond to attribute perturbations, they could be actively manipulated to control the predictions of an attribute trained classifier, turning the framework into a test for targeted model intervention.
\newpage
%


%
%
\bibliographystyle{splncs04}
\bibliography{main}

\newpage
\appendix
\section*{Supplementary Materials}
\setcounter{section}{0}
\renewcommand\thefigure{S\arabic{figure}}
\renewcommand\thetable{S\arabic{table}}
\renewcommand{\thesection}{S\arabic{section}}
\renewcommand{\theHsection}{appendix.\arabic{section}}

\begin{itemize}
    \item \ref{sup:impact}: \textbf{Discussions and Broader Impact}
    \item \ref{sup:matching_metrics}: \textbf{Latent–Concept Matching Details}
    \item \ref{sup:dataset}: \textbf{Synthetic Dataset Construction Details}
    \item \ref{app:subsec:sae_training}: \textbf{Sparse Autoencoder Training Configuration}
    \item \ref{app:qualitative_matching_results}: \textbf{Qualitative Matching Results}
    \item \ref{app:sec:additional_results}: \textbf{Additional Results}

\end{itemize}
\section{Discussions and Broader Impact}
\label{sup:impact}

\textbf{Limitations and Future Work.}
The primary limitation of our framework is its dependence on the availability, quality and granularity of human-annotated concepts. Since latent-concept matching relies on binary attribute annotations, the matching quality is directly bounded by annotation completeness. Concepts absent from the annotation set cannot be matched, and noisy or coarse annotations may yield unreliable matchings. Because TAPAScore builds on matched latent sets, this limitation propagates to the functional evaluation stage. A natural direction for future work is therefore to reduce reliance on manually curated annotations, for instance by leveraging vision-language models to automatically generate attribute vocabularies.

Beyond evaluation, our framework opens a natural extension toward active model intervention. Rather than measuring whether matched latents respond to attribute perturbations, one could actively steer those latents to control downstream predictions, turning TAPAScore into a diagnostic for targeted concept manipulation. An interesting open question concerns the nature of unmatched latents; we hypothesize that some of these may capture low-level or background concepts that, while not directly interpretable in terms of human-annotated attributes, nonetheless play an important role in shaping the model's internal representation. Exploring this direction in the context of fine-grained recognition and vision-language alignment represents a promising avenue for future work.

\noindent\textbf{Broader Impact.}
Our work contributes tools for auditing the internal representations of large vision models. By grounding interpretability evaluation in human-annotated concepts and intervention-style benchmarks, the framework supports more rigorous transparency and accountability in the deployment of such models. In particular, the ability to detect whether SAE features align with semantically meaningful concepts can help identify unintended or spurious representations that might otherwise go undetected. We do not foresee direct negative societal impacts from this work; however, as with any interpretability method, results should be interpreted with care, as apparent alignment with human concepts does not guarantee that a model is free from bias or other failure modes outside the scope of the evaluated attributes.
\section{Latent–Concept Matching Details}
\label{sup:matching_metrics}
This section provides full details of the latent--concept matching procedure introduced in Section 3.1 of the main manuscript. We describe the binary matching metrics used to quantify alignment between SAE latents and human-annotated concepts, and outline both the one-to-one and many-to-one matching schemes evaluated in our experiments.
\subsection{One-to-one Matching}
\label{sup:metrics}
In the one-to-one matching scheme, each attribute $a$ is assigned to exactly one SAE latent $l$ that best captures its activation pattern. To identify this match, we compute standard binary classification metrics between the binarized activation vector $\mathbf{z}^l_{\text{bin}}$ and the ground-truth annotation vector $\mathbf{y}_a$, and select the latent maximizing the F1 score. Formally, for each attribute $a$ and SAE latent $l$, we compute precision $P_{al}$ and recall $R_{al}$ as:
\begin{equation}
P_{al} = \frac{TP_{al}}{TP_{al} + FP_{al} + \varepsilon},
\qquad
R_{al} = \frac{TP_{al}}{TP_{al} + FN_{al} + \varepsilon},
\end{equation}
and define the $F1_{al}$ score as:
\begin{equation}
F1_{al} = \frac{2 P_{al} R_{al}}{P_{al} + R_{al} + \varepsilon}.
\end{equation}

\begin{figure}
    \centering
    \vspace{-2em}
    \includegraphics[width=\linewidth]{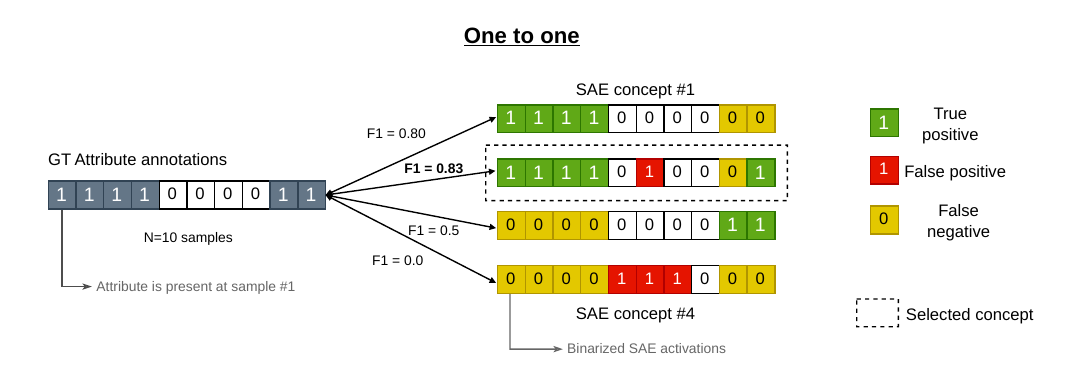}
    \vspace{-2em}
    \caption{One-to-one matching procedure using an F1 similarity criterion.}
    \label{fig:one-to-one}
    \vspace{-2em}
\end{figure}
\begin{figure}
    \centering
    \vspace{-1.5em}
    \includegraphics[width=0.8\linewidth]{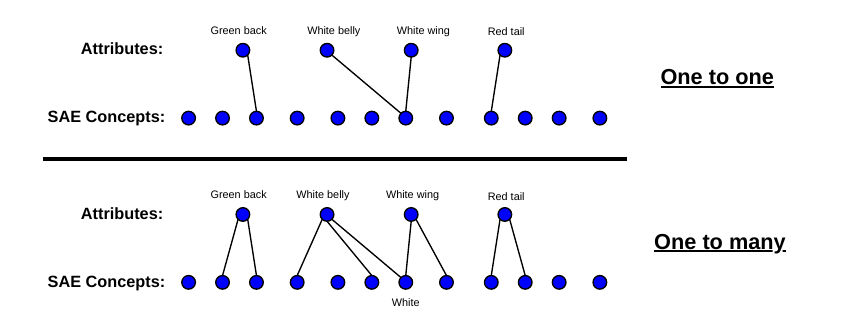}
    \vspace{-2em}
    \caption{Graphical representation of the two discussed matching paradigms. Since matching iterates over attributes independently, a latent can be selected by multiple attributes simultaneously.}
    \label{fig:matching_abstract}
    \vspace{-2em}
\end{figure}

The F1-based matching procedure is summarized in Fig.~\ref{fig:one-to-one}.
Although each attribute $a$ is matched to a single latent, nothing prevents the same latent $l$ from being selected for multiple attributes (see Fig.~\ref{fig:matching_abstract}). Such a case would suggest a polysemantic nature of said latent.
While the same phenomenon may arise under a many-to-one matching, a different interpretation can be drawn in this more complex setting. Because an attribute can be decomposed into multiple latents, it now accommodates feature composition~\cite{wattenberg2024relational}, where co-occurring semantic concepts (i.e., commonalities between attributes) are captured by a common latent. For example, attributes such as “yellow head”, “yellow crown” and “yellow throat” may share a latent representing the concept “yellow”, while their attribute-specific semantics can be expressed through composition with additional, distinct latents.
By construction, such compositional structure cannot be recovered under a strict one-to-one assignment.

\subsection{Fully-Binary Matching Pursuit (FBMP)}
\label{sup:fbmp}

\begin{figure}
    \centering
    \vspace{-2em}
    \includegraphics[width=\linewidth]{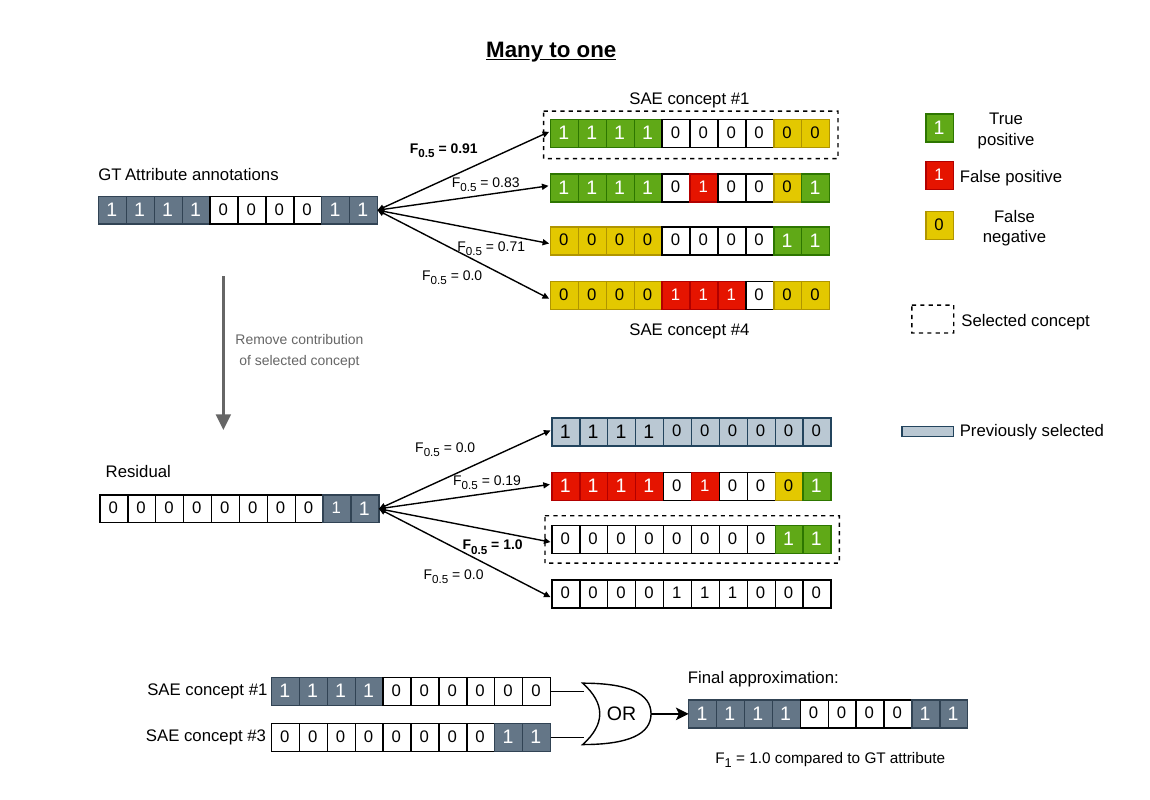}
    \vspace{-2em}
    \caption{Many-to-one matching approach on a sequential residual update procedure.}
    \label{fig:many-to-one}
    \vspace{-2em}
\end{figure}

The many-to-one matching procedure performed by FBMP is summarized in Fig.~\ref{fig:many-to-one}. It consists of a sequential selection procedure where one latent is selected at each step (based on some similarity criterion), then a residual is computed by removing the latent's contribution. The following iteration then proceeds on the residual of the previous step.
In Fig.~\ref{fig:many-to-one}, a $F_{0.5}$ score is used as the selection criterion. Notice that it selects a different latent from the one selected by the F1 score in Fig.~\ref{fig:one-to-one}. This allows FBMP, in this particular case, to achieve a perfect reconstruction of the attribute annotations after the selection of a second latent in the coalition.

\begin{figure*}
    \centering
    \vspace{-2em}
    \begin{subfigure}[t]{0.49\textwidth}
        \centering
        \includegraphics[width=\linewidth]{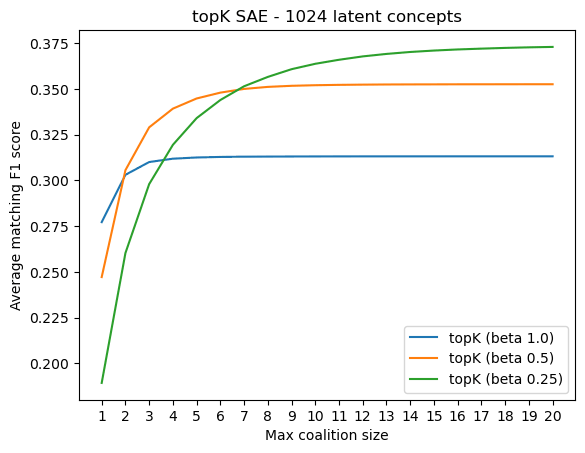}
        \caption{Comparison of $\beta$ values \label{fig:bmp_betas}}
    \end{subfigure}
    \hfill
    \begin{subfigure}[t]{0.49\textwidth}
        \centering
        \includegraphics[width=\linewidth]{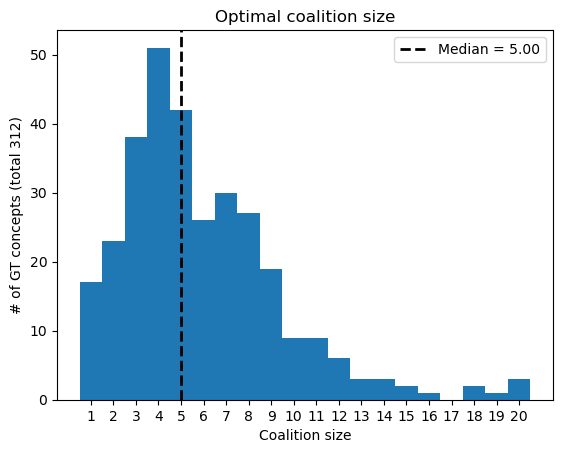}
        \caption{Histogram of optimal coalition sizes \label{fig:bmp_histogram}}
    \end{subfigure}
    \hfill

    \caption{Fully-Binary Matching Pursuit results over a TopK SAE trained on CLIP embeddings of CUB data samples.}
    \label{fig:matching_sae_variants}
    \vspace{-3em}
\end{figure*}

Figure~\ref{fig:bmp_betas} illustrates the effect of the parameter $\beta$ on the $F_\beta$-score used for the latent selection step. The matching F1-score averaged over all 312 CUB attributes is displayed for different maximum coalition sizes ($k$). Lower $\beta$ values privilege precision over recall of the selected latent compared to the attribute presence annotations. As a result, bigger coalitions are required to attain a given F1 matching score, but higher overall scores tend to be achieved for sufficiently large coalition sizes. On the contrary, $\beta=1$ leads to a very quick start (quite high matching after the first selection), but tends to stall earlier. As a good compromise between a higher final matching score and smaller coalition sizes, we select $\beta=0.5$ as the default value for FBMP.

Due to its early-stopping criterion, the sequential procedure of FBMP can terminate earlier than the maximum number of iterations defined by the parameter $k$. In Figure~\ref{fig:bmp_histogram} we show a histogram of the actual coalition sizes obtained over the 312 CUB attributes when running FBMP with a high value of $k=20$. It shows that most of the attributes were matched to a small number of SAE latents, with $50\%$ of them using up to 5 latents only. Very few attributes required a coalition of more than 10 latents. Indeed, for such high coalition sizes, interpretability starts to be compromised and we can even infer that such attributes have not been appropriately learned by the underlying SAE.

\section{Synthetic Dataset Construction Details}
\label{sup:dataset}
This section provides full details of the synthetic dataset construction pipeline introduced in Section 3.3, covering the generation procedure, prompt design, and curation steps for both synCUB and synCOCO
.

\begin{table}[t]
\centering
\setlength{\tabcolsep}{6pt}
\caption{Statistics of the synthetic perturbation datasets used for TAPAS evaluation. The verification classifier flagged 54\% of synCUB and 72\% of synCOCO pairs for manual review (Sec.~\ref{sup:dataset_curation}).}
\label{tab:synthetic_dataset_statistics}
\begin{tabular}{lcccc}
\hline
Dataset &
\shortstack{Generated\\pairs} &
\shortstack{Attributes\\manipulated} &
\shortstack{Avg. pairs\\per attribute} &
\shortstack{Final curated\\pairs} \\
\hline
COCO & 9000 & 80 & 32.08 & 2534 \\
CUB  & 3063 & 43 & 136.42 & 2933 \\
\hline
\end{tabular}
\end{table}
\begin{figure*}[t]
    \centering
    \vspace{-1em}
    \begin{subfigure}[t]{0.9\textwidth}
        \centering
        \includegraphics[width=\linewidth]{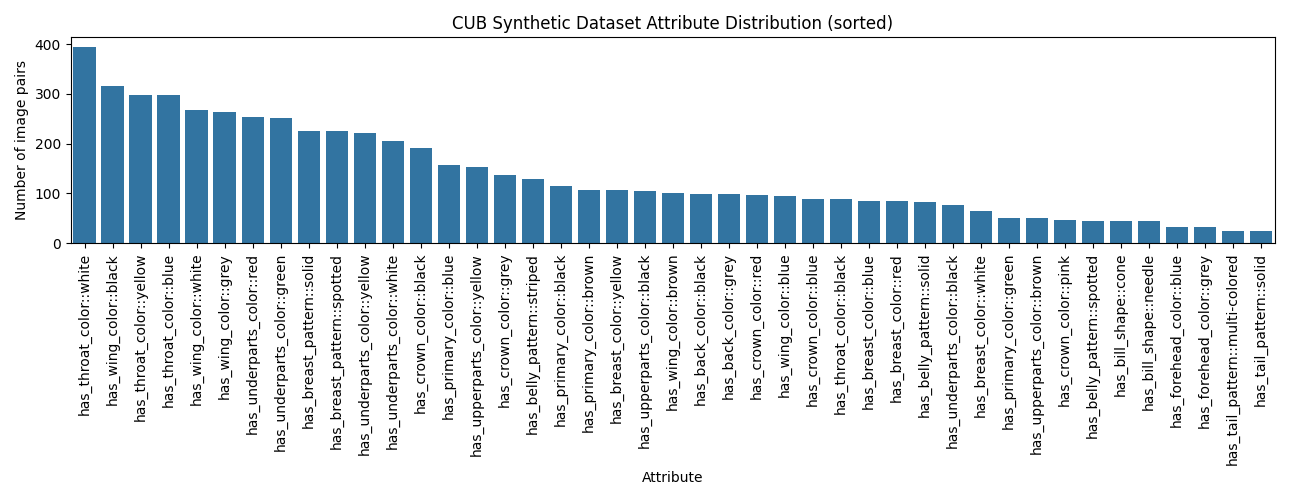}
    \end{subfigure}

    \vspace{0.5em}

    \begin{subfigure}[t]{0.9\textwidth}
        \centering
        \includegraphics[width=\linewidth]{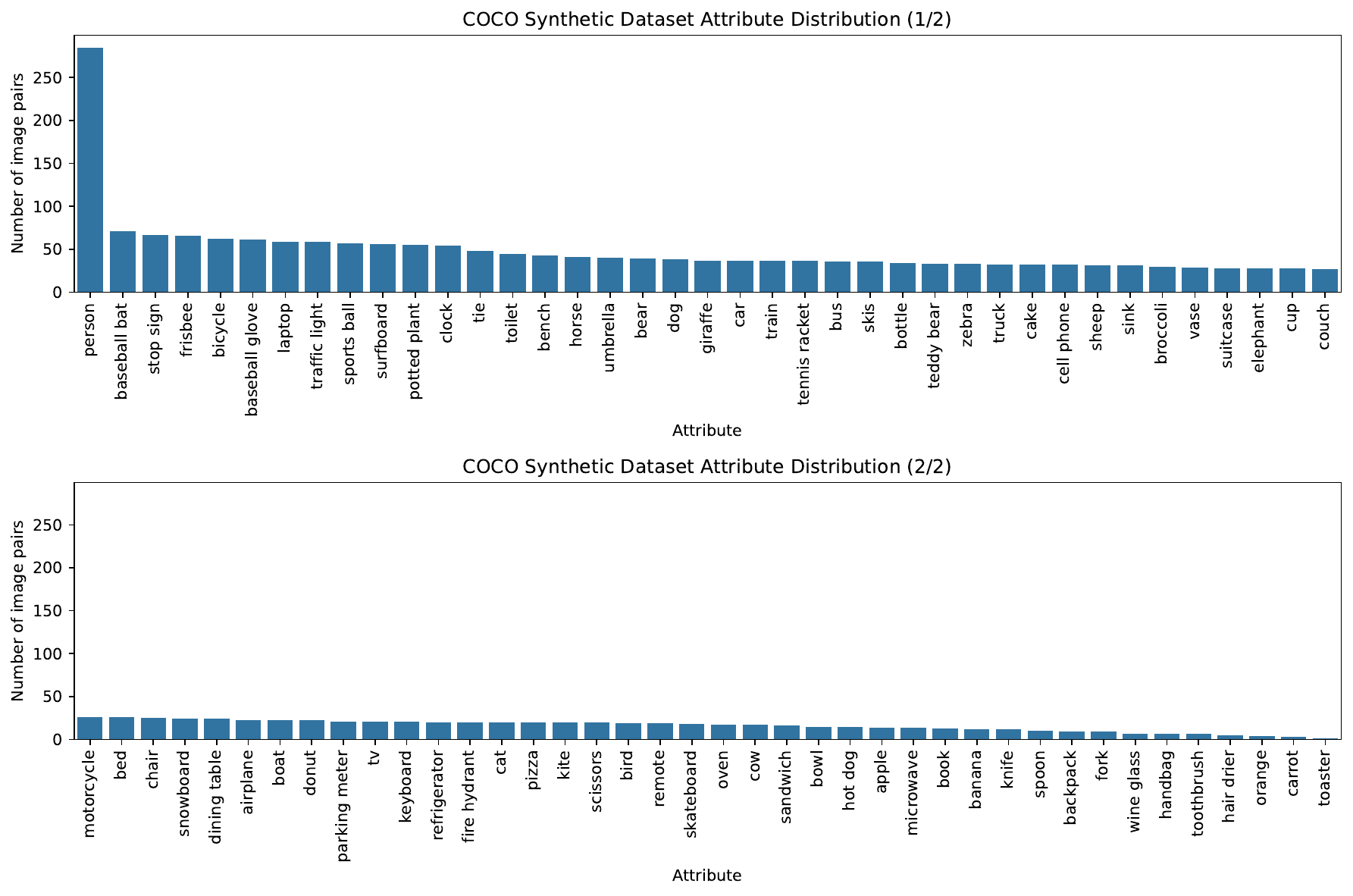}
    \end{subfigure}

    \caption{Attribute histograms for the synthetic datasets. Top: synCUB, bottom: synCOCO.}
    \label{fig:attribute_histograms}
    \vspace{-2em}
\end{figure*}
\subsection{synCUB}
\label{app:subsec:syncub}
synCUB is constructed by applying reference-guided attribute perturbations to images from a controlled subset of CUB-200-2011~\cite{CUB_dataset}. We restrict ourselves to the 33-class subset used in SUB~\cite{bader2025sub} and consider their curated set of 45 attribute concepts. CUB attributes are provided with per-instance confidence scores; for each part family (e.g., \textit{has leg color}, or \textit{has bill shape}), we select the three most frequent attributes that also exhibit the highest annotation confidence. For each selected target attribute, we identify reference images that guide the appearance of the target attribute during editing. Reference candidates are drawn from the three most attribute-frequent classes within the CUB subset that contain at least three high-confidence examples of the attribute.

Table~\ref{tab:synthetic_dataset_statistics} summarizes the resulting dataset statistics. In total, we generated 3063 edited image pairs across 43 manipulated attributes, of which 2933 pairs remained after the automatic and manual filtering stages.
Figure~\ref{fig:attribute_histograms} (top) shows the distribution of manipulated attributes. The distribution is relatively balanced across attributes, with an average of 136.42 pairs per attribute. While some attributes occur more frequently than others, the overall distribution remains well spread across the attribute space, indicating that the generation and filtering pipeline preserves a diverse set of semantic perturbations.
\begin{figure}
    \centering
    \vspace{-2em}
    \includegraphics[width=1\linewidth]{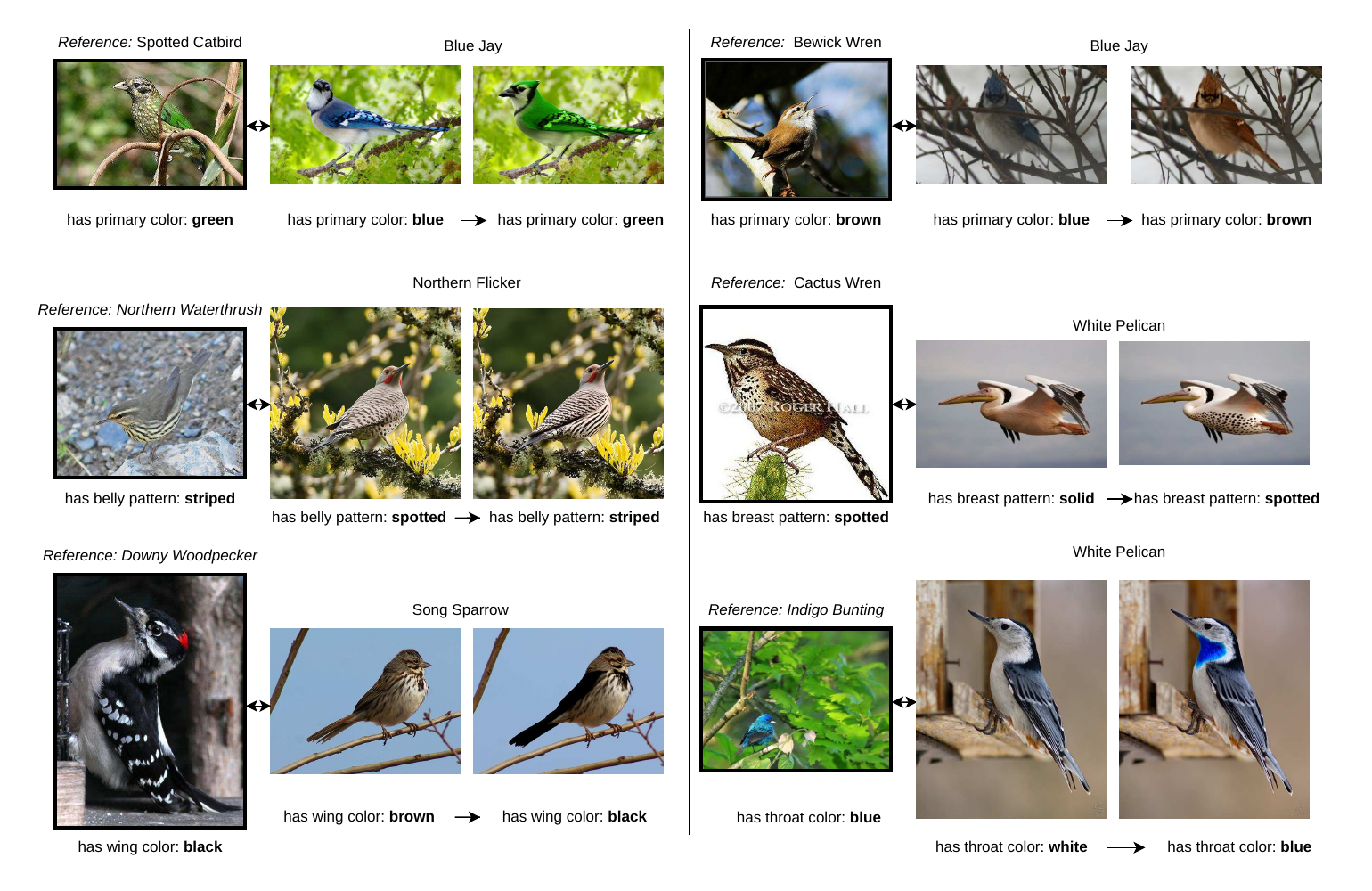}
    \caption{synCUB pairs: Additional example images.}
    \label{fig:app_syncub_qualitative1}
    \vspace{-2em}
\end{figure}

\subsection{synCOCO}
\label{app:subsec:syncoco}

While synCUB evaluates attribute-level alignment in a controlled single-object setting, real-world scenes are considerably more complex. Editing in such complex scenes is considerably more difficult. In particular, adding objects would frequently lead to unrealistic scenes. For example, removing a table does not imply that a car can plausibly be inserted into the same scene, which would result in out-of-distribution images. We therefore restrict synCOCO to object removal operations. Even object removal remains challenging due to the compositional nature of COCO scenes.
For each pair, we select the target object to remove by first choosing the object with the lowest instance count in the scene. Our experiments showed that removing multiple objects simultaneously leads to unstable edits and unreliable verification. If multiple objects have the same instance count, ties are broken by selecting the object with the largest area in the image, ensuring that the perturbation remains visually significant.

Table~\ref{tab:synthetic_dataset_statistics} summarizes the resulting dataset statistics. We generated 9000 candidate image pairs covering all 80 COCO object categories. After automatic filtering and manual verification, 2534 pairs remained in the final curated dataset, covering 79 of the 80 categories; all pairs of one category (\textit{mouse}) were rejected during curation.
Figure~\ref{fig:attribute_histograms} (bottom) shows the resulting concept distribution. Compared to synCUB, the distribution is less balanced, with \textit{person} remaining the most frequent category, as such objects are easier both for the editing model to remove and for the verification classifier to detect reliably. During the manual review of classifier-flagged pairs we therefore prioritized rare object categories, which improves the balance of the final dataset compared to a purely classifier-based selection.

\begin{figure}
    \centering
    \vspace{-1em}
    \includegraphics[width=1\linewidth]{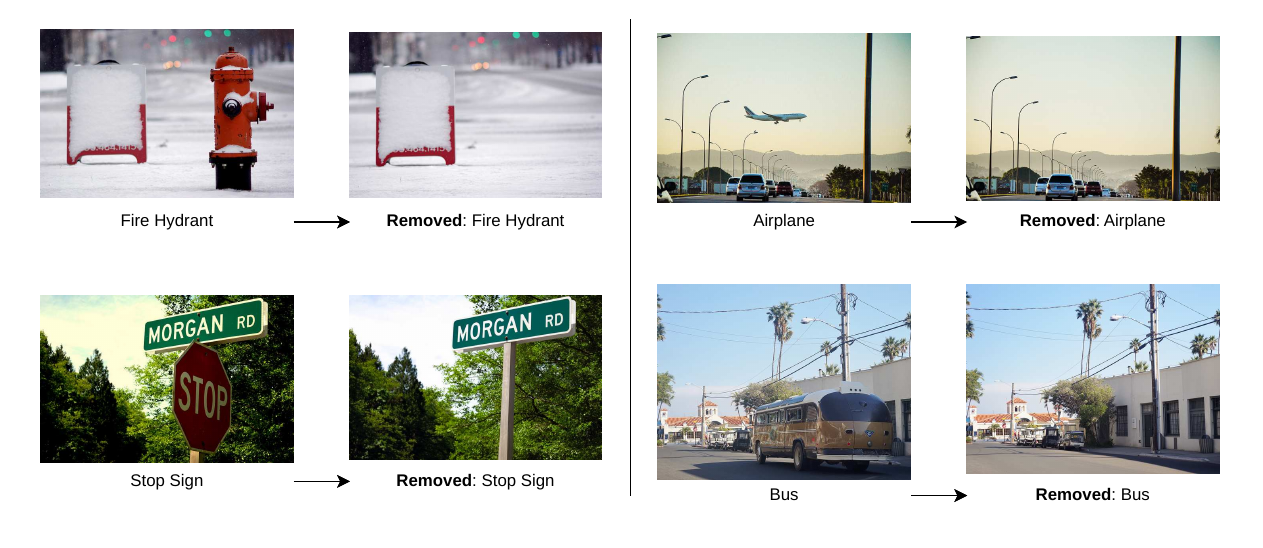}
    \caption{synCOCO pairs: Additional example images.}
    \label{fig:app_syncoco_qualitative1}
    \vspace{-2em}
\end{figure}

\subsection{Prompts}
\label{sup:dataset_prompts}
\begin{figure}
    \centering
    \vspace{-1em}
    \includegraphics[width=0.8\linewidth]{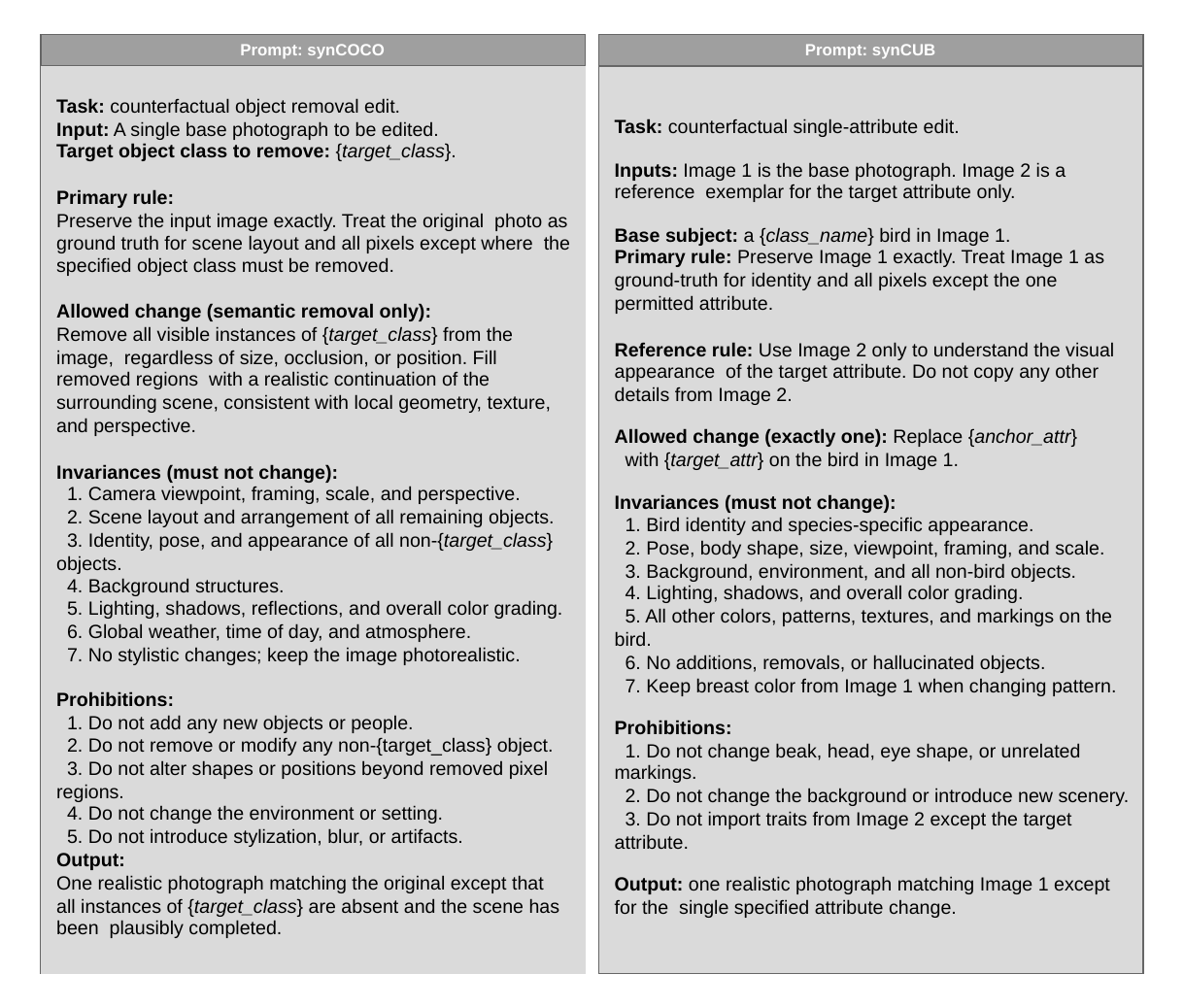}
    \caption{Prompts for the image generation}
    \label{fig:app:prompts}
    \vspace{-2em}
\end{figure}

All attribute perturbations are generated using structured natural language prompts designed to enforce a strict single factor intervention constraint. The prompts are instantiated programmatically to ensure consistency across attributes and object classes.
For synCUB, we use reference-based editing. Given a base image and a reference image representing the target attribute state, the edited image is generated using Flux2~\cite{flux-2-2025}, which allows conditioning on multiple input images. The base image serves as the identity reference, preserving pose, background, and all other attributes, while the attribute-reference image guides the appearance of the target attribute to be added. The prompt explicitly instructs the model to modify only the specified attribute while preserving identity, pose, and background.
For synCOCO, no reference images are used. Since the edits consist solely of removing an object rather than inserting a new one, the model is conditioned only on the original image together with a prompt that instructs the model to remove the specified object while preserving the rest of the scene.
The prompt templates below are used verbatim, with placeholders replaced by the corresponding attribute names, class names, and semantic families. The exact prompts are visualized in Fig.~\ref{fig:app:prompts}.

\subsection{Dataset Curation}

\label{sup:dataset_curation}
\subsubsection{Classifier-based selection of samples}
\label{app:subsubsec:classifier_selection}
To ensure label consistency in the generated synthetic datasets, we trained multi-label ResNet-50 classifiers separately on the original CUB (attribute annotations) and COCO datasets.
Training was performed for 30 epochs on CUB and 10 epochs on COCO. The classifier was used as a filtering mechanism during dataset construction. For each original–synthetic image pair, we verified whether the classifier predictions were consistent with the expected attribute perturbations. If the classifier misclassified either the original or the synthetic image with respect to the manipulated label indices, the pair was manually inspected to ensure correctness of the perturbation and annotation.

\begin{figure*}[t]
    \centering
    \vspace{-1em}
    \begin{subfigure}[t]{0.49\textwidth}
        \centering
        \includegraphics[width=\linewidth]{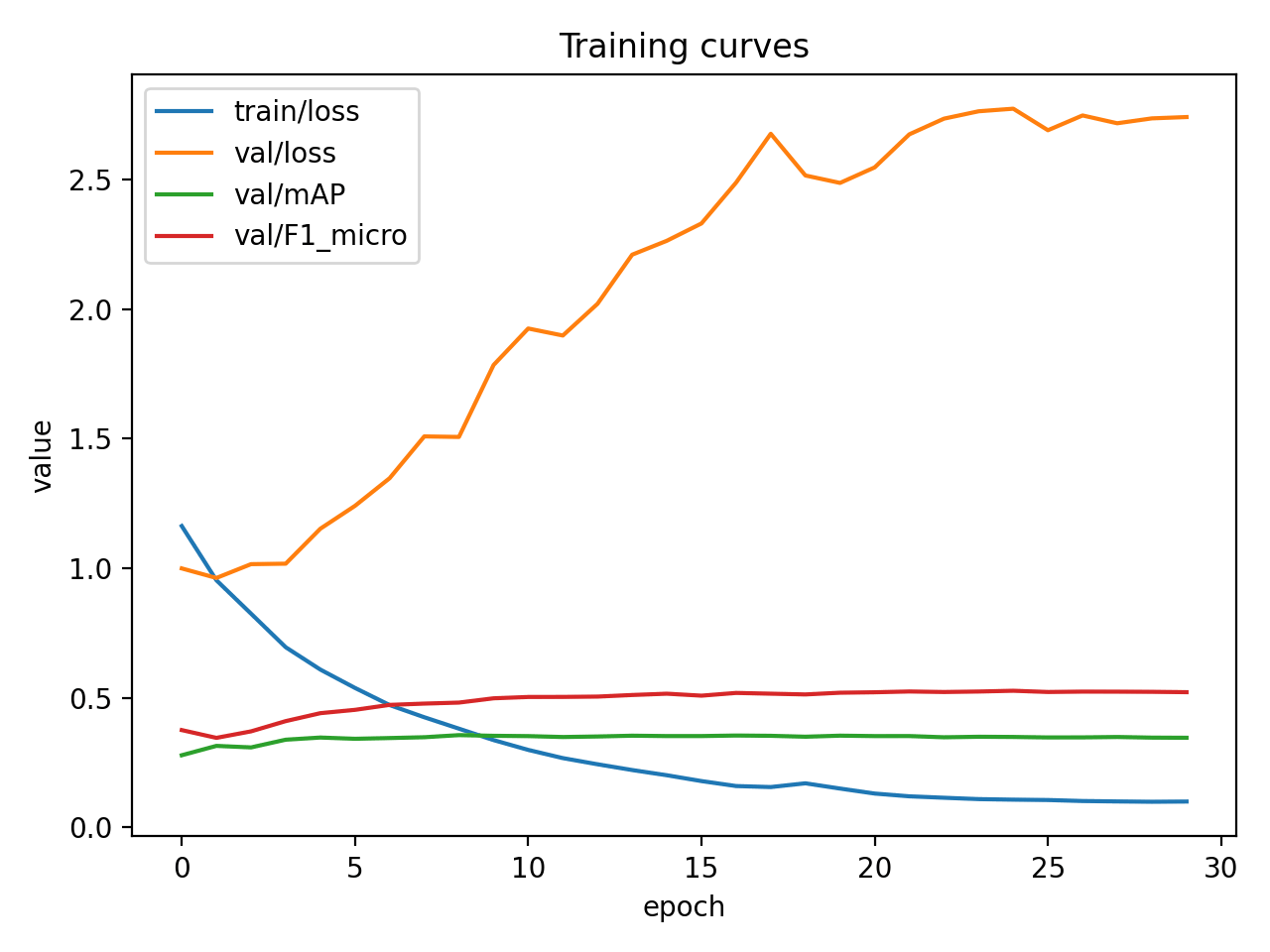}
        \caption{Synthetic CUB}
    \end{subfigure}
    \hfill
    \begin{subfigure}[t]{0.49\textwidth}
        \centering
        \includegraphics[width=\linewidth]{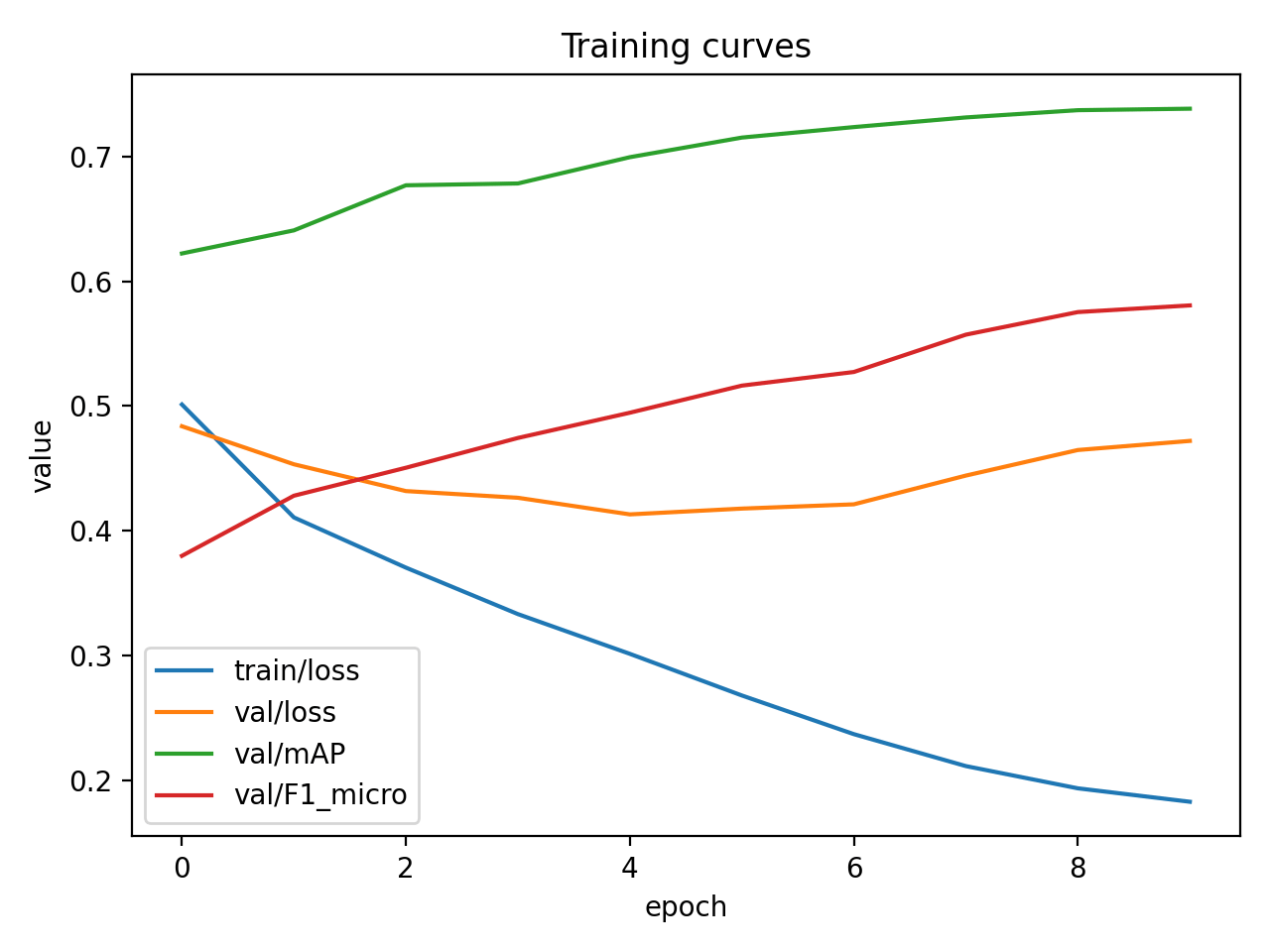}
        \caption{Synthetic COCO}
    \end{subfigure}
    \caption{Training curves of the ResNet-50 classifiers used for dataset filtering.}
    \label{fig:dataset_gen_classifier}
    \vspace{-2em}
\end{figure*}

Table~\ref{tab:app_class_results} reports the final validation performance of the classifiers on original and synthetic data. For both datasets, performance on synthetic images is lower than on original images, reflecting the distribution shift introduced by controlled attribute manipulation. The gap is more pronounced for CUB in terms of macro F1, consistent with its fine-grained attribute structure. The training dynamics are shown in Fig.~\ref{fig:dataset_gen_classifier}. The curves illustrate stable convergence on the original data, with decreasing training loss and increasing validation metrics. 

\begin{table}[t]
\centering
\caption{Validation performance of the ResNet-50 classifiers used for sample filtering.}
\label{tab:app_class_results}
\begin{tabular}{l cc cc}
\toprule
 & \multicolumn{2}{c}{Original} & \multicolumn{2}{c}{Synthetic} \\
\cmidrule(lr){2-3} \cmidrule(lr){4-5}
Dataset & F1 (mi) & F1 (ma) & F1 (mi) & F1 (ma) \\
\midrule
Synthetic COCO & 0.63 & 0.62 & 0.42 & 0.37 \\
Synthetic CUB  & 0.58 & 0.42 & 0.42 & 0.28 \\
\bottomrule
\end{tabular}
\end{table}

\subsubsection{Human validation of samples}

\begin{wrapfigure}{r}{0.5\linewidth}
    \centering
    \vspace{-2em}
    \includegraphics[width=\linewidth]{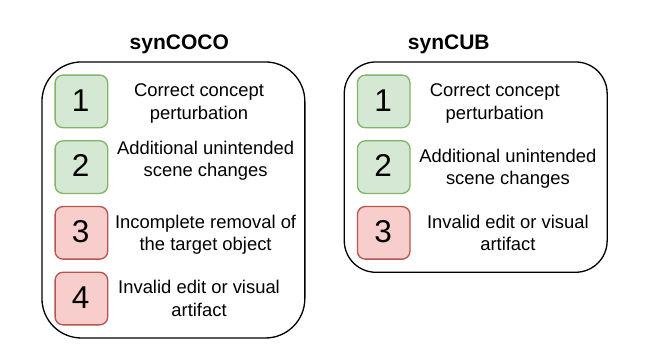}
    \caption{Human validation categories used during dataset curation. Green categories denote edits that satisfy the concept perturbation constraint and are retained, while red categories correspond to generation failures that are filtered from the final dataset.}
    \label{fig:app:selection}
    \vspace{-2em}
\end{wrapfigure}

Image pairs flagged by the filtering classifier described in Section~\ref{app:subsubsec:classifier_selection} were manually curated. Annotation was carried out by four of the authors; each image pair was assigned to one of the categories shown in Fig.~\ref{fig:app:selection}.
The categories describe whether the intended concept perturbation was successfully applied and whether additional unintended changes occurred. For both datasets, the first two categories correspond to valid edits and are retained, while the remaining categories correspond to generation failures and are removed from the dataset.
For synCUB, images were annotated as either a correct concept perturbation or as containing additional unintended attribute changes. Edits that produced visual artifacts or otherwise invalid modifications were categorized as invalid edits and filtered out. Flagged synCUB pairs were reviewed in multiple rounds by different annotators, and the per-pair scores were averaged; 130 pairs were removed in this way.
For synCOCO, images were annotated using the same criteria with one annotator per image pair, and with an additional category capturing incomplete removal of the target object. Here, we extended the manual validation to the entire dataset: all 2504 pairs that passed the classifier check were also manually reviewed, of which 903 (36.1\%) were rejected, predominantly due to incomplete removal of the target object. In addition, 1488 classifier-flagged pairs were reviewed, prioritizing rare object categories, of which 933 (62.7\%) were accepted. The final synCOCO dataset therefore contains exclusively human-verified pairs.

\noindent\textbf{Validity of the classifier flag.}
To verify that the automatic flagging carries signal, we performed a blind re-curation experiment: for each dataset, 150 flagged and 150 unflagged pairs were sampled, mixed, and re-curated without knowledge of their flag status. Rejection rates were consistently higher among flagged pairs (synCUB: 20.7\% vs.\ 10.7\%, Fisher's exact test $p=0.025$; synCOCO: 41.3\% vs.\ 32.7\%, $p=0.15$), confirming that the flag is informative. For synCOCO, any residual dependence on the classifier is eliminated by the subsequent manual validation of every retained pair; for synCUB, the rejection rate among unflagged pairs bounds the residual error rate at roughly 10\%.

\section{Sparse Autoencoder Training Configuration}
\label{app:subsec:sae_training}

We train Sparse Autoencoders (SAEs) on frozen feature representations extracted from pretrained vision backbones. Specifically, we use the normalized class-token representation from the final transformer layer. Experiments are conducted on features from CLIP-ViT-L/14 and DINOv2-ViT-S/14. Training data is obtained from the full training splits of the CUB and COCO datasets.
For each configuration, SAEs are trained for $50$ epochs with dictionary sizes $\{128,$ $ 256,$ $ 512,$ $ 1024,$ $2048,$ $ 4096\}$. We evaluate several SAE training variants: TopK, BatchTopK, GlobalBatchTopKMatryoshkaSAE (referred to as Matryoshka), and JumpReLU~\cite{rajamanoharan2024jumping}. As reference baselines, we additionally report results for randomly activated dictionaries and untrained (frozen) autoencoders. For the TopK-based variants, the sparsity constraint is fixed to $K=32$ active features per input.
Training follows the implementation and hyperparameters of \cite{gao2024scaling}. Optimization is performed using Adam with learning rate $5\times10^{-4}$, $\epsilon=6.25\times10^{-10}$, and $(\beta_1,\beta_2)=(0.9,0.999)$. Gradients are clipped during training and early stopping is applied based on the validation loss.

\noindent\textbf{Training Losses.}
For the TopK and BatchTopK SAEs, the training objective consists of a reconstruction loss, an $\ell_1$ sparsity penalty on the activations, and an auxiliary loss following \cite{gao2024scaling}. Given input features $x$ and reconstruction $\hat{x}$, the reconstruction loss and sparsity penalty are defined as:
\begin{equation}
\mathcal{L}_{\text{rec}} = \|x - \hat{x}\|_2^2 , 
\qquad
\mathcal{L}_{\ell_1} = \lambda \|a\|_1 ,
\end{equation}
where $a$ denotes the top-$k$ activations and $\lambda$ controls the sparsity strength. The overall training objective is
\begin{equation}
\mathcal{L} = \mathcal{L}_{\text{rec}} + \mathcal{L}_{\ell_1} + \mathcal{L}_{\text{aux}} .
\end{equation}
The auxiliary loss $\mathcal{L}_{\text{aux}}$ is computed using inactive (dead) latents as proposed by \cite{gao2024scaling}, encouraging them to model the residual reconstruction error.
\begin{wrapfigure}{r}{0.7\linewidth}
    \centering
    \vspace{-2em}
    \includegraphics[width=\linewidth]{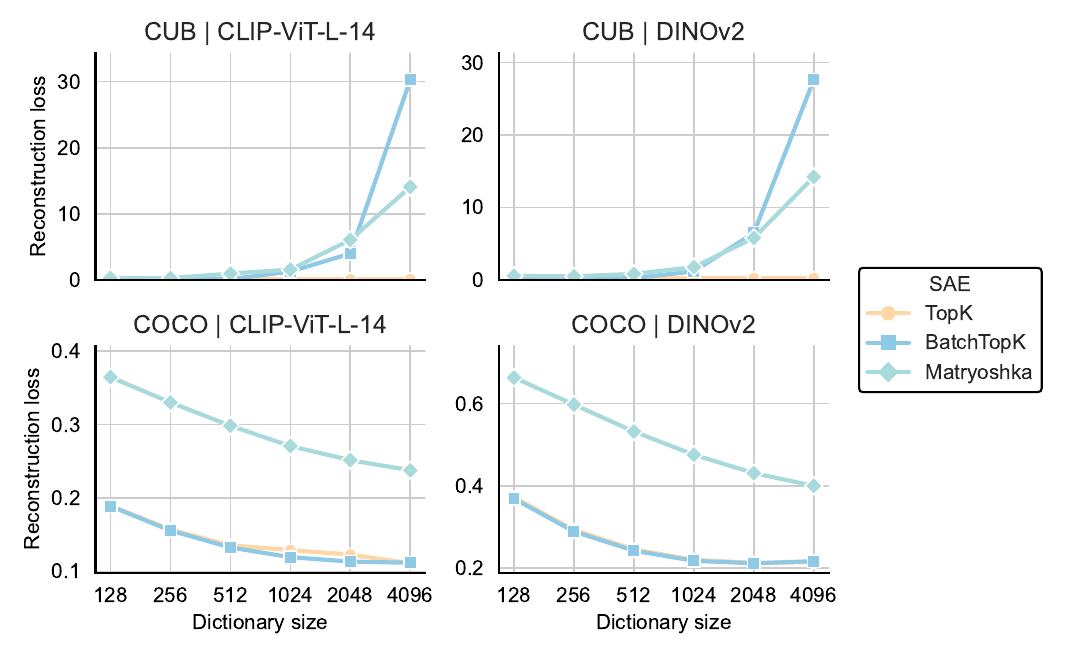}
    \caption{Test reconstruction loss of the TopK, BatchTopK, and Matryoshka SAE variants as a function of dictionary size. Results are shown for features extracted from CLIP-ViT-L/14 and DINOv2 on the CUB and COCO datasets.}
    \label{fig:app:sae_training}
    \vspace{-2em}
\end{wrapfigure}
For the Matryoshka SAE, multiple nested intermediate reconstructions are produced during decoding. Reconstruction losses are computed for each intermediate reconstruction and averaged to form the final reconstruction term. The resulting objective is:
$
\mathcal{L} = \overline{\mathcal{L}}_{\text{rec}} + \mathcal{L}_{\ell_1} + \mathcal{L}_{\text{aux}},
$
where $\overline{\mathcal{L}}_{\text{rec}}$ denotes the mean reconstruction loss across all intermediate reconstruction stages and the final output.

For the JumpReLU SAE~\cite{rajamanoharan2024jumping}, sparsity is not enforced through a fixed activation budget but through a learned per-latent threshold: the JumpReLU activation zeroes all pre-activations below their threshold. The training objective combines the reconstruction loss with an $\ell_0$ sparsity penalty on the activations,
\begin{equation}
\mathcal{L} = \mathcal{L}_{\text{rec}} + \lambda \|a\|_0 ,
\end{equation}
where the thresholds are optimized using straight-through estimators following \cite{rajamanoharan2024jumping}. We use a sparsity coefficient of $\lambda = 0.001$ and a kernel bandwidth of $0.001$ for the rectangle pseudo-gradients of both the JumpReLU activation and the $\ell_0$ step function.

Figure~\ref{fig:app:sae_training} reports the reconstruction loss on the test set as a function of dictionary size, illustrating the scaling behavior of the different SAE variants across datasets and backbone models. Figure~\ref{fig:app:sae_training} shows that reconstruction loss decreases consistently with increasing dictionary size on COCO for both CLIP-ViT-L/14 and DINOv2 features, indicating that larger dictionaries improve reconstruction capacity on this more diverse dataset. In contrast, on CUB the reconstruction loss for BatchTopK and Matryoshka increases as the dictionary size grows. This behavior suggests overfitting when the dictionary becomes too large relative to the complexity of the CUB dataset, where additional latent capacity does not translate into improved reconstruction performance.

\section{Qualitative Matching Results}
\label{app:qualitative_matching_results}

\begin{figure}
    \centering
        \includegraphics[width=1\linewidth]{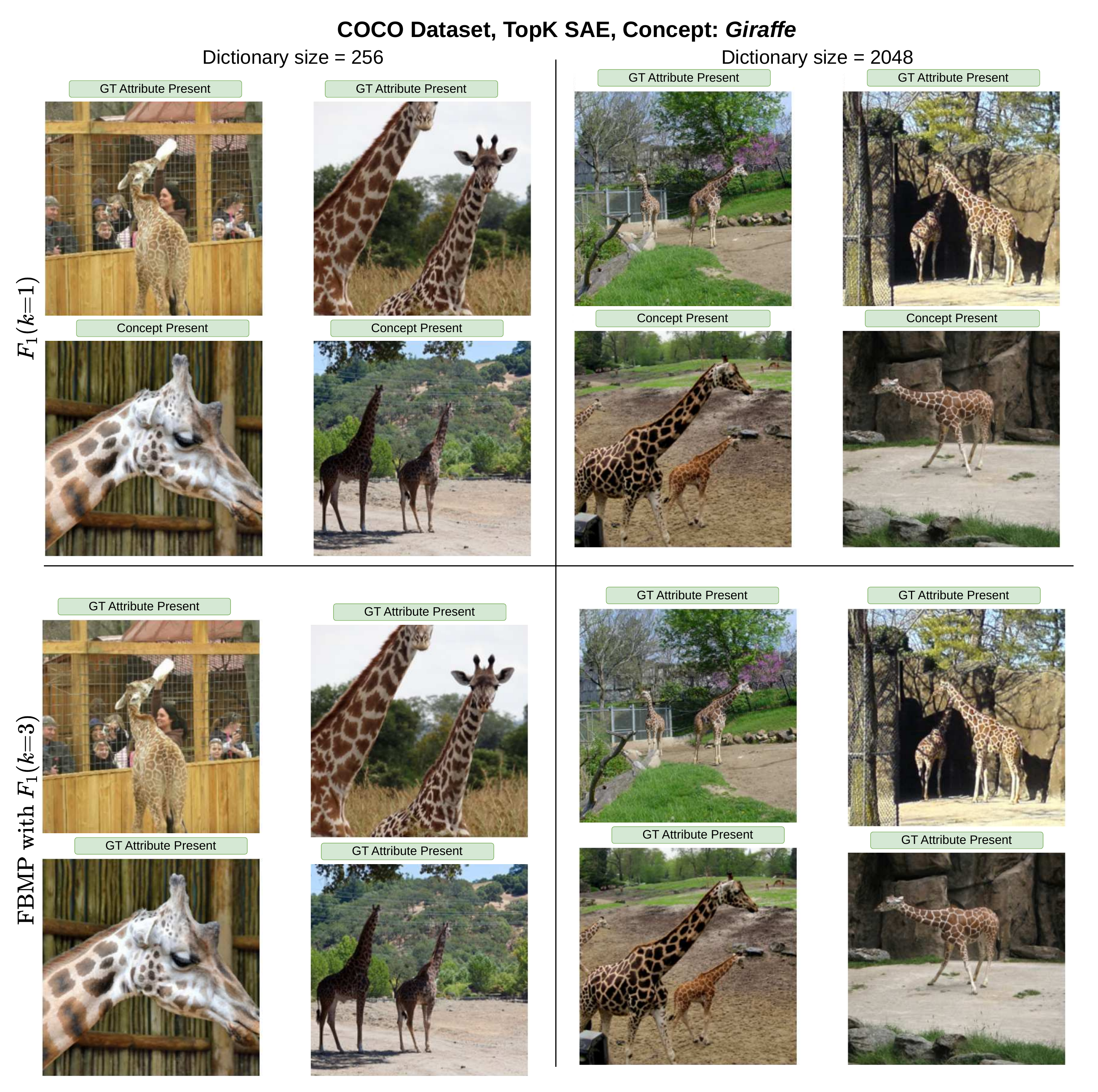}
    \caption{Qualitative latent--concept matching example for the COCO concept ``Giraffe''. Columns correspond to SAE dictionary sizes ($d=256$ and $d=2048$), while rows compare the matching procedures (F1 one-to-one matching and FBMP many-to-one matching).}
    \label{fig:matching_qualitative_giraffe}
\end{figure}

\begin{figure}
    \centering
        \includegraphics[width=1\linewidth]{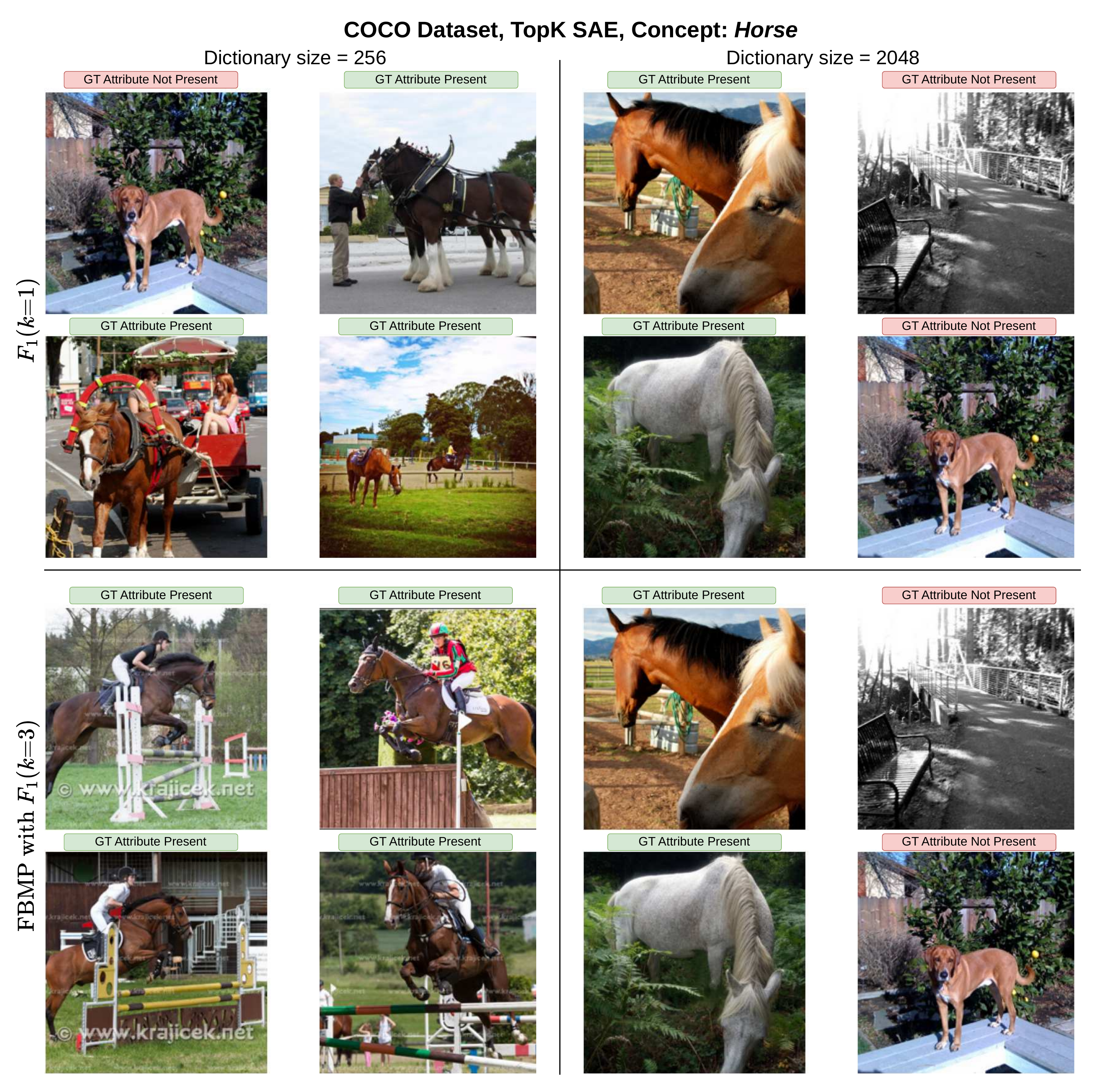}
    \caption{Qualitative latent--concept matching example for the COCO concept ``Horse''. Columns correspond to SAE dictionary sizes ($d=256$ and $d=2048$), while rows compare the matching procedures (F1 one-to-one matching and FBMP many-to-one matching).}
    \label{fig:matching_qualitative_horse}
\end{figure}

Each qualitative plot shows the four highest-activating samples for a matched latent--concept pair. More precisely, given a ground-truth concept, we first identify the corresponding latent or latent coalition using either F1 matching or FBMP. For one-to-one matching and FBMP with coalition size $k=1$, retrieval is based on the activation of the single matched latent. For FBMP coalitions with $k>1$, we rank samples by the maximum activation across all latents in the matched coalition, such that the displayed images correspond to those with the strongest evidence for the matched concept. The green box ``GT Attribute Present'' indicates that the image was labeled with the corresponding attribute.

Figure~\ref{fig:matching_qualitative_giraffe} and Figure~\ref{fig:matching_qualitative_horse} show qualitative results on COCO concepts. For the concept ``giraffe'', all four matching configurations correctly identify the top four activating images, suggesting that this concept is robustly and consistently captured across dictionary sizes and matching procedures. For the concept ``horse'', rows compare F1 one-to-one matching and FBMP many-to-one matching, while columns correspond to dictionary sizes $d=256$ and $d=2048$. For the larger dictionary size, both matching procedures retrieve very similar samples, with two failure cases: one image containing a bench and one containing a dog, suggesting that the corresponding latent is not fully disentangled and partially captures visually related but semantically distinct content. For the smaller dictionary size, the retrieved samples differ more strongly between F1 and FBMP: F1 retrieves three correct horse images alongside one dog image, whereas FBMP retrieves four correct horse images, demonstrating the benefit of coalition-based matching under limited dictionary capacity.

\begin{wrapfigure}{r}{0.4\textwidth}
    \centering
    \vspace{-2.5em}
    \includegraphics[width=\linewidth]{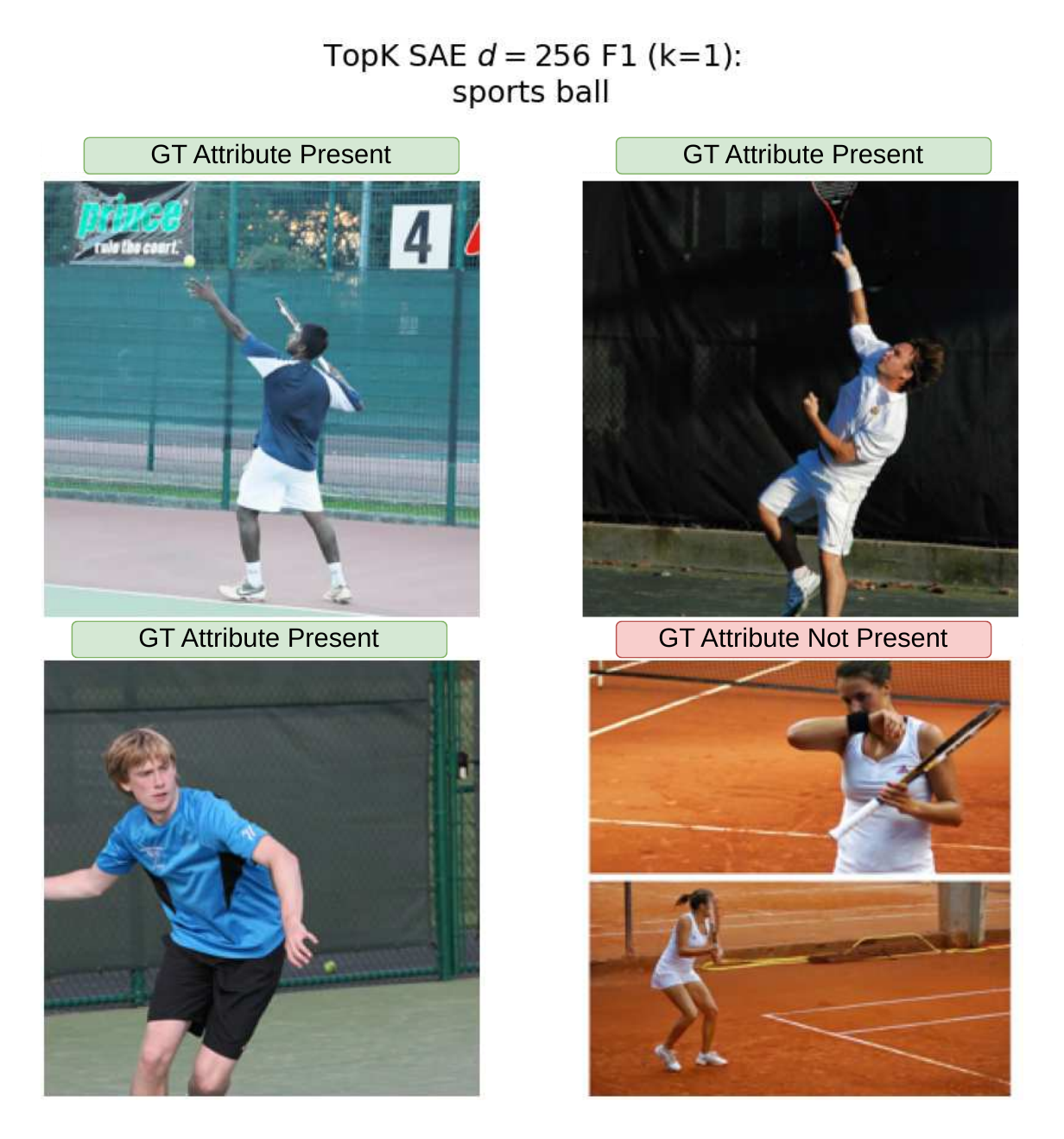}
    \caption{Qualitative matching results for the COCO concept \textit{sports ball} using $F_1$ matching for a TopK SAE with $d=2048$. All four retrieved images are tennis-related.}
    \label{fig:matching_qualitative_ball}
    \vspace{-2.5em}
\end{wrapfigure}

Figure~\ref{fig:matching_qualitative_ball} illustrates a more subtle failure mode for the concept ``sports ball''. While two of the four retrieved images clearly contain a tennis ball and one carries the corresponding ground-truth label, the fourth image is neither labeled nor visibly contains the object. Notably, all four images are strongly tennis-related, suggesting that the matched latent has conflated the ``sports ball'' concept with the broader ``tennis'' context. This type of semantic entanglement, where a latent captures a correlated concept rather than the target one, is precisely the kind of error that TAPAScore is designed to detect, as a truly selective latent should respond to the presence of the object itself rather than to its co-occurring context.

\begin{figure}
    \centering
    \includegraphics[width=1\linewidth]{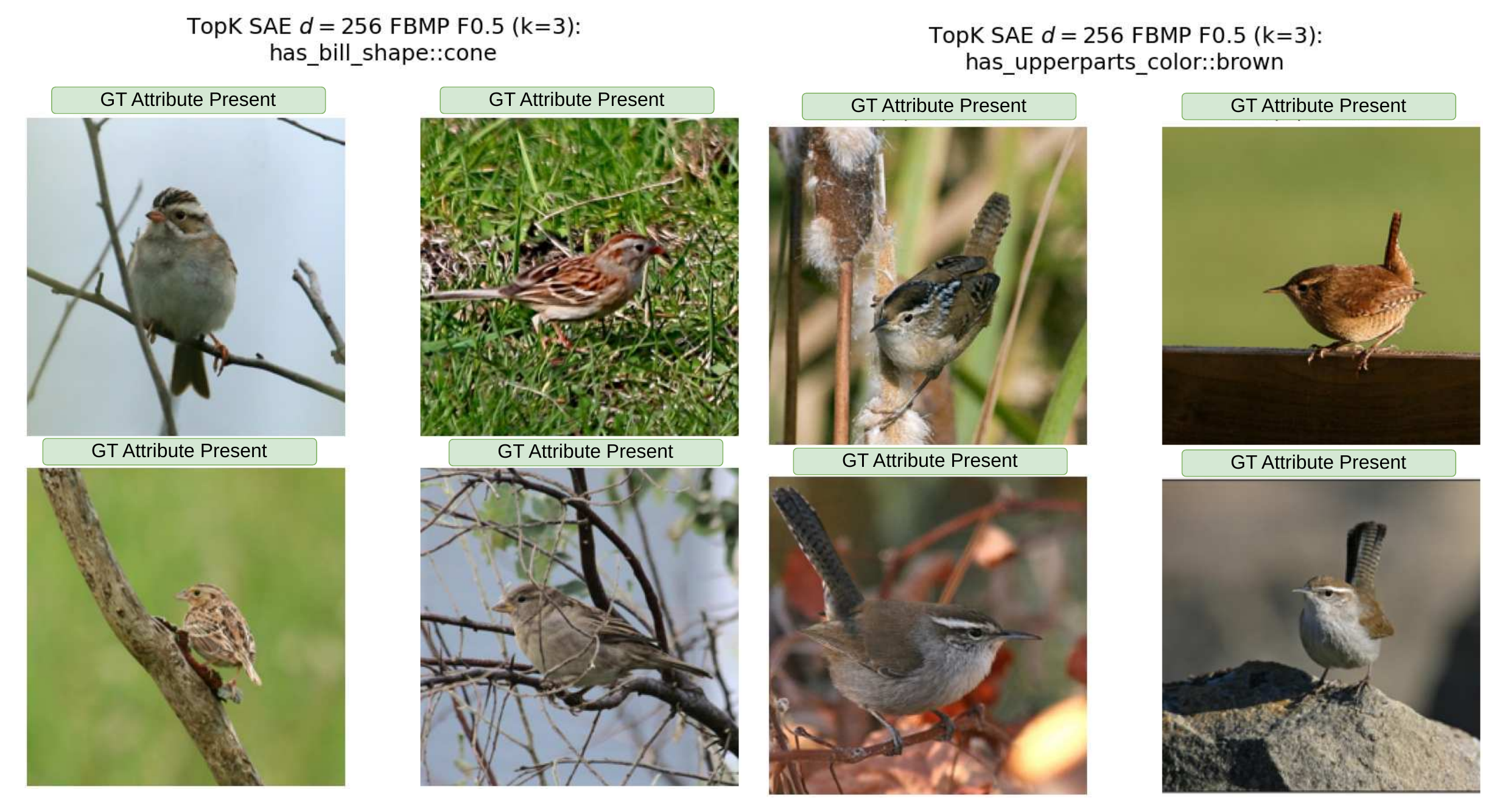}
    \caption{Qualitative matching results for two CUB concepts using a TopK SAE with $d=256$ and FBMP ($k=3$, $F_{0.5}$). All four highest-activating images are correctly retrieved and labeled for both \textit{has bill shape: cone} (left) and \textit{has upperparts color: brown} (right), demonstrating that FBMP can reliably identify both fine-grained morphological and color concepts.}
    \label{fig:matching_qualitative_sparrow}
\end{figure}

\begin{figure}
    \centering
    \includegraphics[width=1\linewidth]{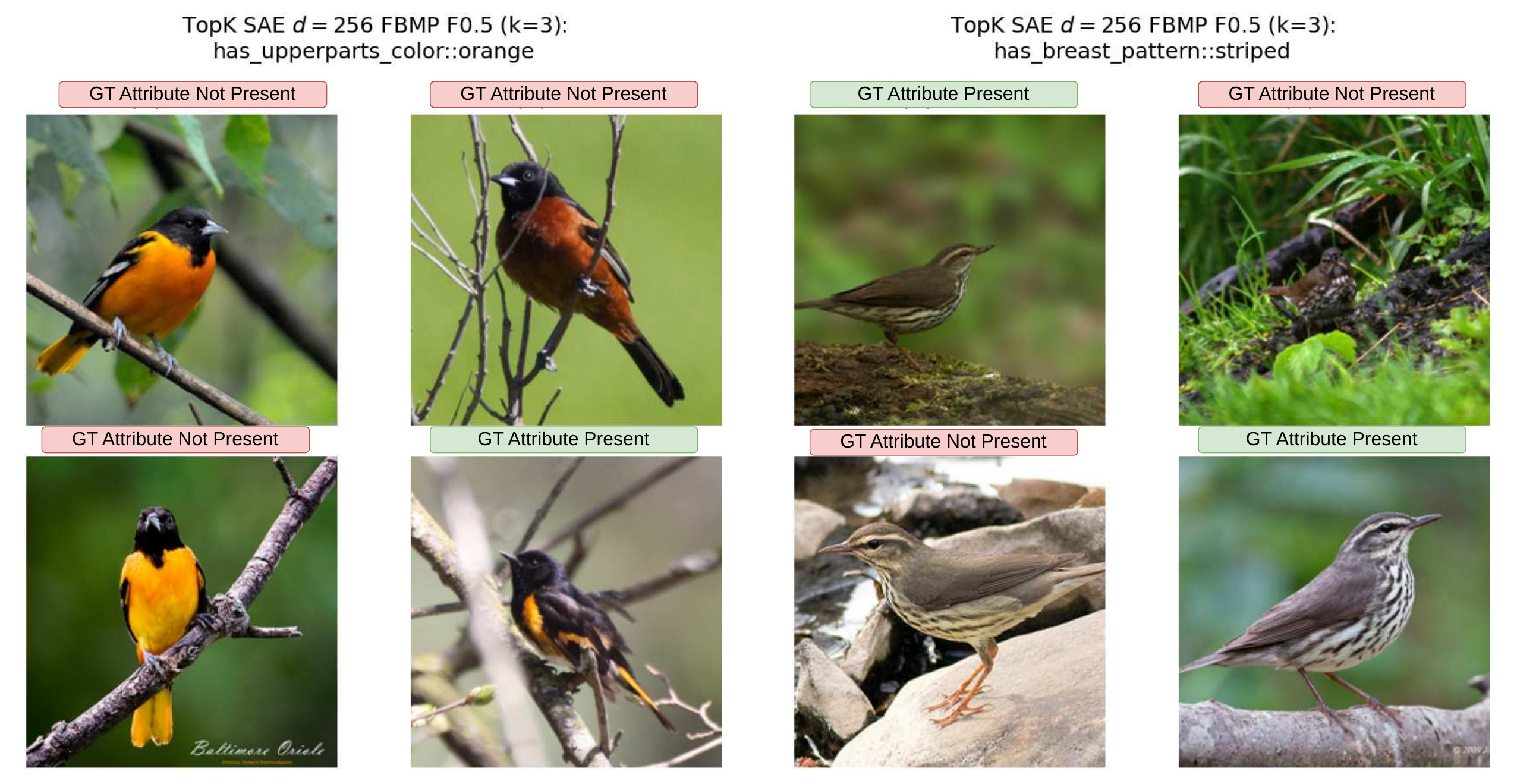}
    \caption{Qualitative matching results illustrating limitations caused by spatial specificity and annotation noise in CUB. For \textit{has upperparts color: orange} (left), all retrieved birds display orange plumage, suggesting that the latents do capture the general color concept but not its precise spatial localization (upperparts vs.\ underparts). For \textit{has breast pattern: striped} (right), the retrieved images are visually consistent with the striped pattern, even though the ground-truth annotations are not always correctly aligned with the visual evidence.}    
    \label{fig:matching_qualitative_problems_cub}
\end{figure}

Figures~\ref{fig:matching_qualitative_sparrow} and~\ref{fig:matching_qualitative_problems_cub} show qualitative matching results for four CUB concepts using the same configuration: a TopK SAE with $d=256$ and FBMP ($k=3$, $F_{0.5}$). Figure~\ref{fig:matching_qualitative_sparrow} illustrates two successful cases: for \textit{has bill shape: cone}, all four highest-activating images are correctly retrieved and carry the ground-truth label, demonstrating that even fine-grained morphological concepts are reliably identified; a similar result holds for \textit{has upperparts color: brown}, confirming that color concepts are also well captured by the learned latent coalitions.

\begin{wrapfigure}{r}{0.4\textwidth}
    \centering
    \vspace{-2.5em}
    \includegraphics[width=\linewidth]{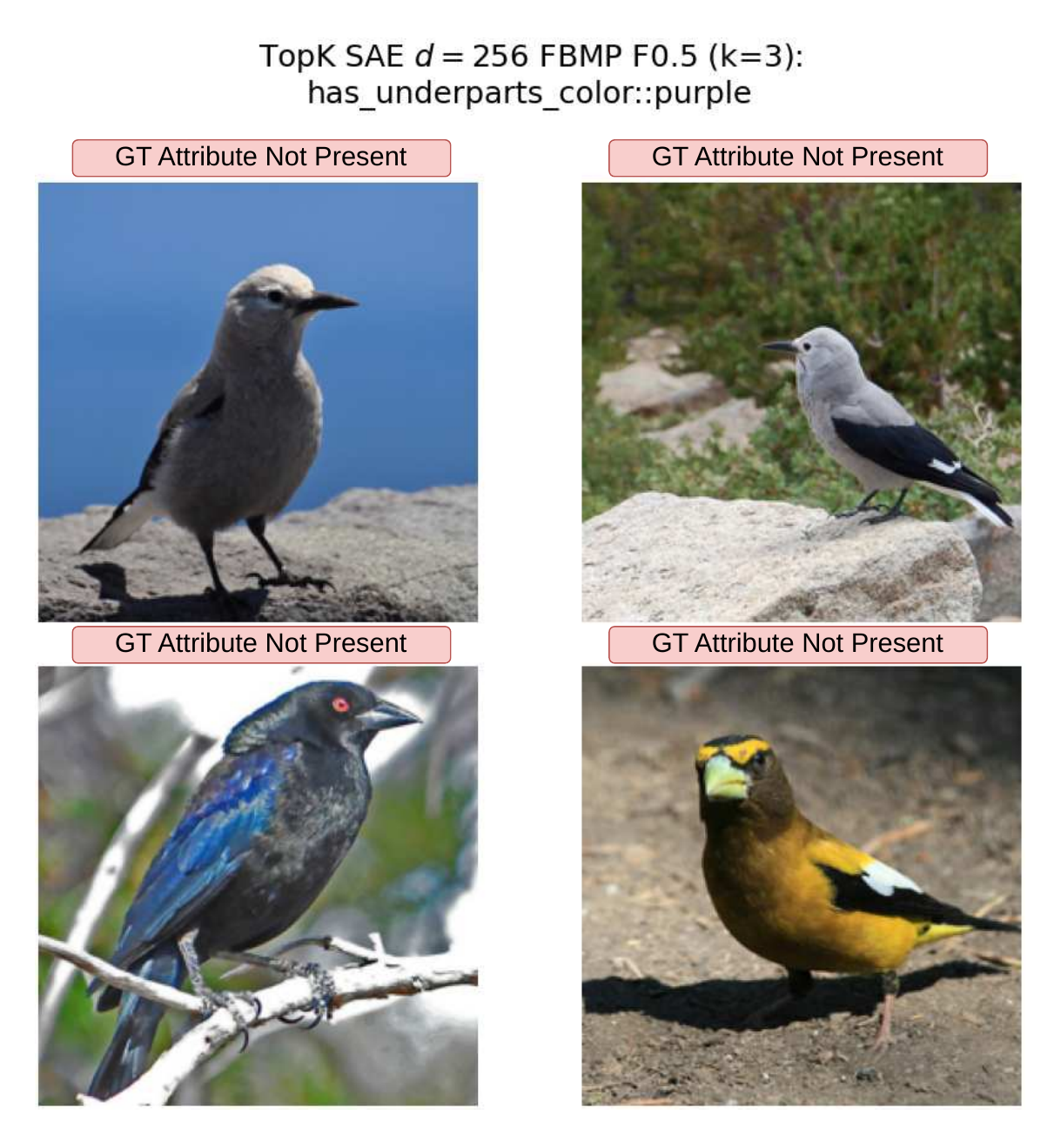}
    \caption{Qualitative matching results for the CUB concept \textit{has underparts color: purple} using a TopK SAE with $d{=}256$ and FBMP ($k=3$, $F_{0.5}$).}
    \label{fig:matching_qualitative_purple}
    \vspace{-2.5em}
\end{wrapfigure}

Figure~\ref{fig:matching_qualitative_problems_cub} highlights two cases where retrieval quality is imperfect. For \textit{has upperparts color: orange}, three of the four retrieved images display orange plumage but have negative ground-truth labels. Closer inspection reveals that the orange coloration in these images is located on the underparts rather than the upperparts, indicating that the matched latent captures the color concept without encoding its spatial localization. This suggests that the granularity of attribute definitions, distinguishing \textit{upperparts} from \textit{underparts} color, poses a challenge for the SAE, which may learn a more general color representation that does not respect part-level distinctions. The ambiguity inherent in such spatially specific annotations can lead to annotation inconsistencies that negatively impacts the matching quality. For \textit{has breast pattern: striped}, retrieved images are visually correct, but one shows the striped pattern only partially and another is clearly mislabeled, further illustrating how annotation noise can artificially degrade matching scores even when the underlying latent captures the intended concept.

Figure~\ref{fig:matching_qualitative_purple} illustrates a failure case for the rare concept \textit{has underparts color: purple}, where none of the four retrieved images contain the intended attribute. 
Unlike the annotation noise observed in Figure~\ref{fig:matching_qualitative_problems_cub},  this failure is related to concept rarity: purple underparts occur in very few CUB samples, providing insufficient training signal for the SAE to learn a dedicated latent. This suggests that matching quality is not only sensitive to annotation noise but also to the frequency of concepts in the underlying data distribution.

\section{Additional Results}
\label{app:sec:additional_results}
\label{app:subsubsec:dino}
\begin{figure*}[t]
\centering
  \begin{subfigure}[t]{0.24\textwidth}
    \centering
    \includegraphics[width=\linewidth]{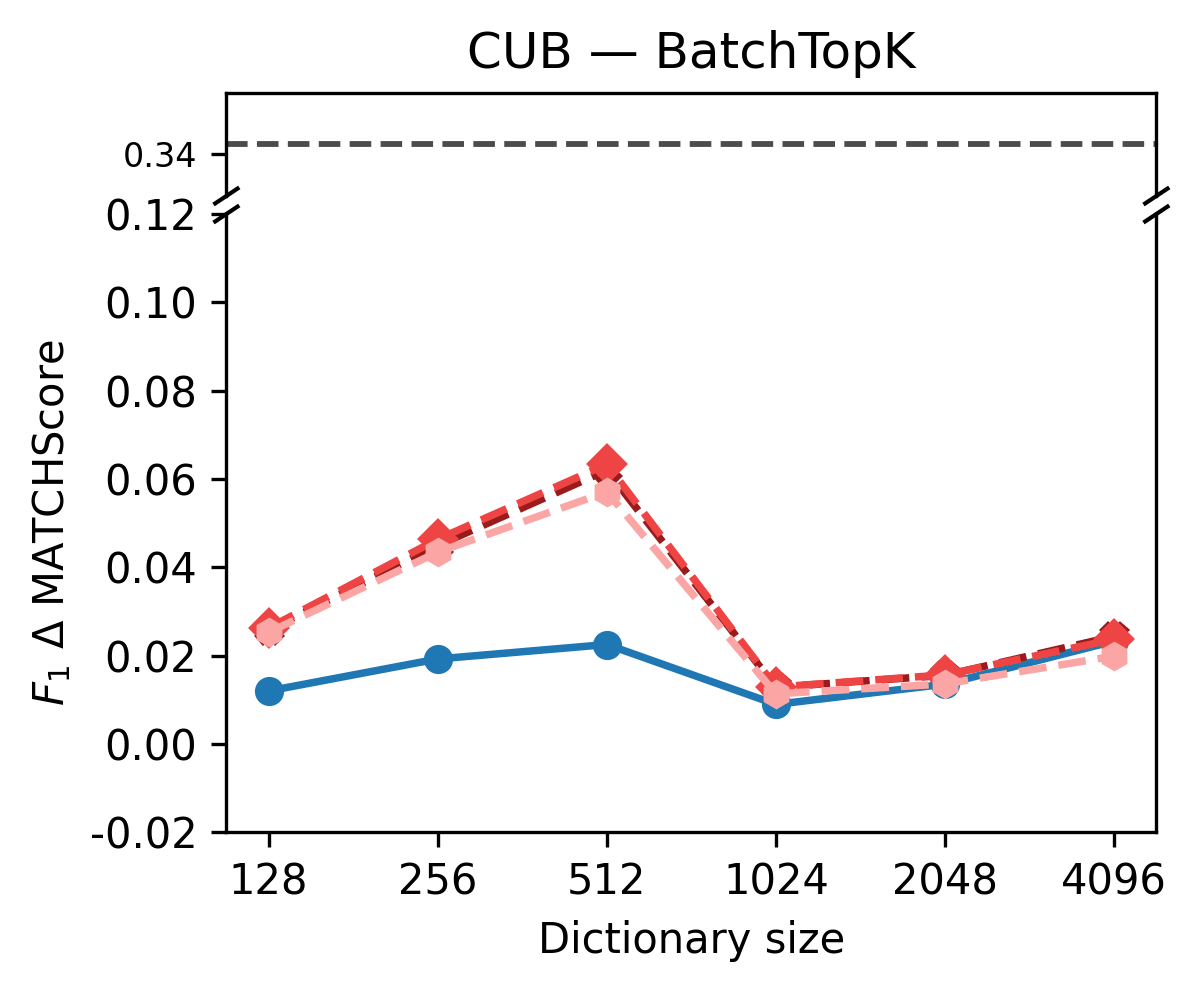}
  \end{subfigure}
  \hfill
  \begin{subfigure}[t]{0.24\textwidth}
    \centering
    \includegraphics[width=\linewidth]{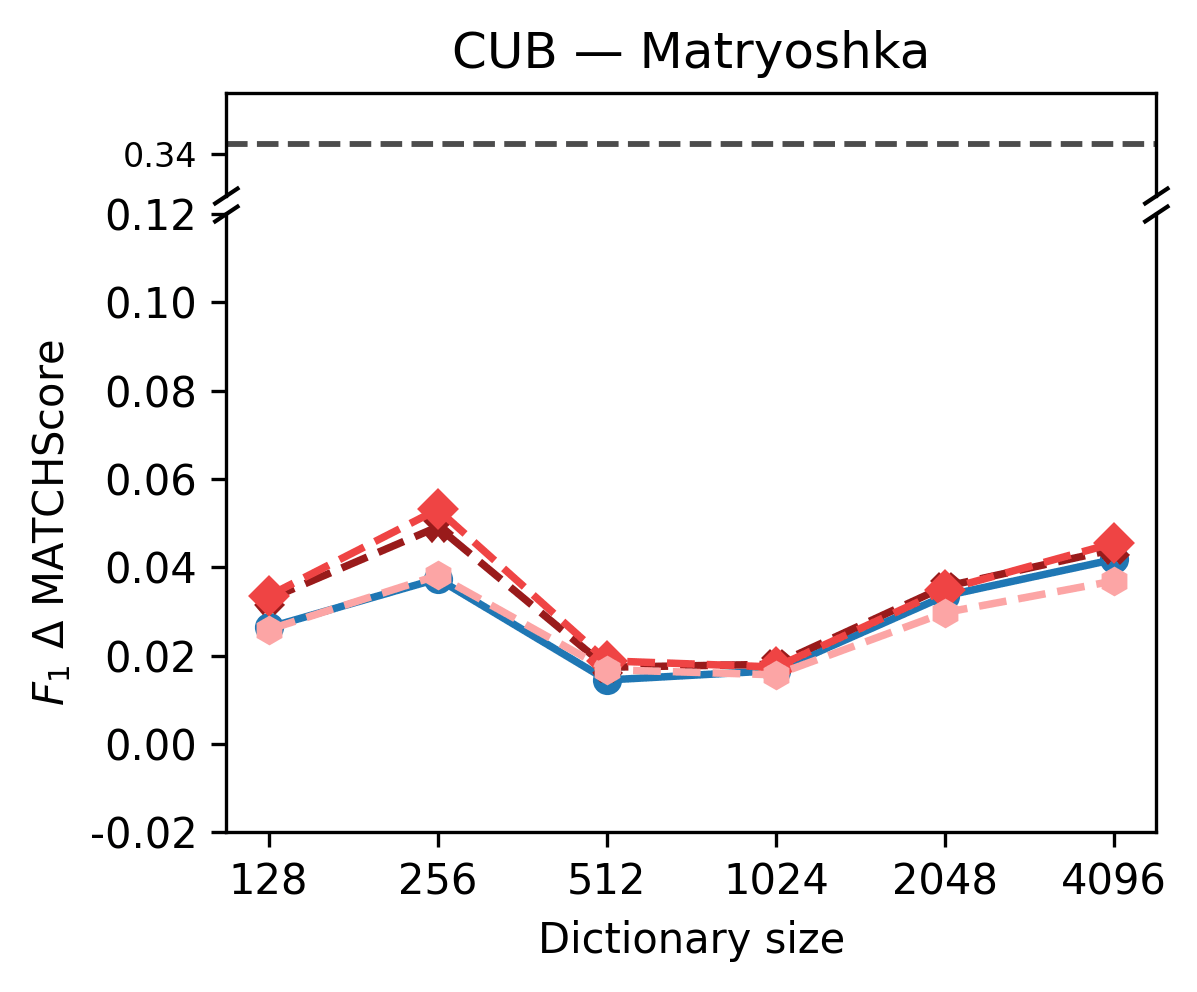}
  \end{subfigure}
  \hfill
  \begin{subfigure}[t]{0.24\textwidth}
    \centering
    \includegraphics[width=\linewidth]{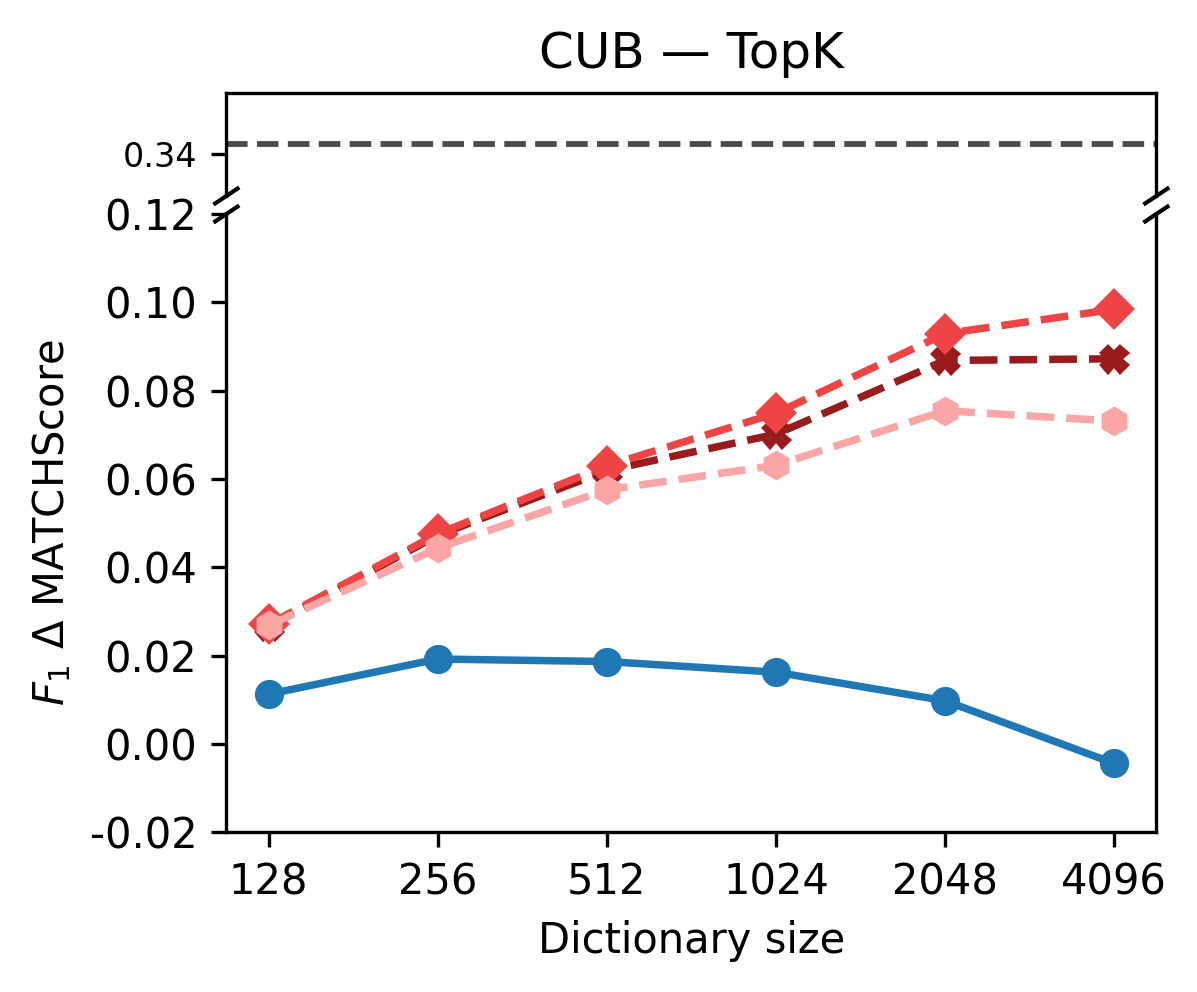}
  \end{subfigure}
  \hfill
  \begin{subfigure}[t]{0.24\textwidth}
    \centering
    \includegraphics[width=\linewidth]{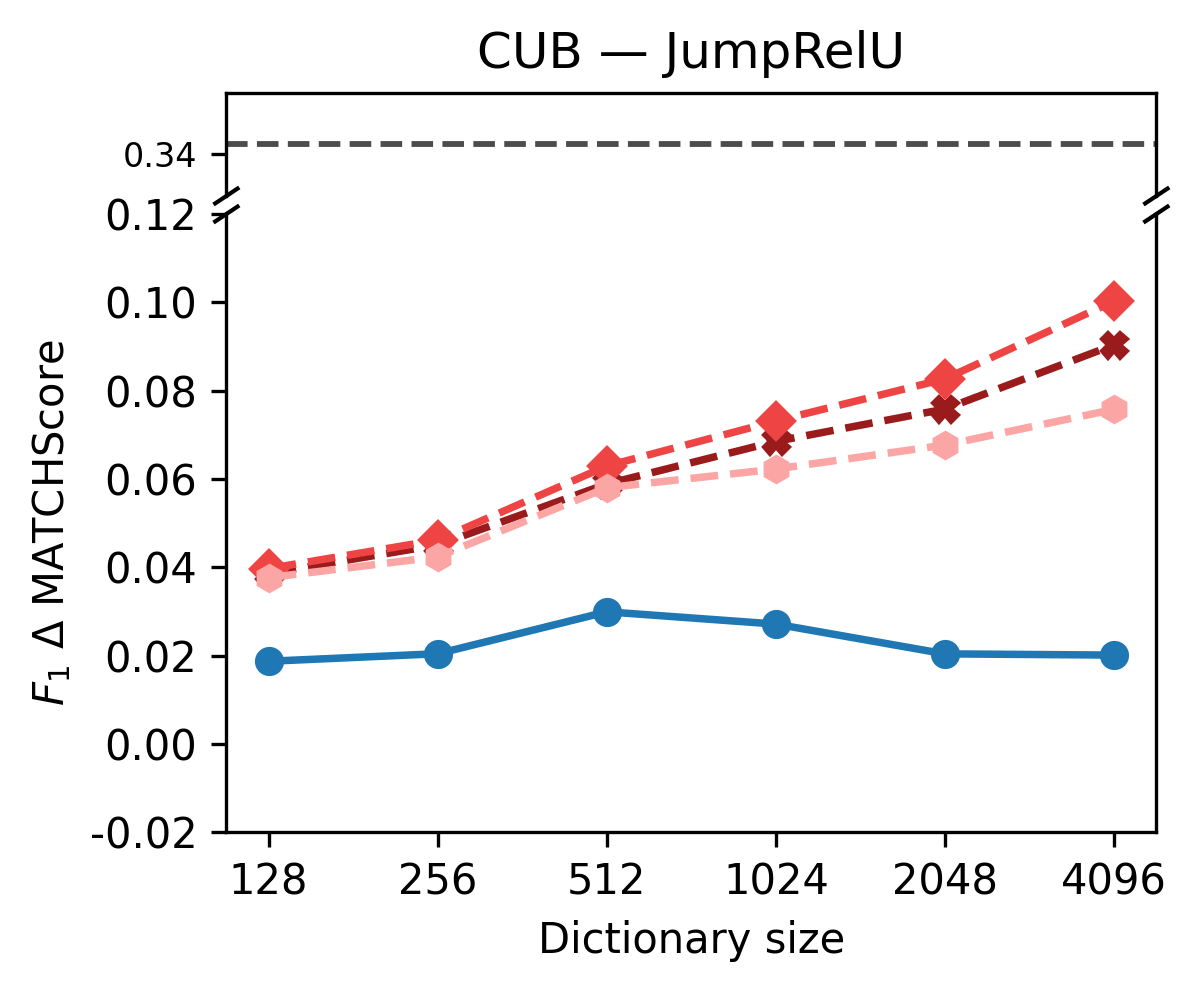}
  \end{subfigure}

  \vspace{-1mm}

  \begin{subfigure}[t]{0.24\textwidth}
    \centering
    \includegraphics[width=\linewidth]{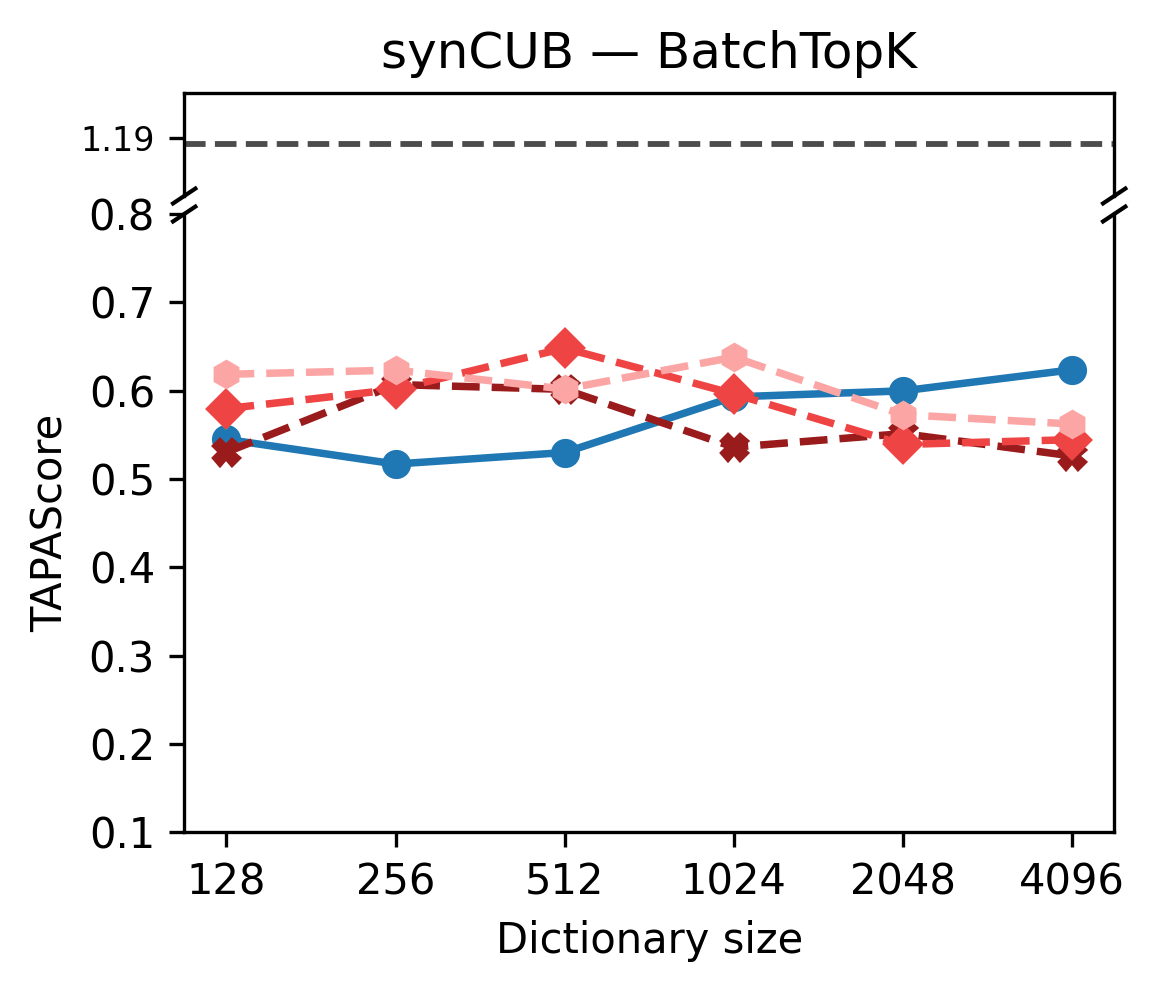}
  \end{subfigure}
  \hfill
  \begin{subfigure}[t]{0.24\textwidth}
    \centering
    \includegraphics[width=\linewidth]{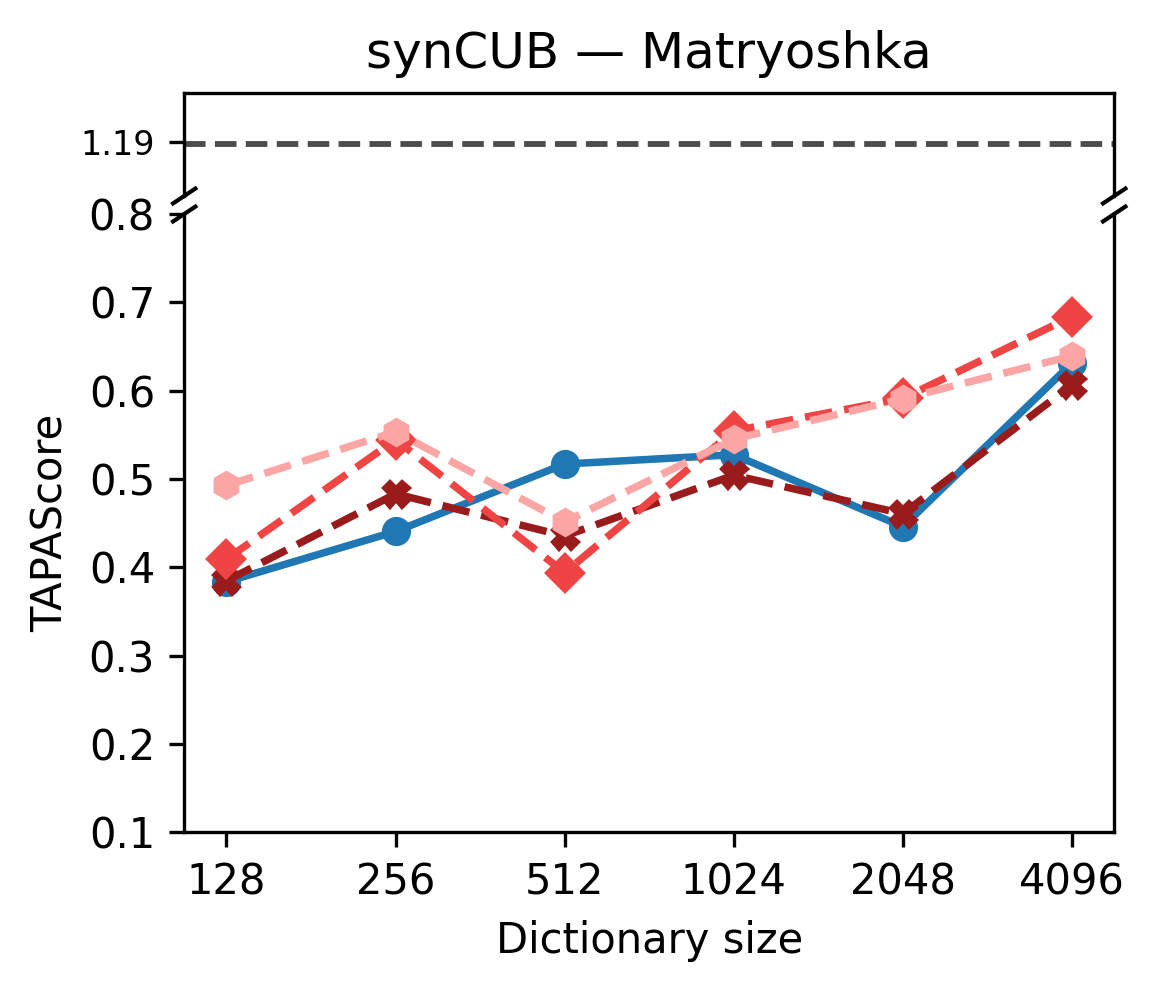}
  \end{subfigure}
  \hfill
  \begin{subfigure}[t]{0.24\textwidth}
    \centering
    \includegraphics[width=\linewidth]{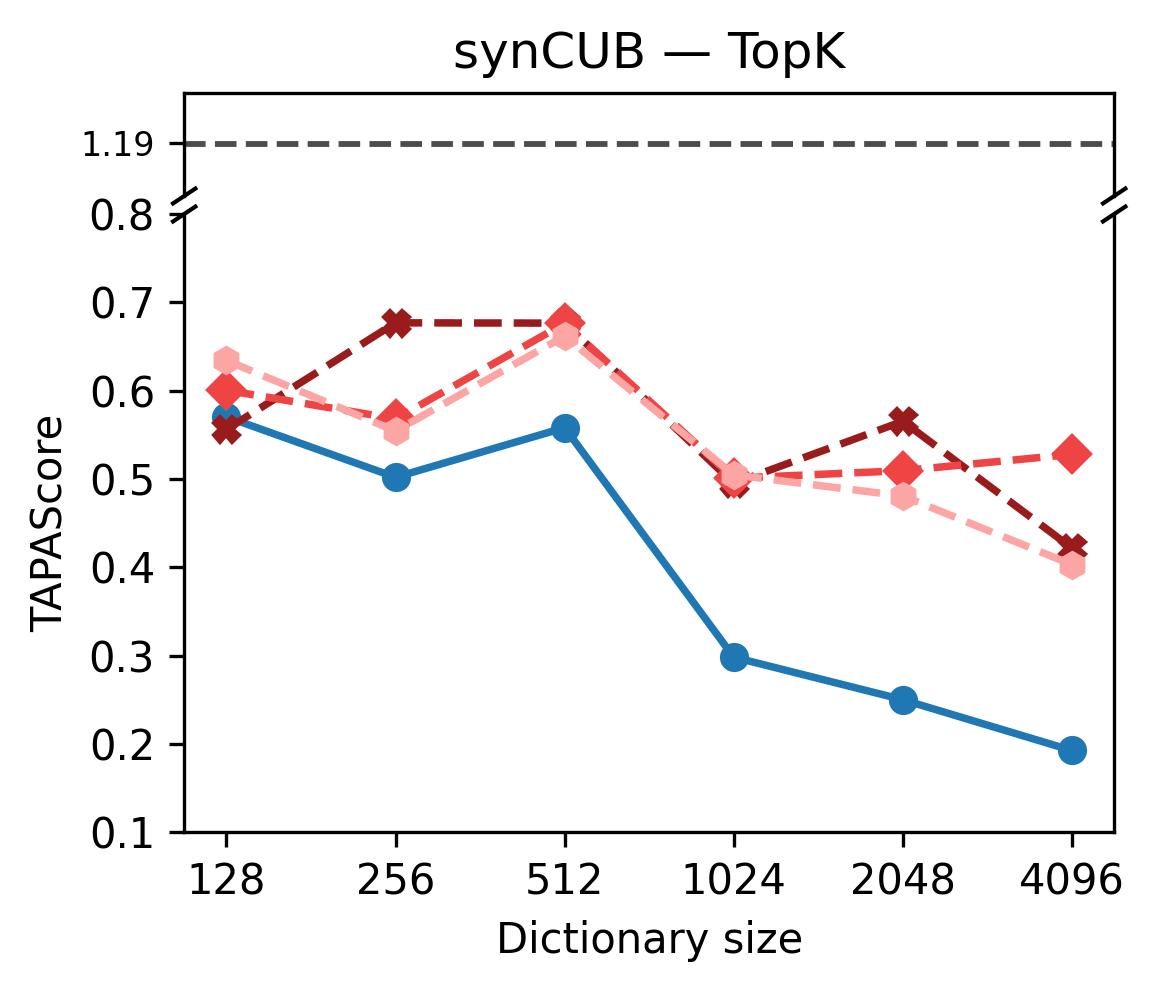}
  \end{subfigure}
  \hfill
  \begin{subfigure}[t]{0.24\textwidth}
    \centering
    \includegraphics[width=\linewidth]{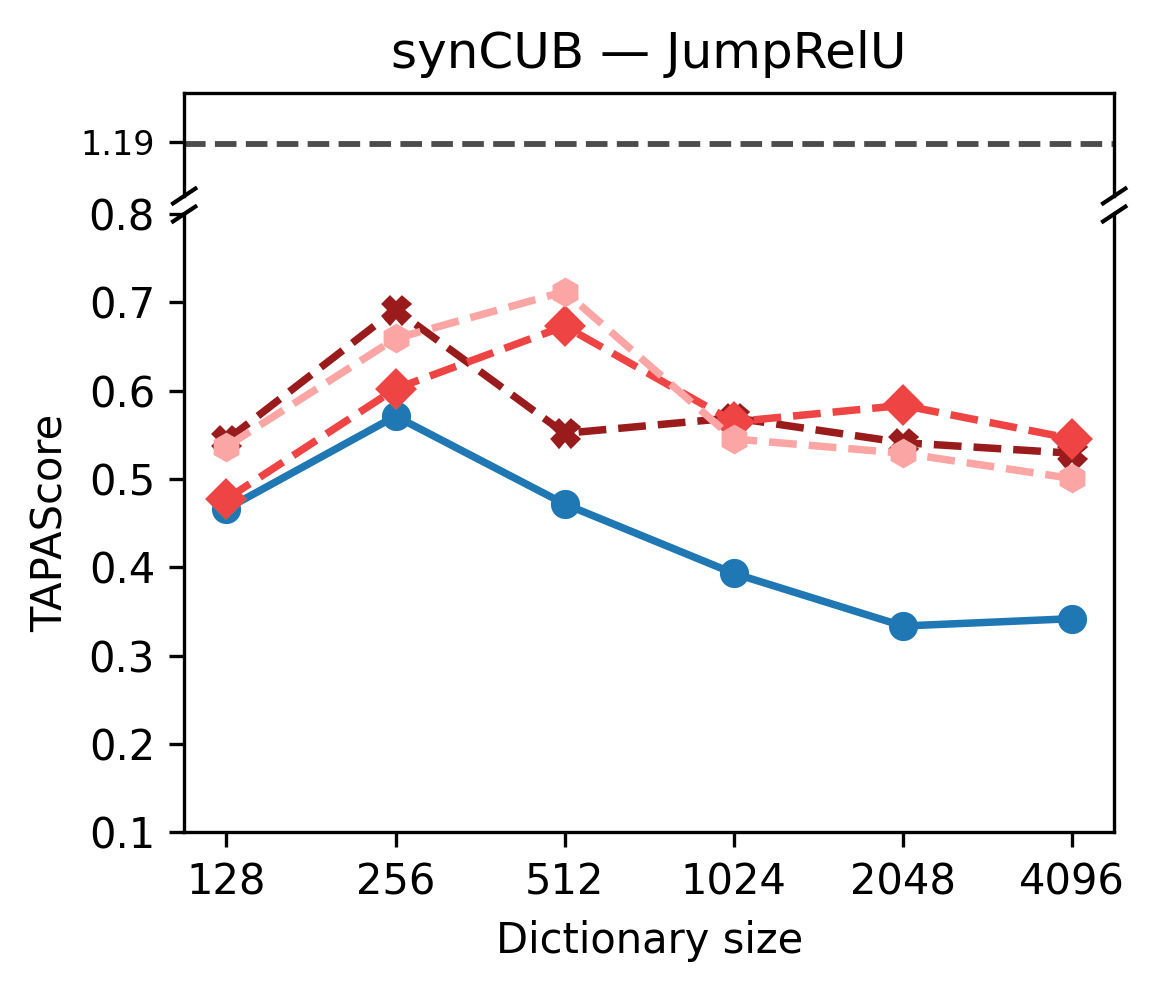}
  \end{subfigure}

  \vspace{-1mm}

  \includegraphics[width=\linewidth]{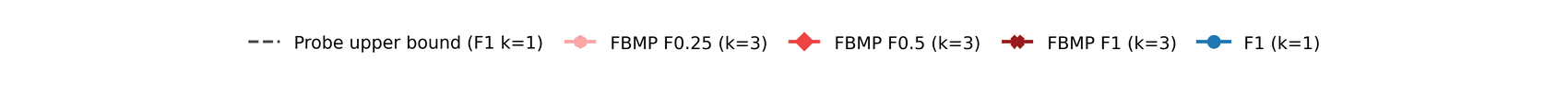}

\vspace{-1.2em}
\caption{$F_1$ $\Delta$MATCHScore on CUB (\textit{top row}) and TAPAScore on synCUB (\textit{bottom row}) as a function of dictionary size for SAEs trained on DINOV2 embeddings of CUB, across BatchTopK, Matryoshka, TopK and JumpReLU SAE variants (\textit{left to right}) and different matching criteria. The gray dashed line marks the supervised linear-probe upper bound.}
\label{fig:matching_and_tapas_cub_dinov2}
\vspace{-2em}
\end{figure*}

\begin{figure*}[t]
\centering
  \begin{subfigure}[t]{0.24\textwidth}
    \centering
    \includegraphics[width=\linewidth]{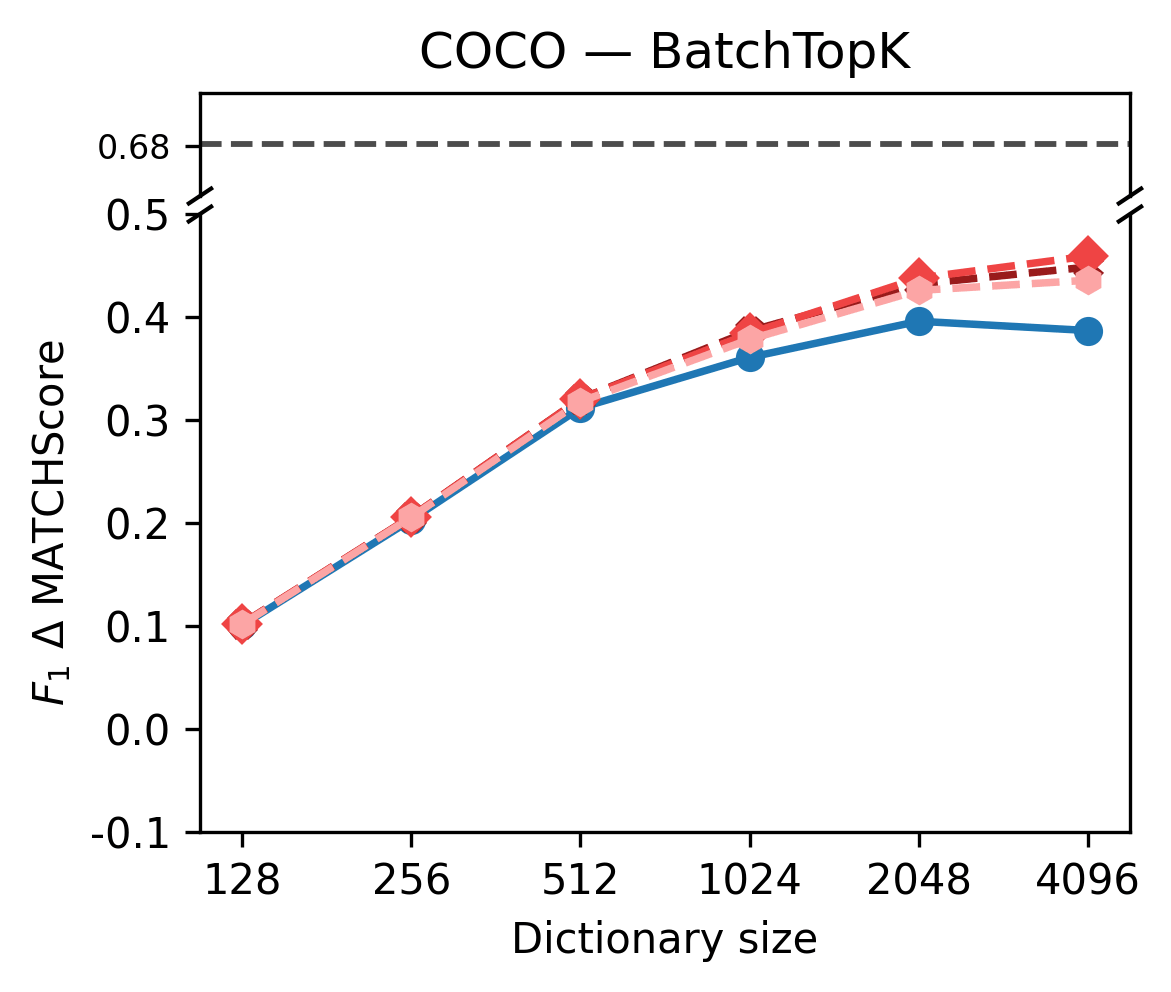}
  \end{subfigure}
  \hfill
  \begin{subfigure}[t]{0.24\textwidth}
    \centering
    \includegraphics[width=\linewidth]{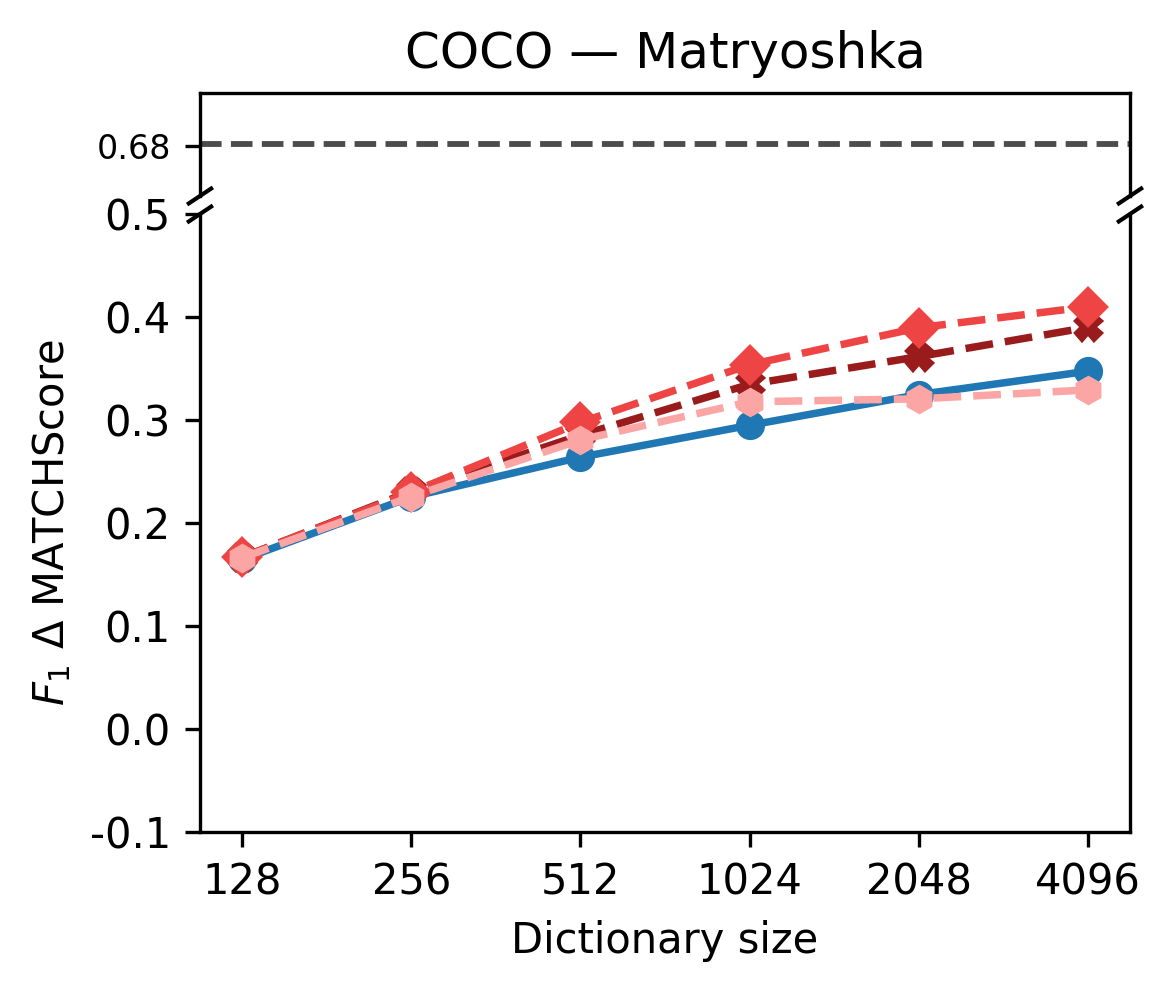}
  \end{subfigure}
  \hfill
  \begin{subfigure}[t]{0.24\textwidth}
    \centering
    \includegraphics[width=\linewidth]{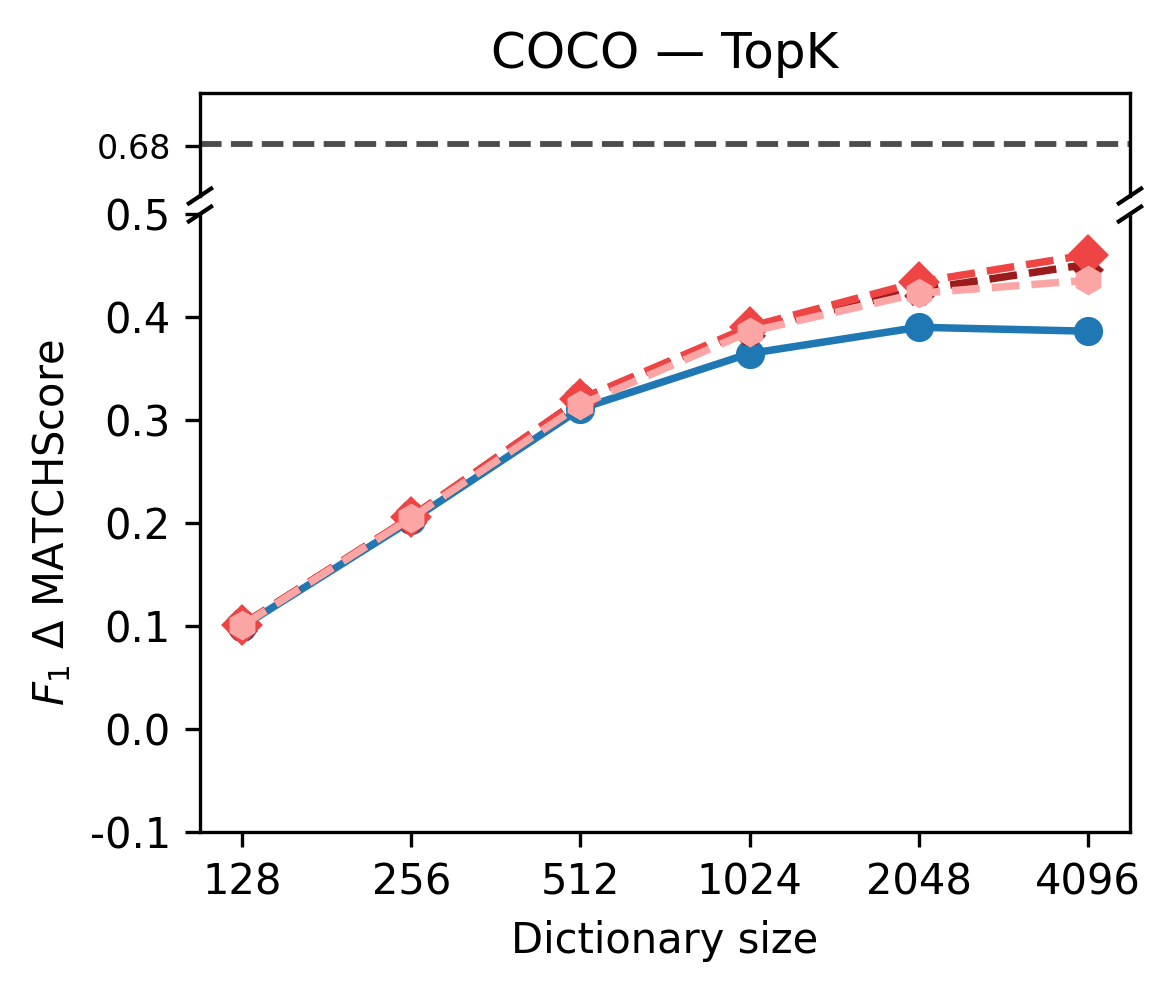}
  \end{subfigure}
  \hfill
  \begin{subfigure}[t]{0.24\textwidth}
    \centering
    \includegraphics[width=\linewidth]{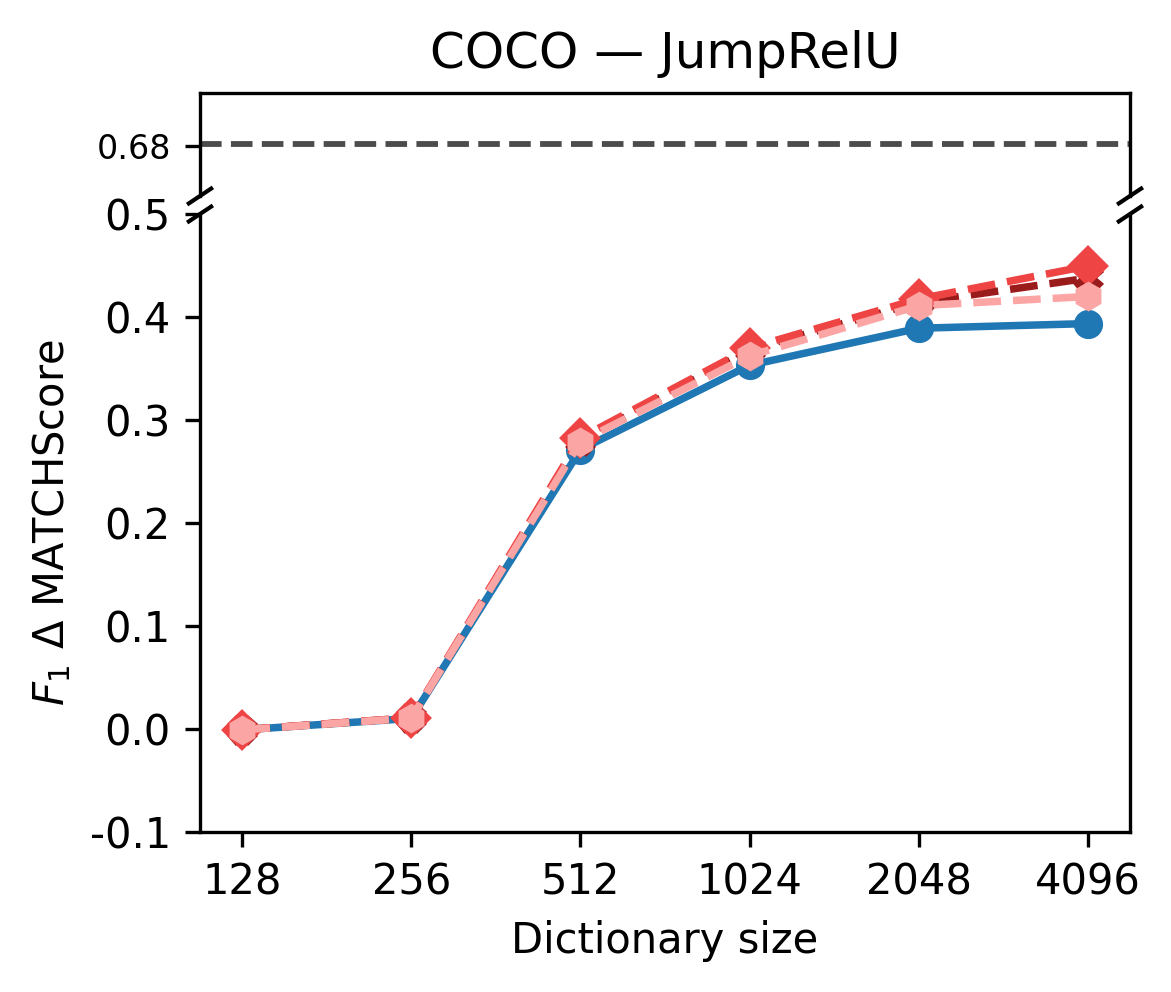}
  \end{subfigure}

  \vspace{-1mm}

  \begin{subfigure}[t]{0.24\textwidth}
    \centering
    \includegraphics[width=\linewidth]{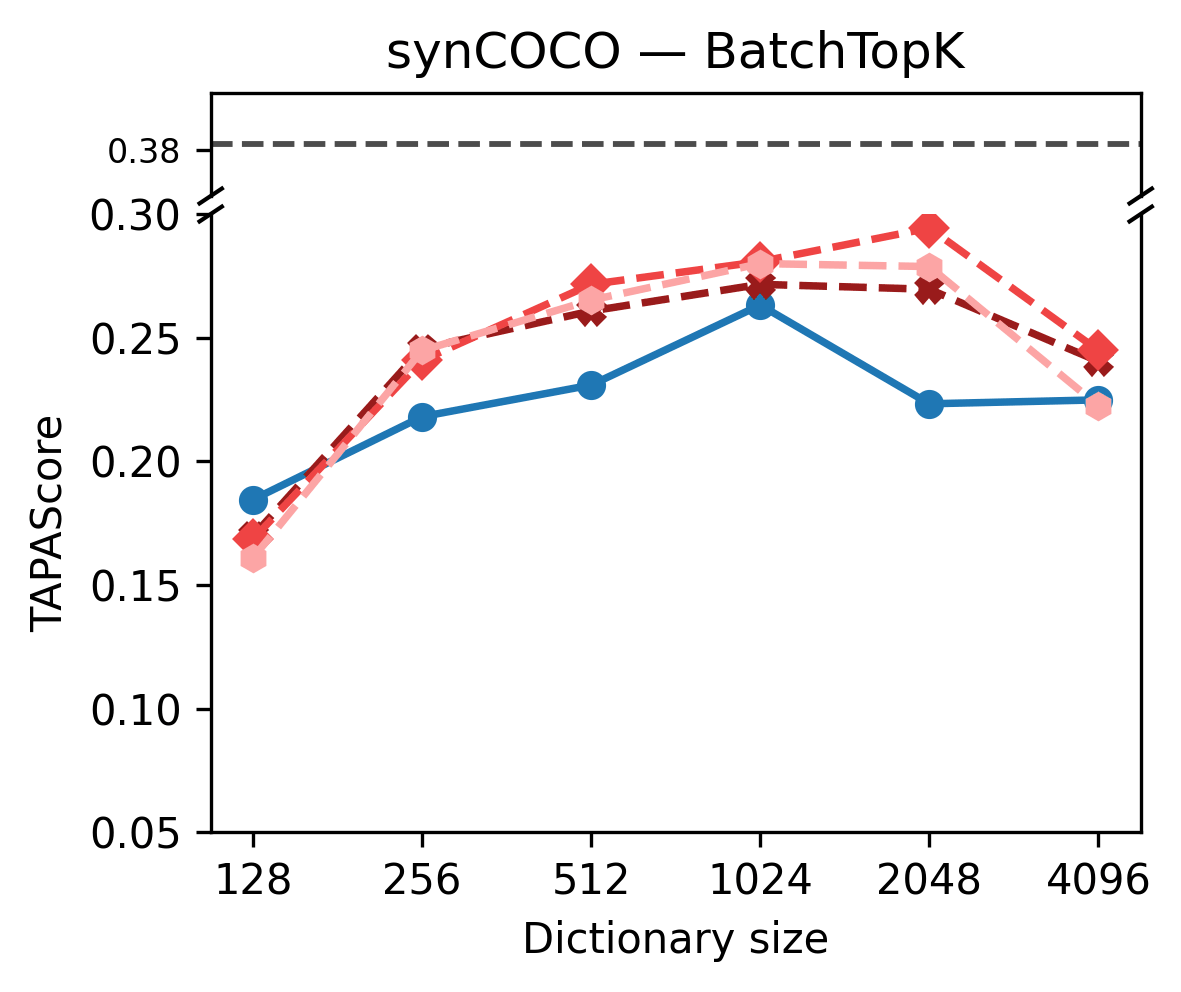}
  \end{subfigure}
  \hfill
  \begin{subfigure}[t]{0.24\textwidth}
    \centering
    \includegraphics[width=\linewidth]{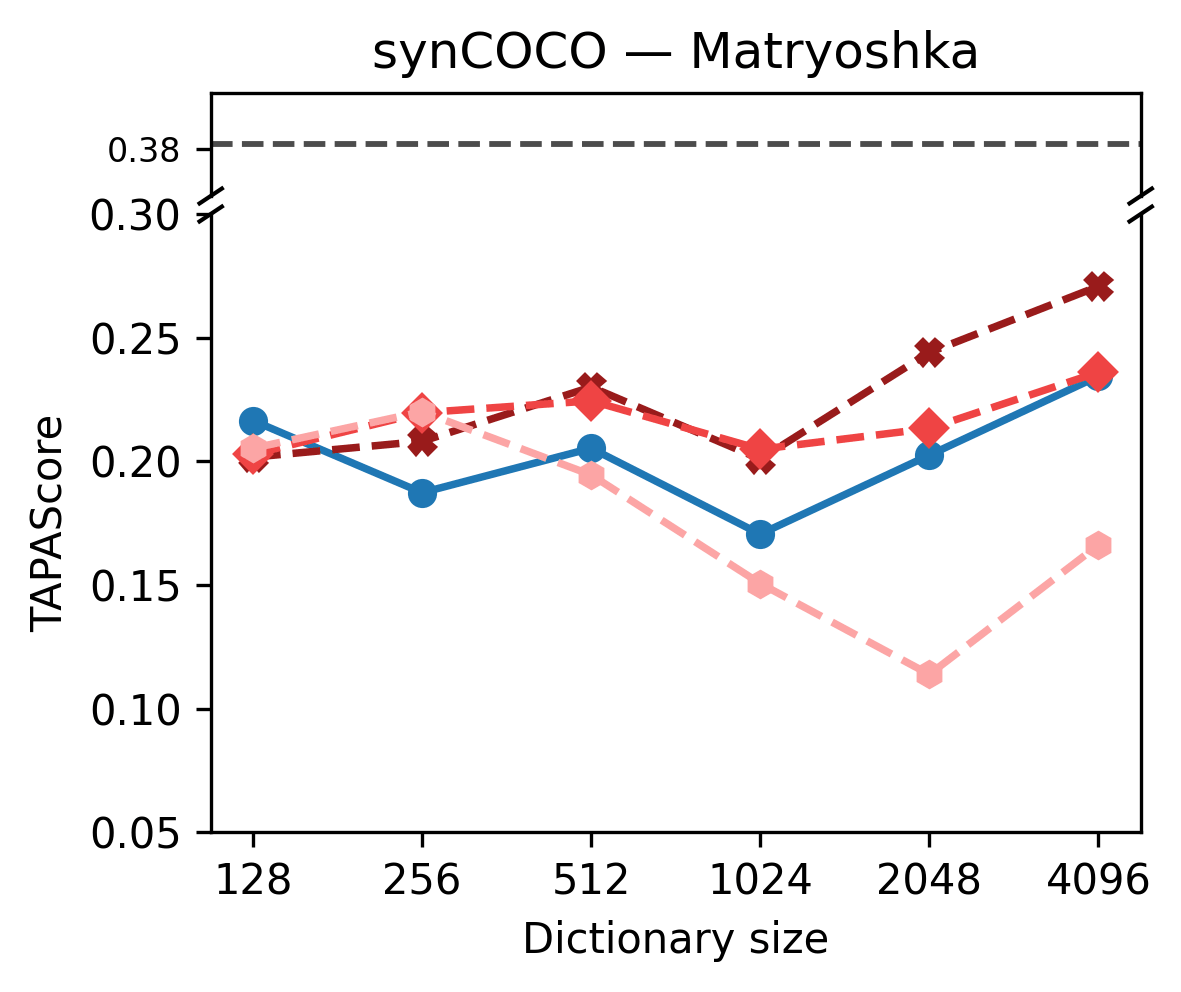}
  \end{subfigure}
  \hfill
  \begin{subfigure}[t]{0.24\textwidth}
    \centering
    \includegraphics[width=\linewidth]{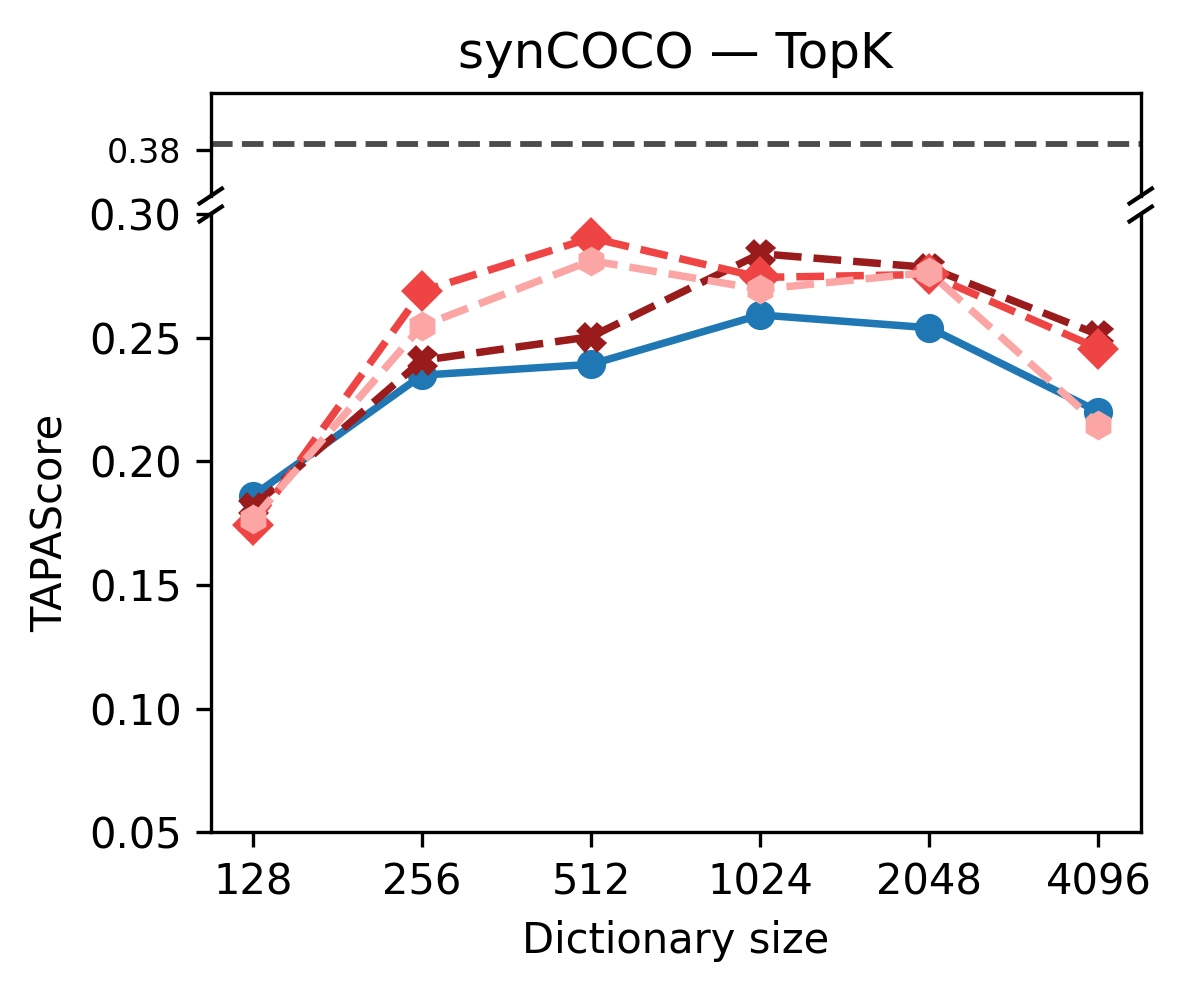}
  \end{subfigure}
  \hfill
  \begin{subfigure}[t]{0.24\textwidth}
    \centering
    \includegraphics[width=\linewidth]{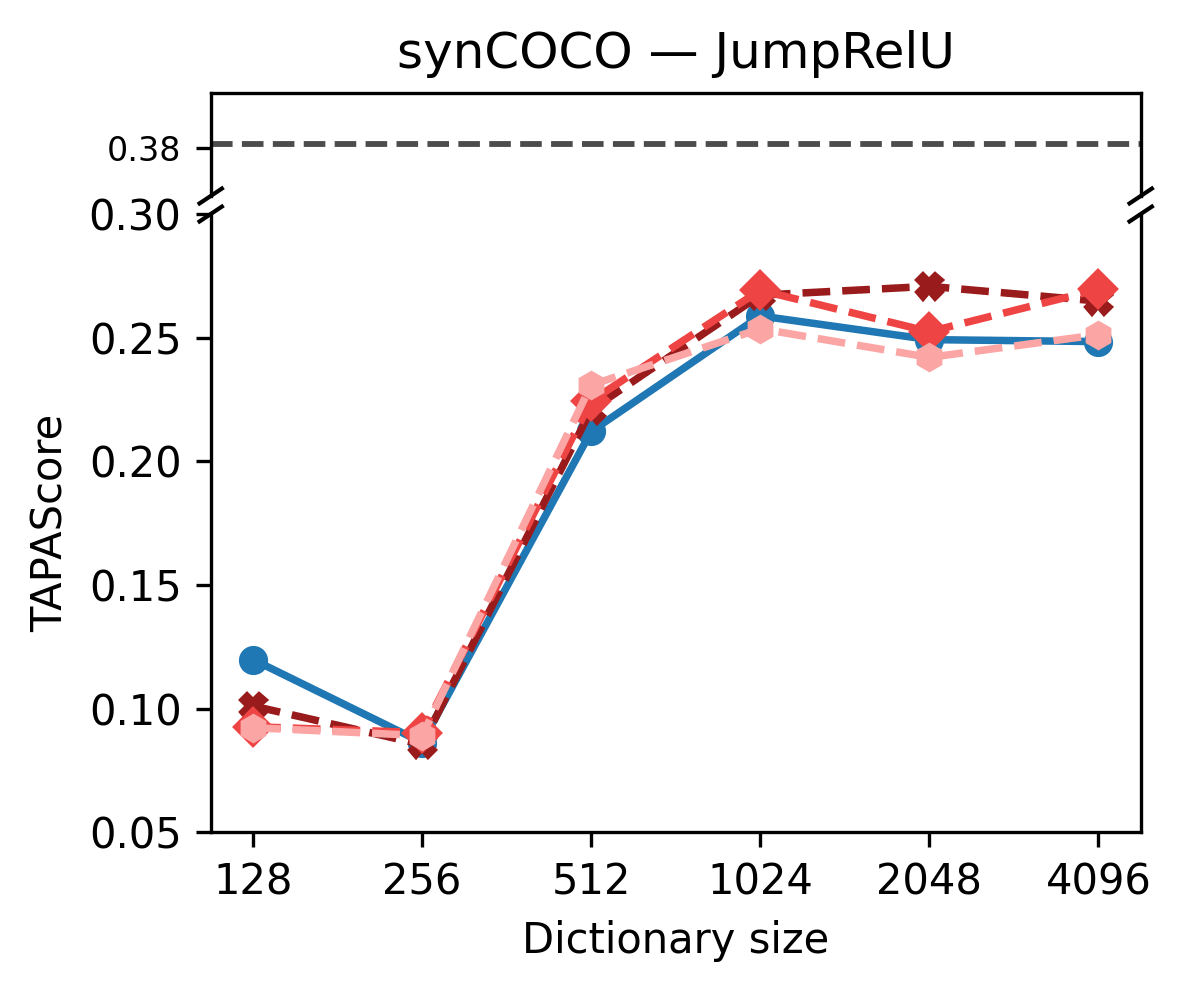}
  \end{subfigure}

  \vspace{-1mm}

  \includegraphics[width=\linewidth]{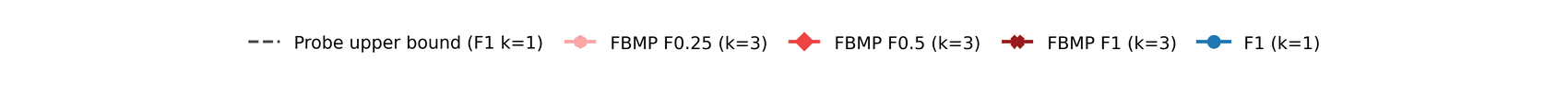}

\vspace{-1.2em}
\caption{$F_1$ $\Delta$MATCHScore on COCO (\textit{top row}) and TAPAScore on synCOCO (\textit{bottom row}) as a function of dictionary size for SAEs trained on DINOV2 embeddings of COCO, across BatchTopK, Matryoshka, TopK and JumpReLU SAE variants (\textit{left to right}) and different matching criteria. The gray dashed line marks the supervised linear-probe upper bound.}
\label{fig:matching_and_tapas_coco_dinov2}
\vspace{-1em}
\end{figure*}
\subsection{Latent to Concept Matching}
\label{app:matching_results}
Fig.~\ref{fig:matching_and_tapas_cub_dinov2} (\textit{top row}) and Fig.~\ref{fig:matching_and_tapas_coco_dinov2} (\textit{top row}) show the $F_1$ $\Delta$MATCHScore as a function of dictionary size for BatchTopK, Matryoshka, TopK, and JumpReLU SAEs trained on DINOv2 embeddings of CUB and COCO, respectively.
Across all four SAE variants on CUB, FBMP consistently outperforms the one-to-one baseline ($F_1, k{=}1$), reinforcing the finding that attributes are better captured by a coalition of complementary latents than by a single unit. The advantage of FBMP is even more pronounced than with CLIP, as one-to-one scores remain low across all dictionary sizes. BatchTopK peaks sharply at dictionary size 512 before declining at larger sizes. Matryoshka exhibits a peak at 256 followed by a sharp decline at 512, after which scores partially recover at larger dictionary sizes without reaching the earlier peak again. TopK exhibits a broadly monotonic increase up to dictionary size 2048, mirroring the pattern observed with CLIP. JumpReLU increases steadily with dictionary size and reaches the highest $\Delta$MATCHScores of all variants at 4096; together with TopK it slightly exceeds the corresponding CLIP peaks, whereas BatchTopK and Matryoshka peak lower than with CLIP.
On COCO, results largely mirror the trends observed in the results with CLIP: FBMP consistently outperforms one-to-one matching, and the relative ordering of matching criteria is preserved. Absolute score differences between SAE variants are smaller, and the overall trends with dictionary size are consistent with the CLIP findings, indicating that matching behavior on COCO is robust to the choice of vision backbone. An exception is JumpReLU, which fails to learn meaningful matchings at the smallest dictionary sizes ($\Delta$MATCHScore close to zero at 128 and 256) and only approaches the other variants from dictionary size 512 onwards.

\begin{figure*}[t]
    \centering
    \begin{subfigure}[t]{0.24\textwidth}
        \centering
        \includegraphics[width=\linewidth]{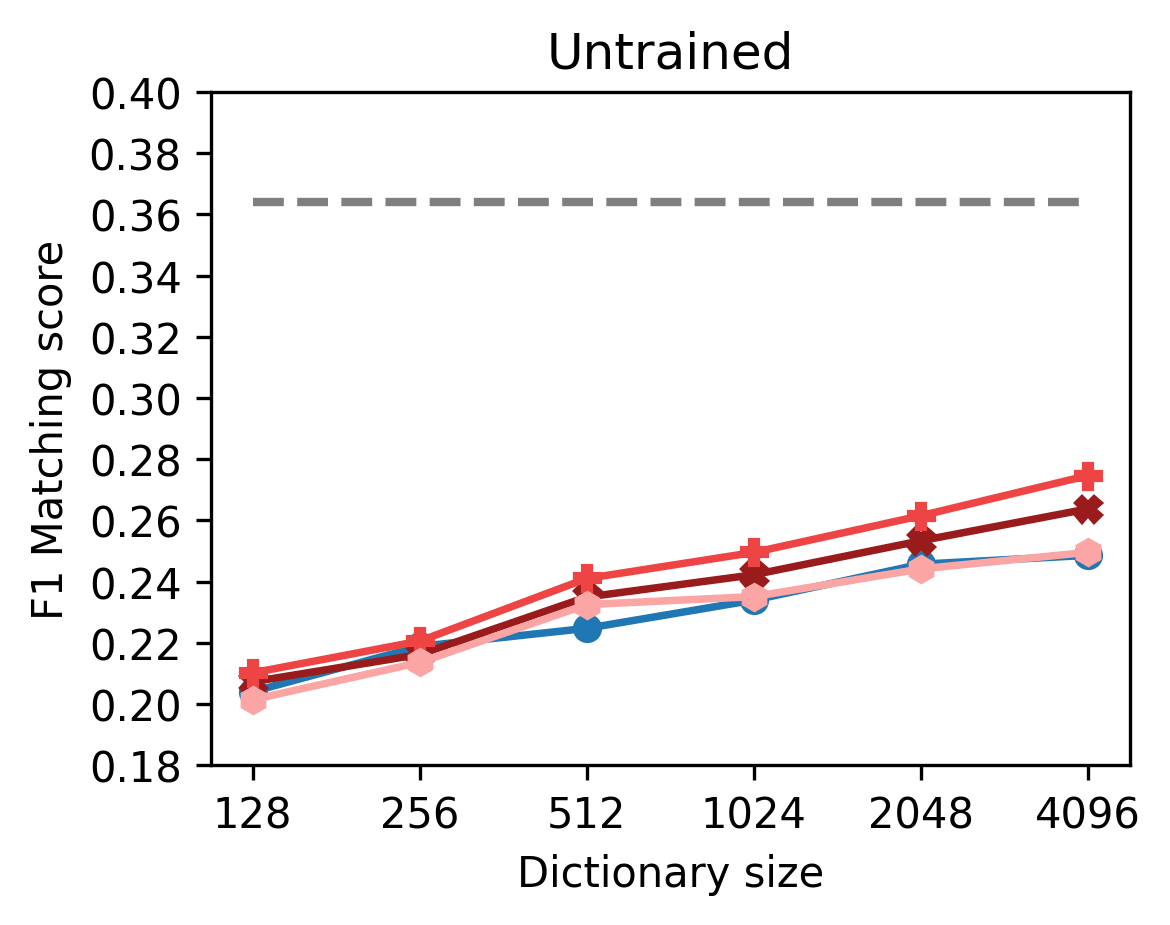}
    \end{subfigure}
    \hfill
    \begin{subfigure}[t]{0.24\textwidth}
        \centering
        \includegraphics[width=\linewidth]{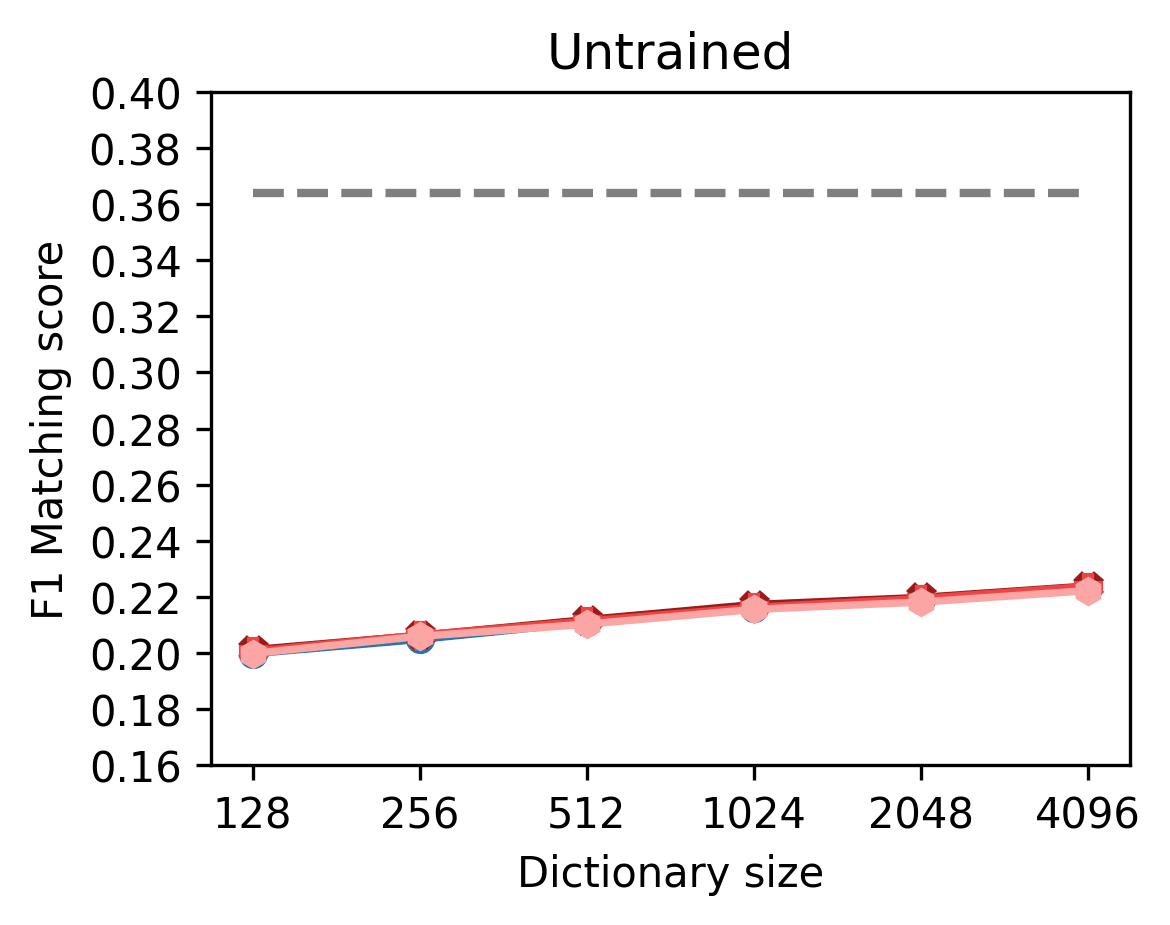}
    \end{subfigure}
    \hfill
    \begin{subfigure}[t]{0.24\textwidth}
        \centering
        \includegraphics[width=\linewidth]{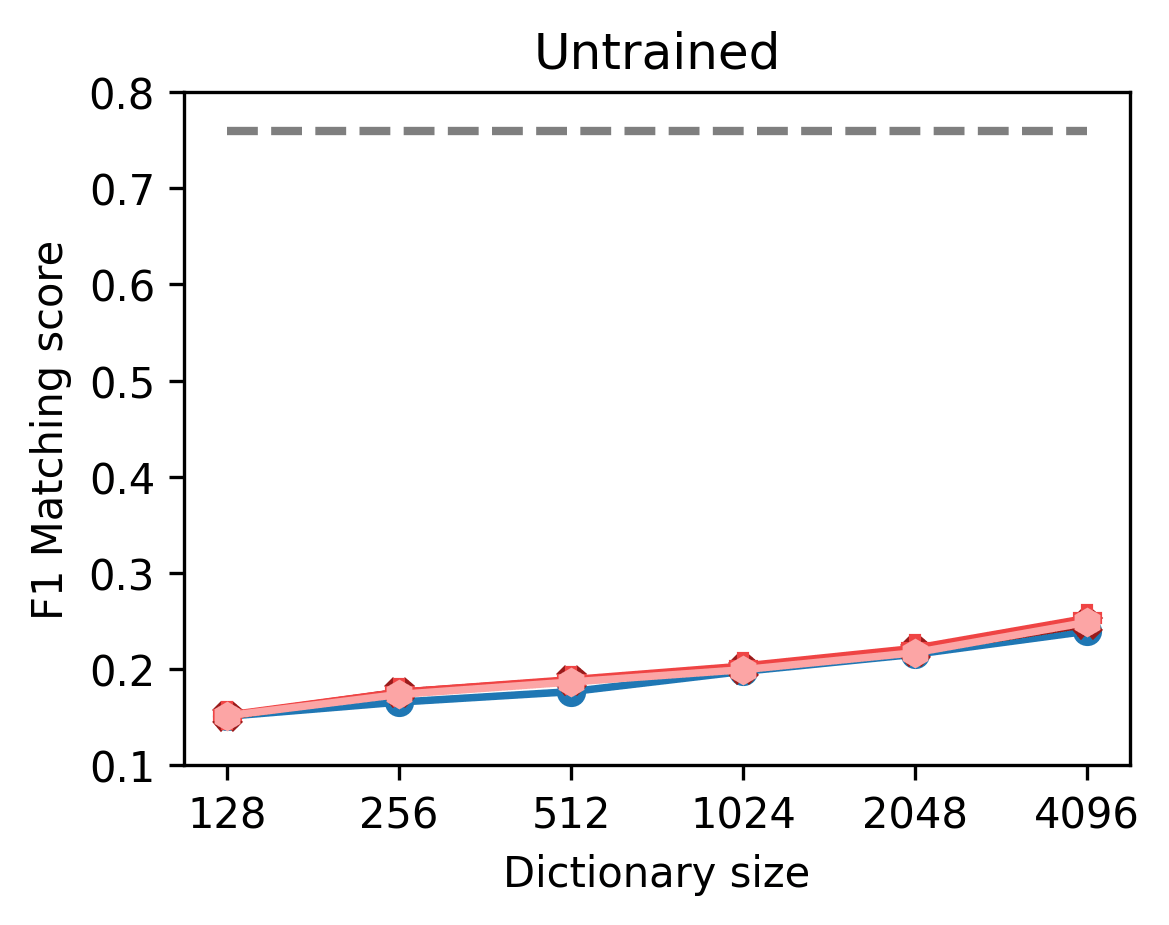}
    \end{subfigure}
        \hfill
    \begin{subfigure}[t]{0.24\textwidth}
        \centering
        \includegraphics[width=\linewidth]{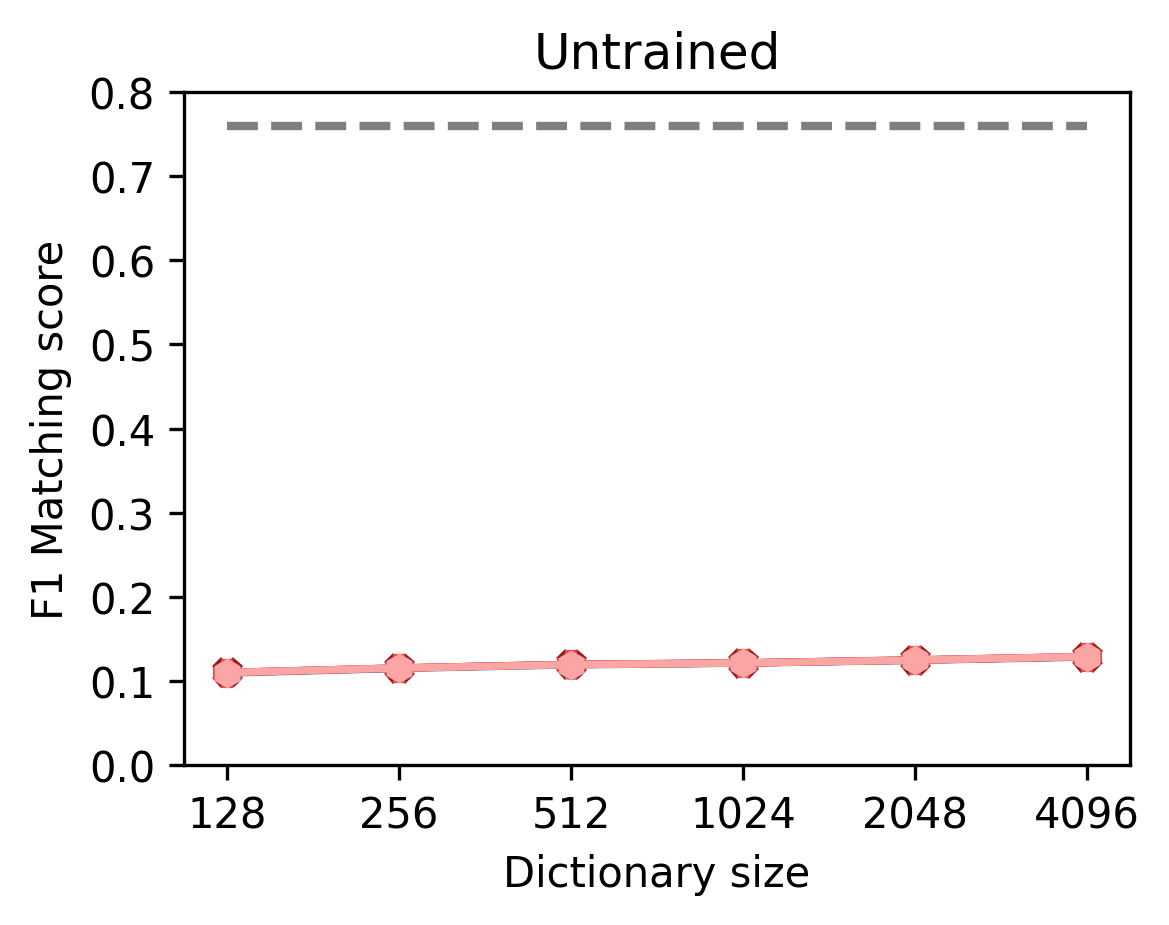}
    \end{subfigure}
    \includegraphics[width=\linewidth]{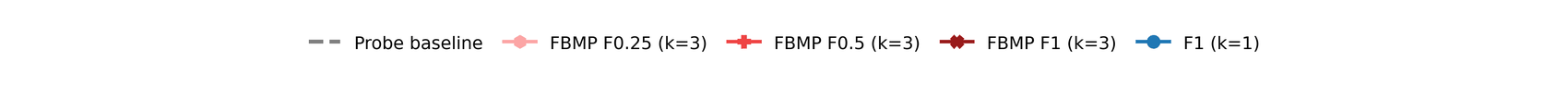}
\caption{Unnormalized $F_1$ matching scores of an untrained TopK SAE as a function of dictionary size, shown for CUB CLIP, CUB DINOv2, COCO CLIP, and COCO DINOv2 (\textit{left to right}).}\label{fig:app_matching_untrained}
\vspace{-2em}
\end{figure*}

\noindent\textbf{Matching Scores of Untrained SAE.} Figure~\ref{fig:app_matching_untrained} validates the monotonic increase of matching scores with dictionary size for an untrained TopK SAE across all four dataset and backbone combinations (CUB CLIP, CUB DINOv2, COCO CLIP, COCO DINOv2). In all cases, the untrained SAE baseline grows steadily as dictionary size increases from 128 to 4096, confirming that larger dictionaries inflate matching scores irrespective of whether meaningful representations have been learned. This inflation arises because a larger pool of candidate latents increases the probability of spurious statistical alignment between random activations and ground-truth attribute annotations. The effect is consistent across both FBMP variants and one-to-one matching, and is observed on both datasets and both backbones. This motivates the use of $\Delta$MATCHScore, which subtracts the untrained baseline to isolate genuine improvements in semantic alignment from dictionary-size-induced inflation.

\noindent\textbf{Matching over coalition sizes ($k$).}
\label{app:subsec:tapas_over_k}
%
  \begin{figure}[p]
    \centering
    \vspace{-2em}
    \begin{subfigure}[t]{0.24\textwidth}
      \includegraphics[width=\linewidth]{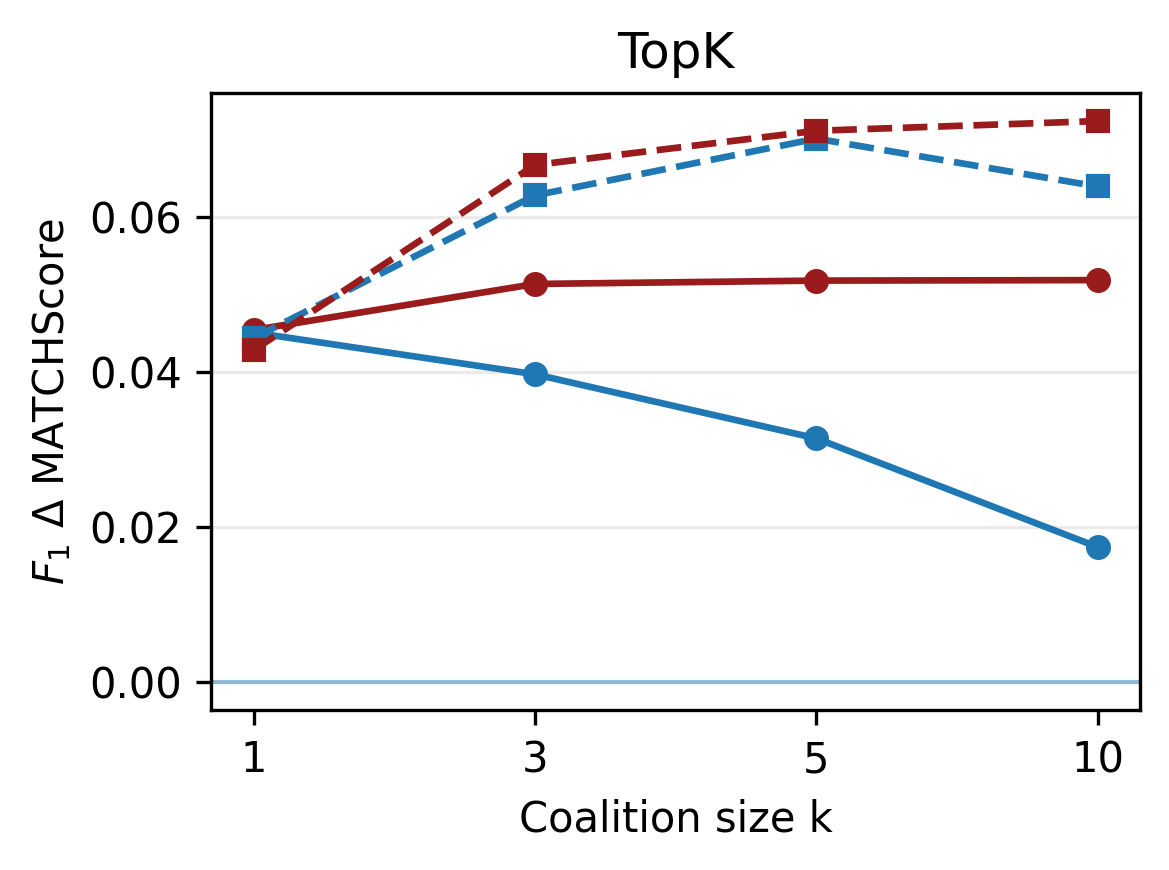}
    \end{subfigure}\hfill
    \begin{subfigure}[t]{0.24\textwidth}
      \includegraphics[width=\linewidth]{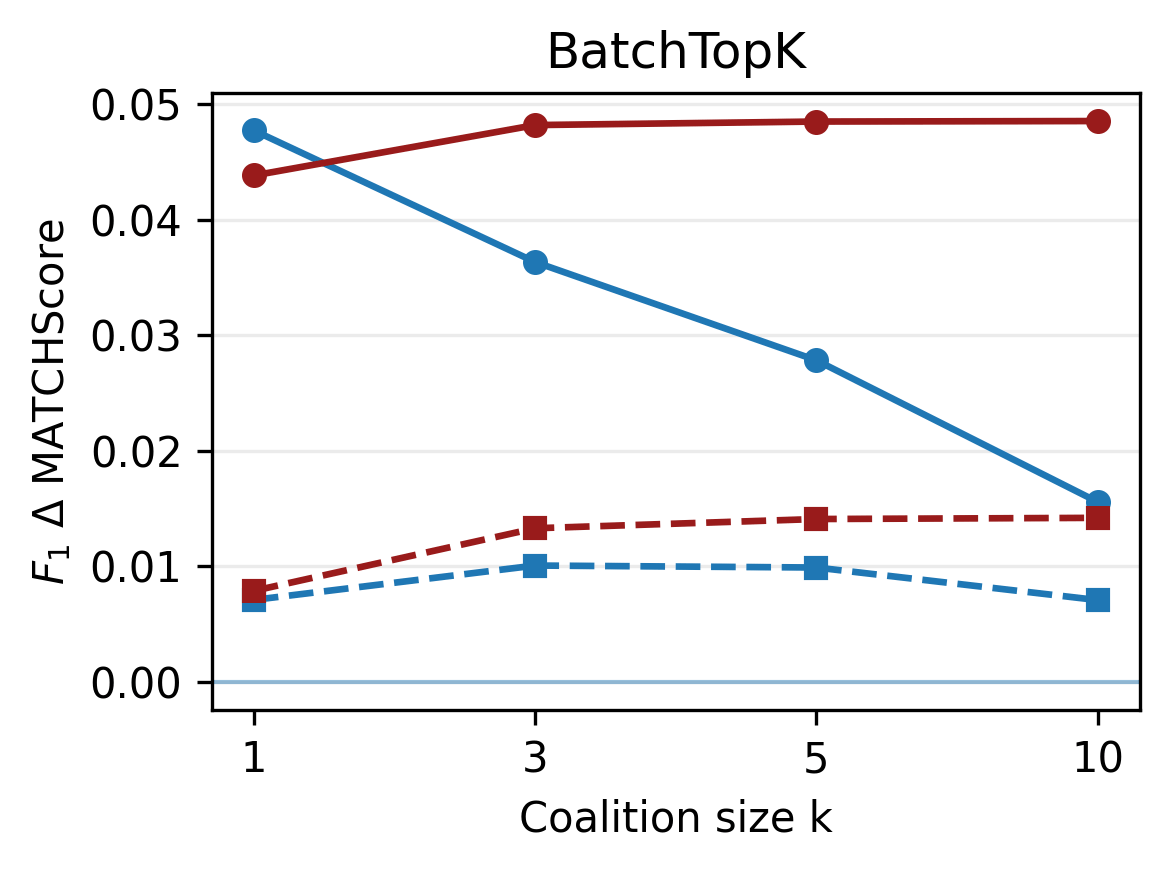}
    \end{subfigure}\hfill
    \begin{subfigure}[t]{0.24\textwidth}
      \includegraphics[width=\linewidth]{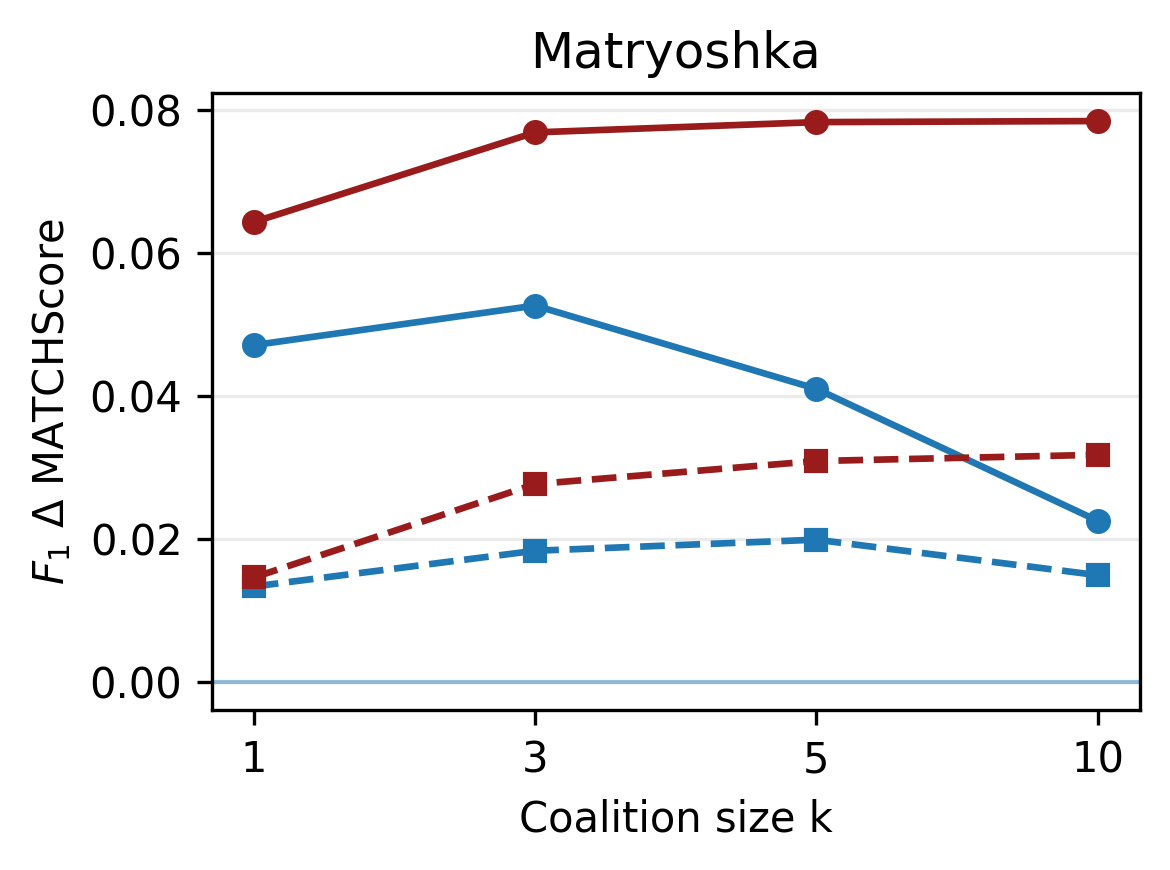}
    \end{subfigure}\hfill
    \begin{subfigure}[t]{0.24\textwidth}
      \includegraphics[width=\linewidth]{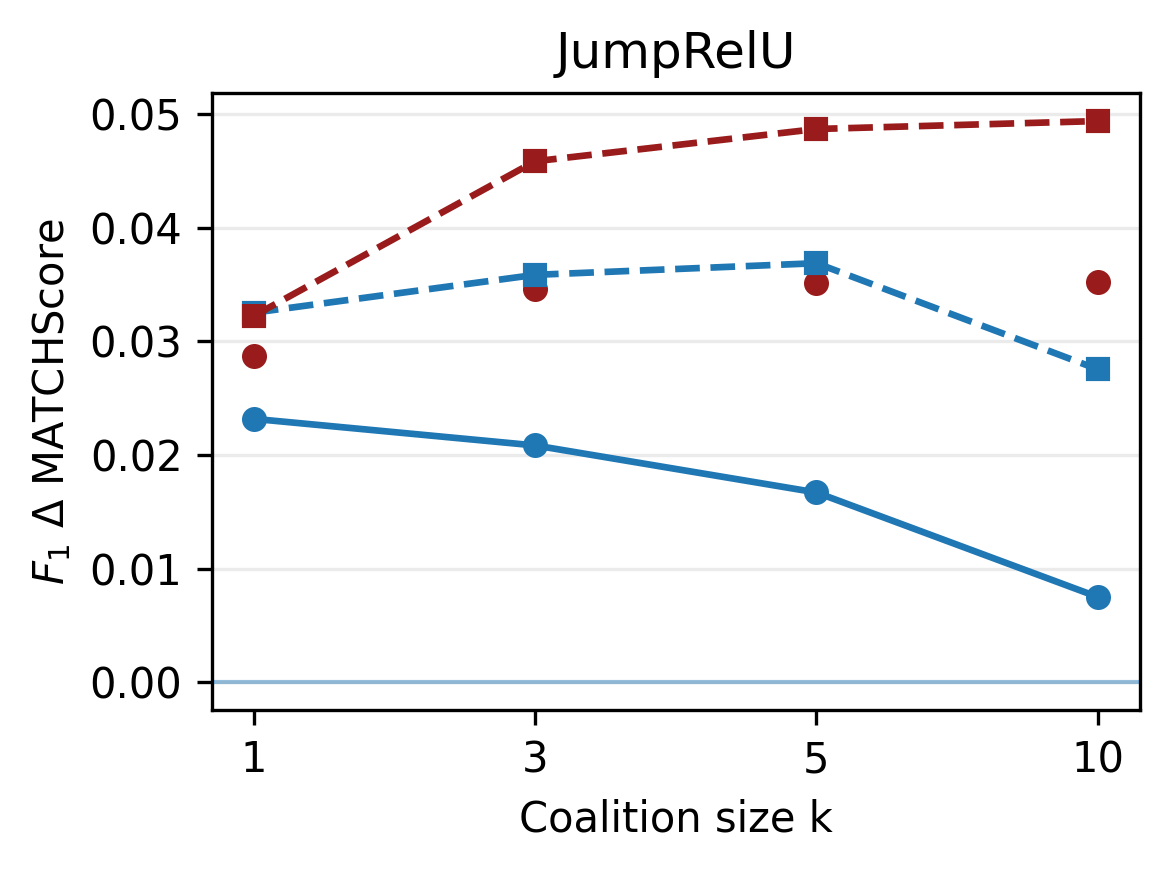}
    \end{subfigure}
    \\[6pt]
    \begin{subfigure}[t]{0.24\textwidth}
      \includegraphics[width=\linewidth]{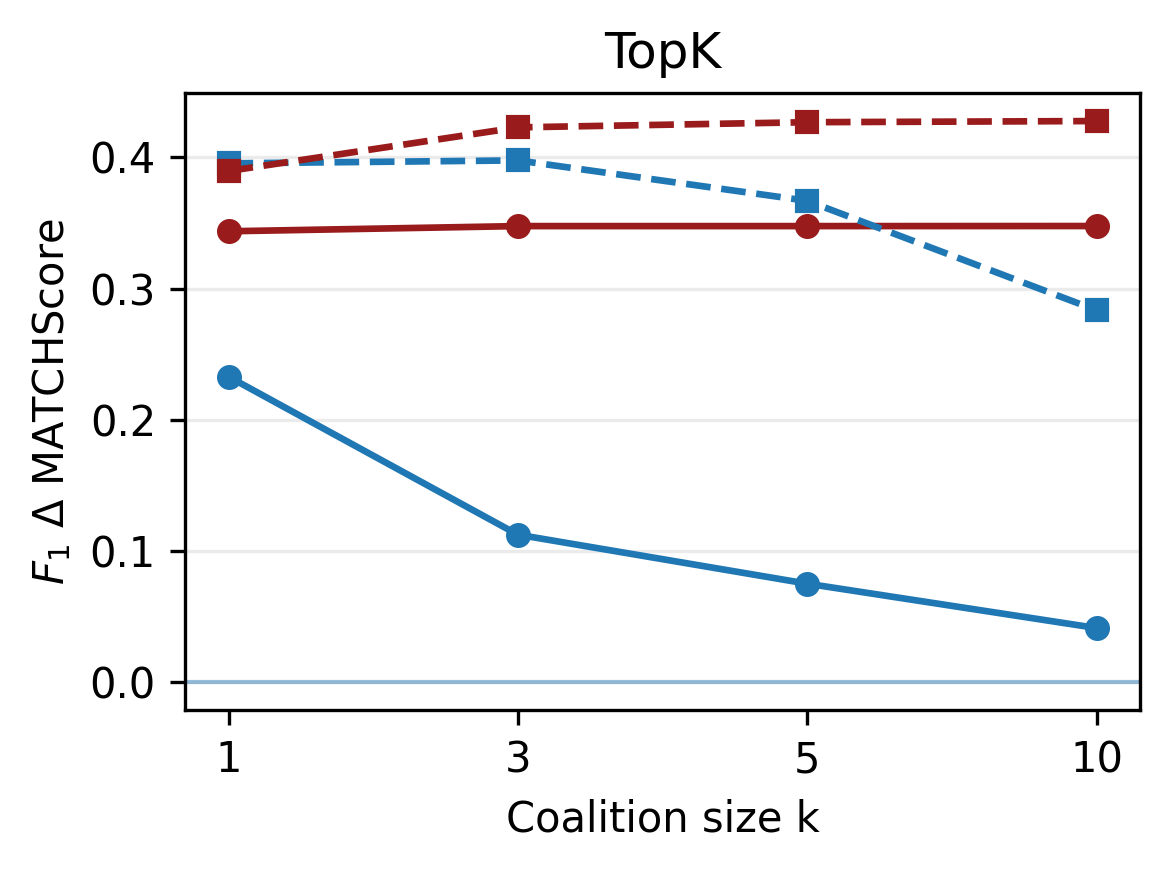}
    \end{subfigure}\hfill
    \begin{subfigure}[t]{0.24\textwidth}
      \includegraphics[width=\linewidth]{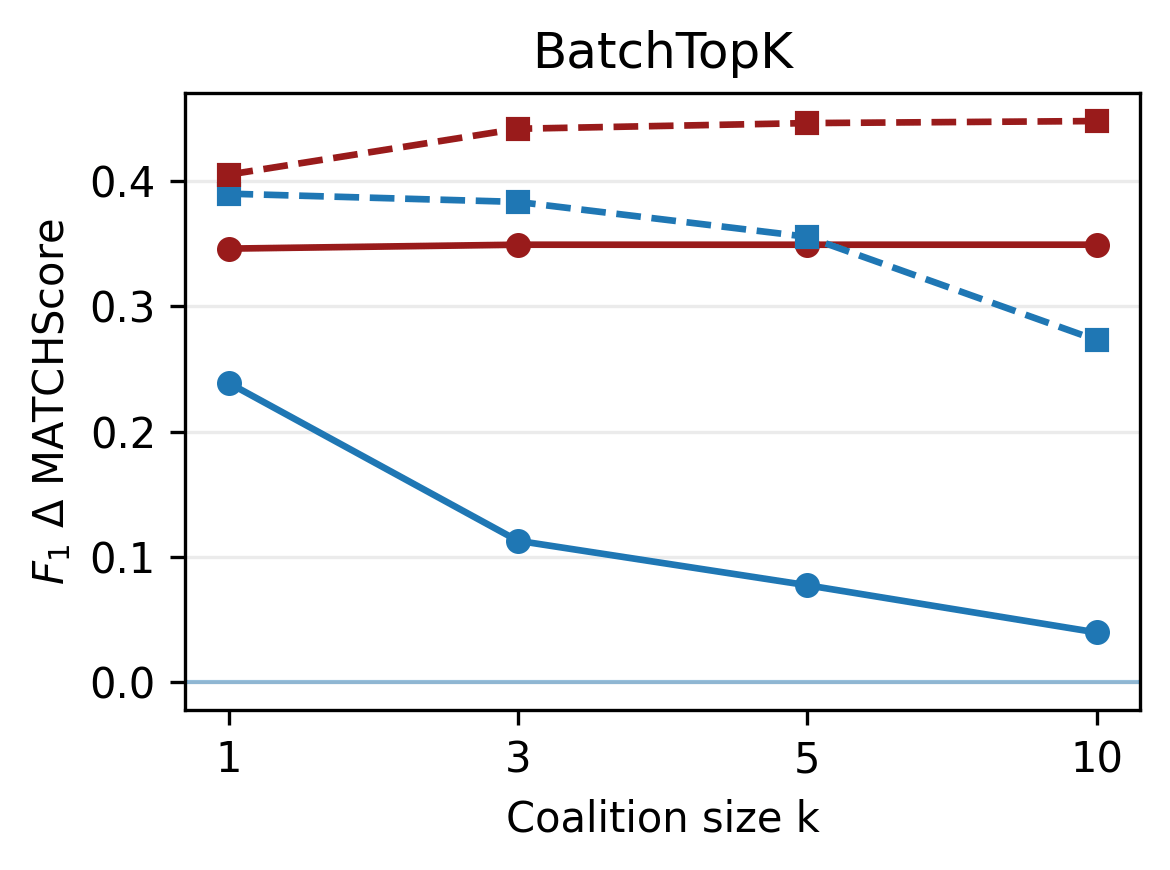}
    \end{subfigure}\hfill
    \begin{subfigure}[t]{0.24\textwidth}
      \includegraphics[width=\linewidth]{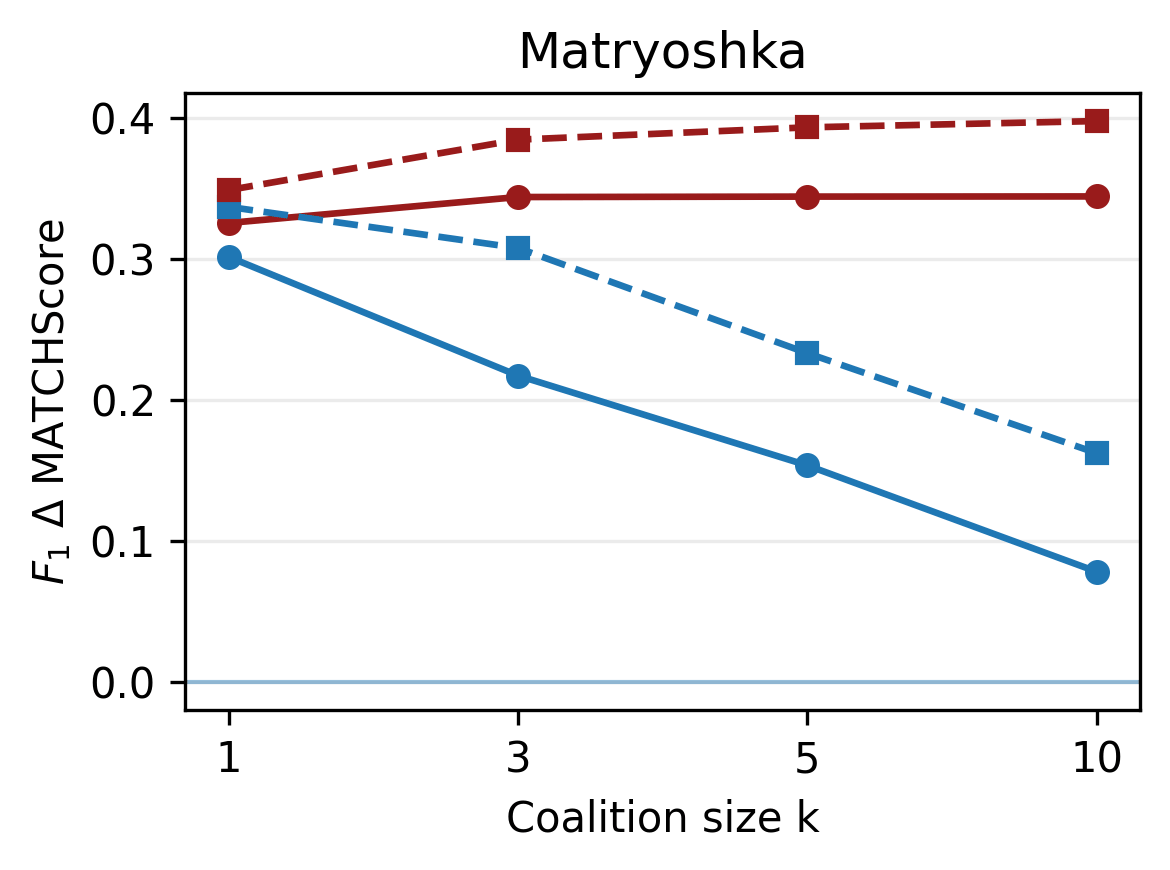}
    \end{subfigure}\hfill
    \begin{subfigure}[t]{0.24\textwidth}
      \includegraphics[width=\linewidth]{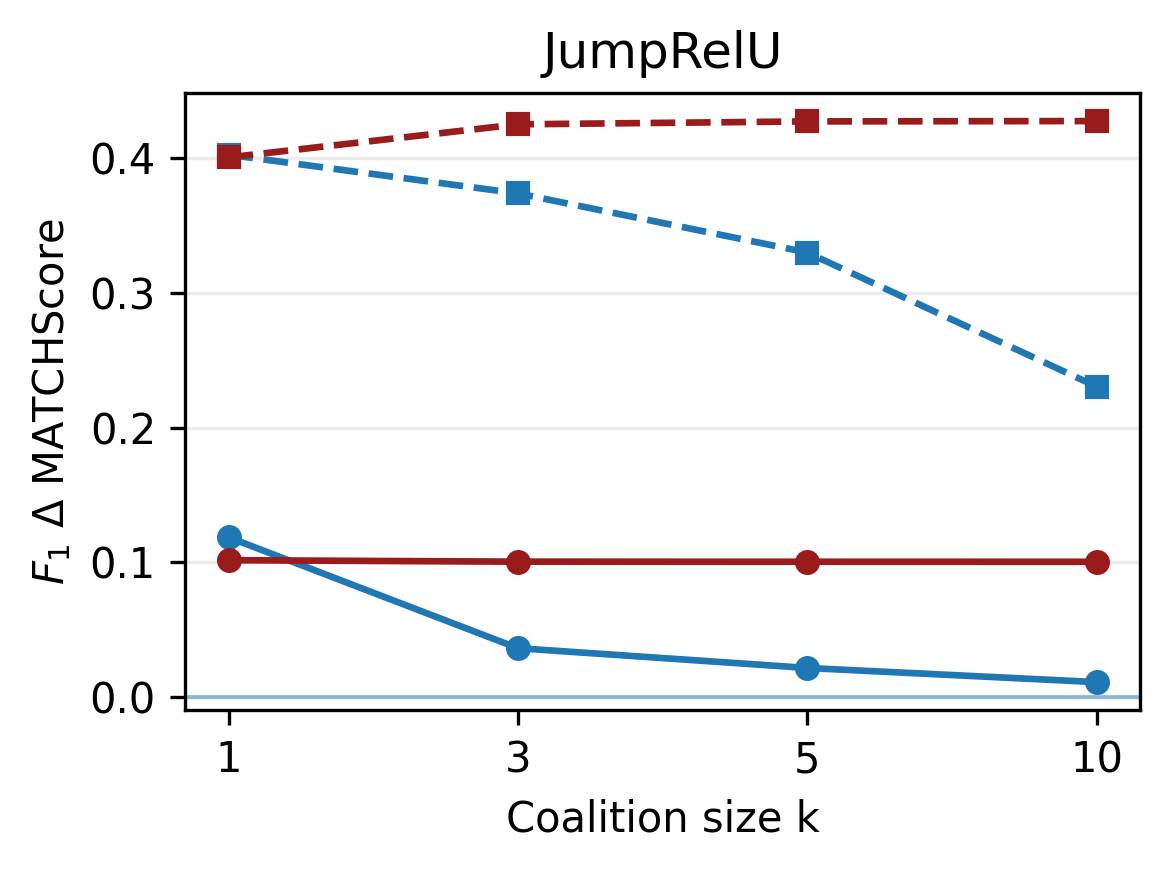}
    \end{subfigure}
    \\[4pt]
    \includegraphics[width=0.7\linewidth]{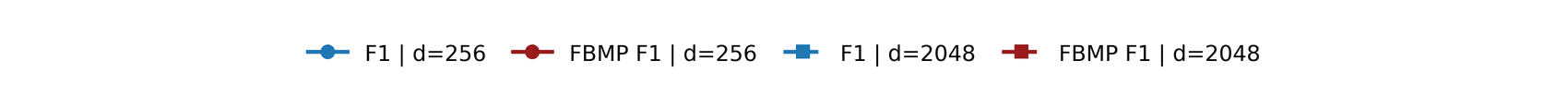}
    \caption{$\Delta$MATCHScore as a function of coalition size $k$ for all SAE families on CLIP,
             dict size $d \in \{256, 2048\}$. Top row: CUB. Bottom row: COCO.}
    \label{fig:appendix_coalition_matching}
    \vspace{-2em}
  \end{figure}
Figure~\ref{fig:appendix_coalition_matching} shows $\Delta$MATCHScore as a function of coalition size $k$ for BatchTopK, Matryoshka, TopK, and JumpReLU SAEs trained on CLIP embeddings, at dictionary sizes $d{=}256$ (solid) and $d{=}2048$ (dashed).
Across both datasets and all SAE variants, FBMP matches or outperforms naive top-$k$ matching with $F_1$ at every coalition size $k$. While FBMP improves or remains stable as $k$ increases, the naive $F_1$ matching score degrades steadily, as additional latents are added without regard for their relevance. For example, for the Matryoshka SAE trained on COCO embeddings, the naive score drops to $0.08$ ($d{=}256$) and $0.16$ ($d{=}2048$) at $k{=}10$, while FBMP remains stable at $0.34$ and $0.40$, respectively.
This confirms that the sequential residual selection of FBMP is essential for maintaining matching quality at larger coalition sizes. Based on these results, we further select $k{=}3$ as the default coalition size for FBMP, as scores plateau beyond this point.

\subsection{Functional Validation of Concept Alignment}
\label{app:tapas_results}
Fig.~\ref{fig:matching_and_tapas_cub_dinov2} (\textit{bottom row}) and Fig.~\ref{fig:matching_and_tapas_coco_dinov2} (\textit{bottom row}) report TAPAScore on synCUB and synCOCO for SAEs trained on DINOv2 embeddings.
From Fig.~\ref{fig:matching_and_tapas_cub_dinov2} (\textit{bottom row}) it can be seen that the TAPAScore results for synCUB differ markedly from the corresponding CLIP results (main paper). Whereas the CLIP results often showed a peak at intermediate dictionary sizes followed by a sharp decline, this degradation is considerably less pronounced for the DINOv2 results. BatchTopK remains comparatively stable across all dictionary sizes. TopK peaks at dictionary sizes 256--512 and then decreases, qualitatively consistent with CLIP but with a shallower drop for the FBMP criteria, while its one-to-one criterion degrades sharply. JumpReLU similarly peaks at 256--512 before declining moderately. Matryoshka, however, exhibits a notably different pattern: TAPAScore continues to improve with dictionary size and achieves its highest values at the largest dictionary sizes, consistent with the recovery of its matching scores in the top row.
The TAPAScore results for synCOCO with DINOv2 are more stable across dictionary sizes compared to their CLIP counterparts, with less pronounced divergence between matching and perturbation alignment at larger dictionary sizes. Unlike the CLIP results, where overcompleteness led to a clear TAPAScore decline for BatchTopK and TopK, DINOv2 results exhibit only a mild decline at the largest dictionary sizes, suggesting that the backbone choice influences the degree to which overcompleteness reduces causal alignment.

\subsection{Binary vs.\ Magnitude-aware Matching}
\label{app:nnomp_results}

\begin{figure*}[t]
  \centering
  \vspace{-2em}
    \begin{subfigure}[t]{0.24\textwidth}
      \centering
      \includegraphics[width=\linewidth]{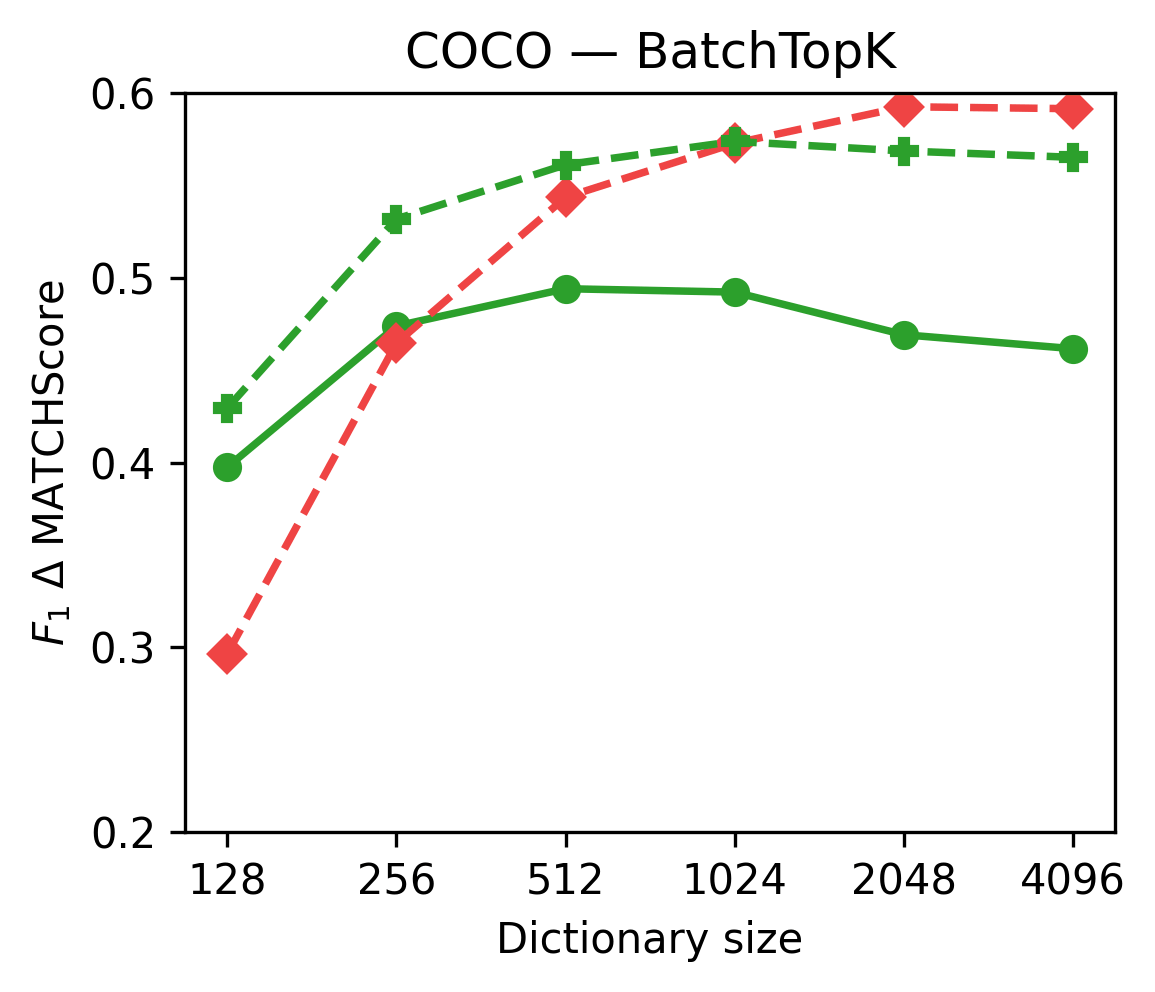}
    \end{subfigure}
    \hfill
    \begin{subfigure}[t]{0.24\textwidth}
      \centering
      \includegraphics[width=\linewidth]{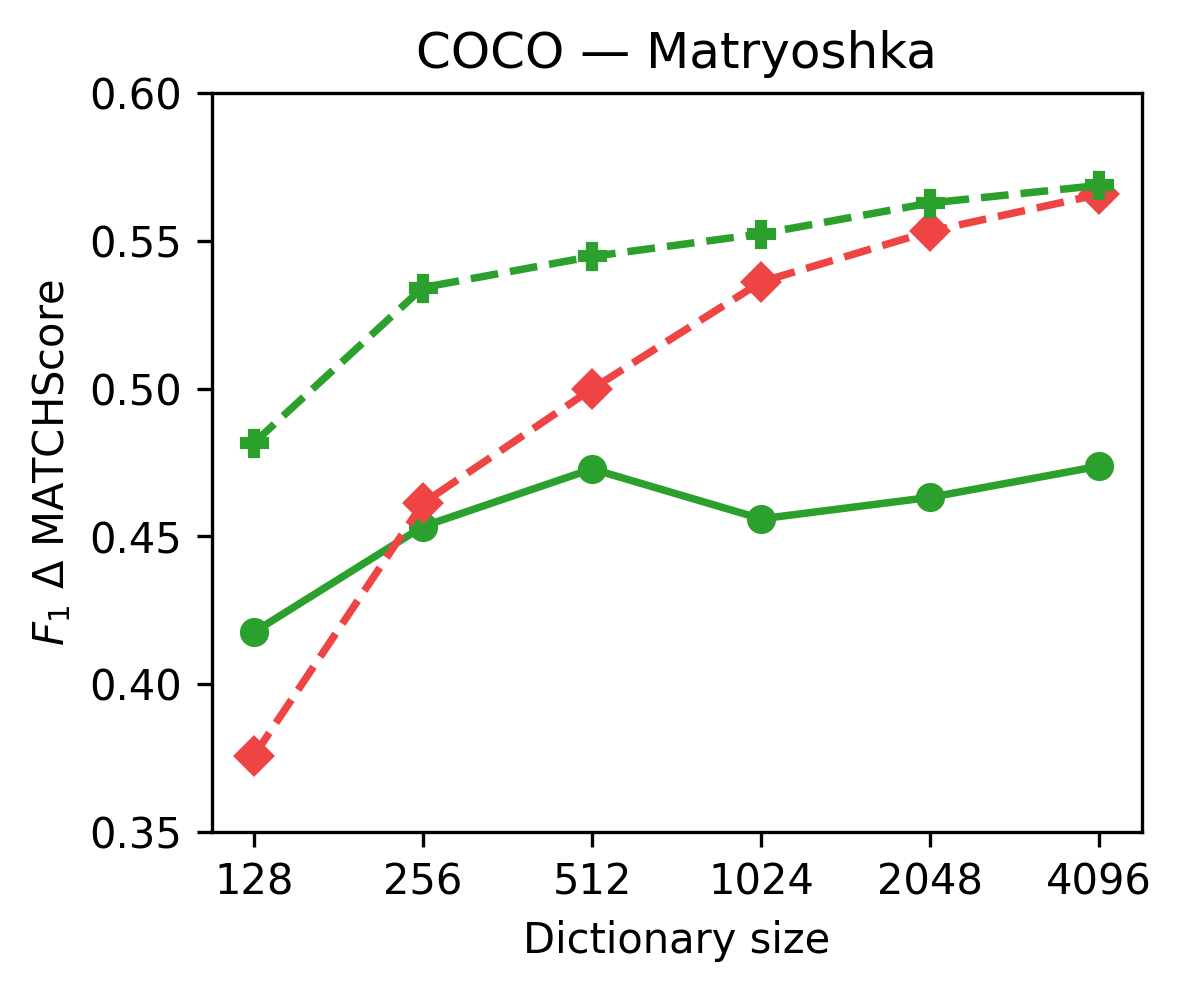}
    \end{subfigure}
    \hfill
    \begin{subfigure}[t]{0.24\textwidth}
      \centering
      \includegraphics[width=\linewidth]{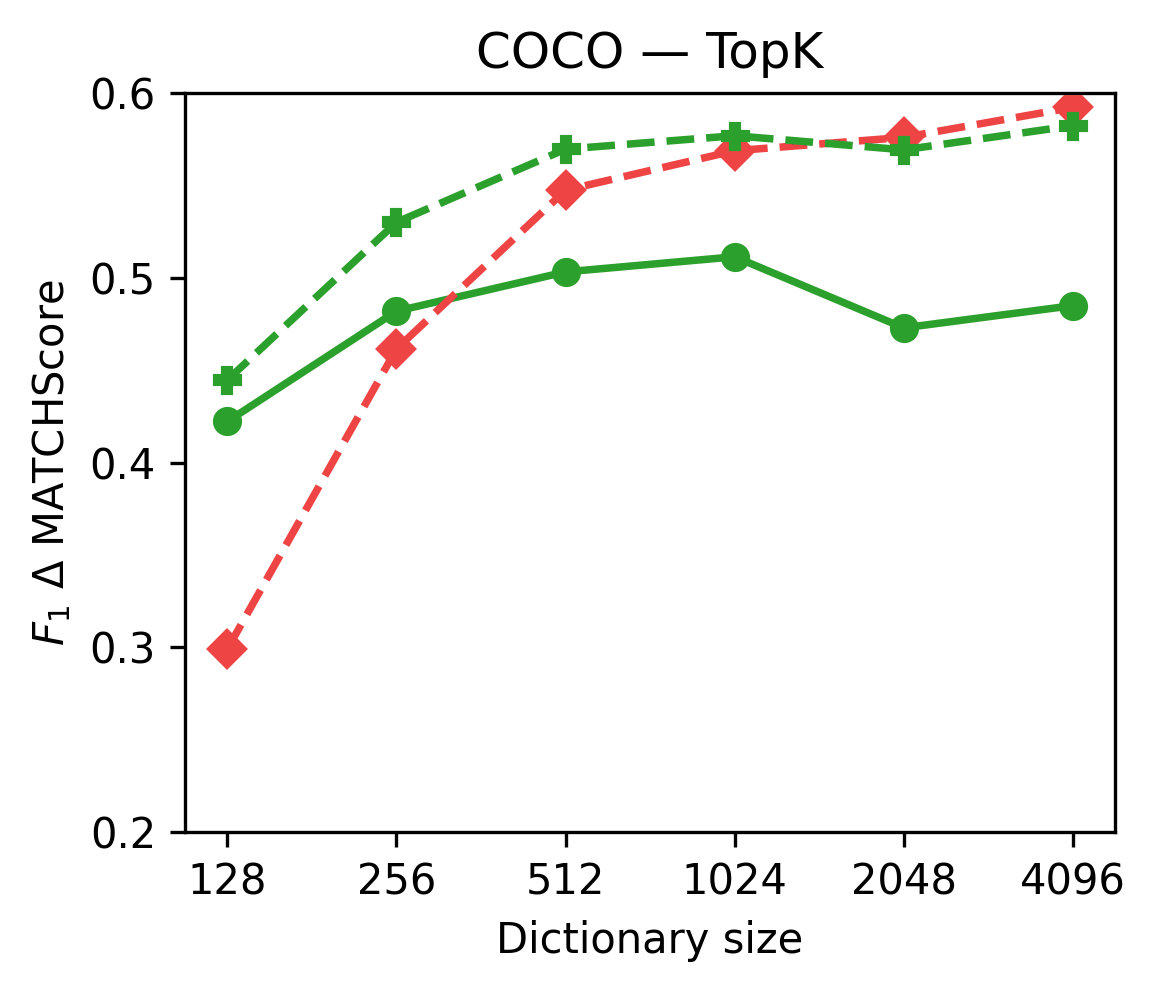}
    \end{subfigure}
    \hfill
    \begin{subfigure}[t]{0.24\textwidth}
      \centering
      \includegraphics[width=\linewidth]{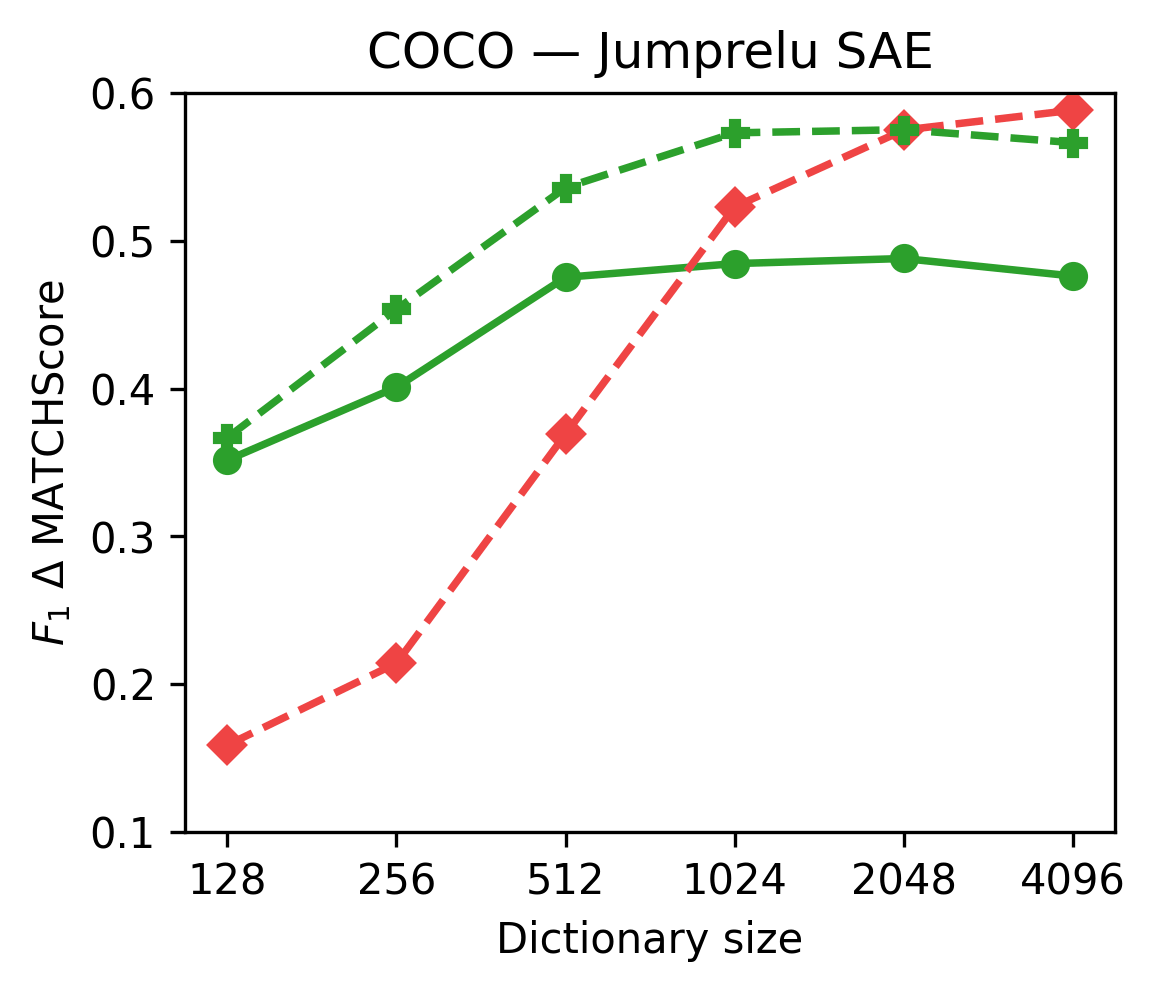}
    \end{subfigure}

    \vspace{2mm}

    \begin{subfigure}[t]{0.24\textwidth}
      \centering
      \includegraphics[width=\linewidth]{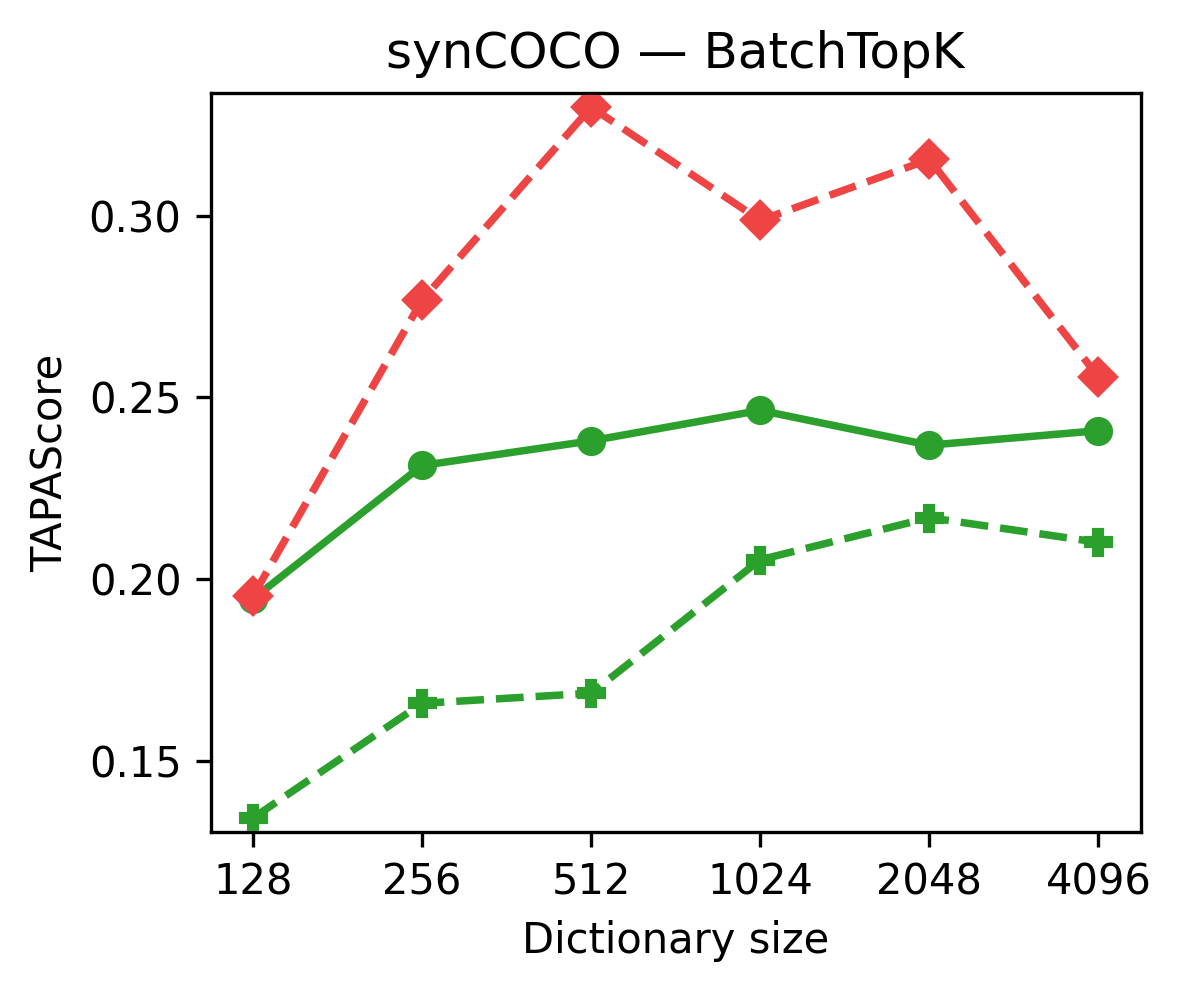}
    \end{subfigure}
    \hfill
    \begin{subfigure}[t]{0.24\textwidth}
      \centering
      \includegraphics[width=\linewidth]{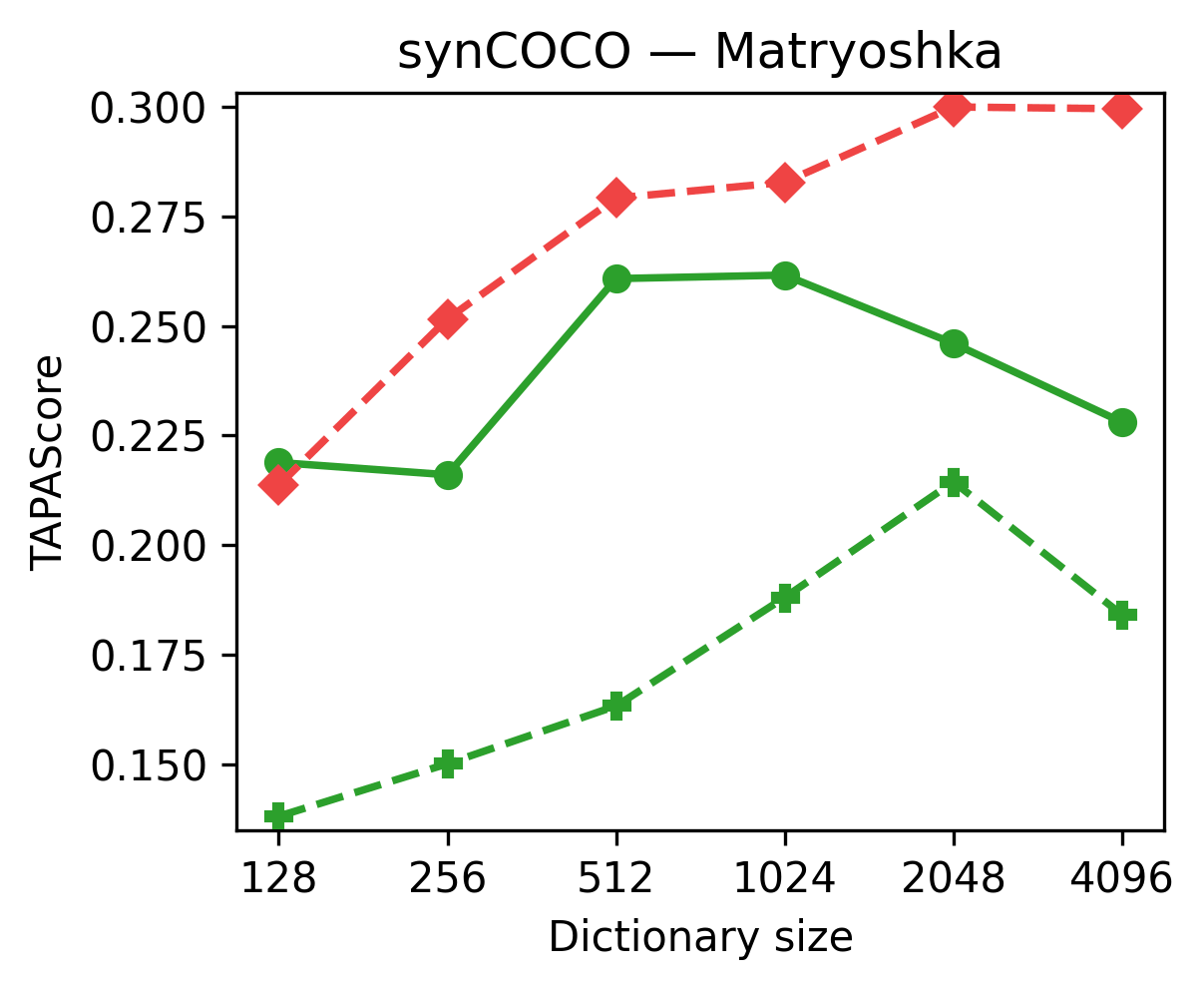}
    \end{subfigure}
    \hfill
    \begin{subfigure}[t]{0.24\textwidth}
      \centering
      \includegraphics[width=\linewidth]{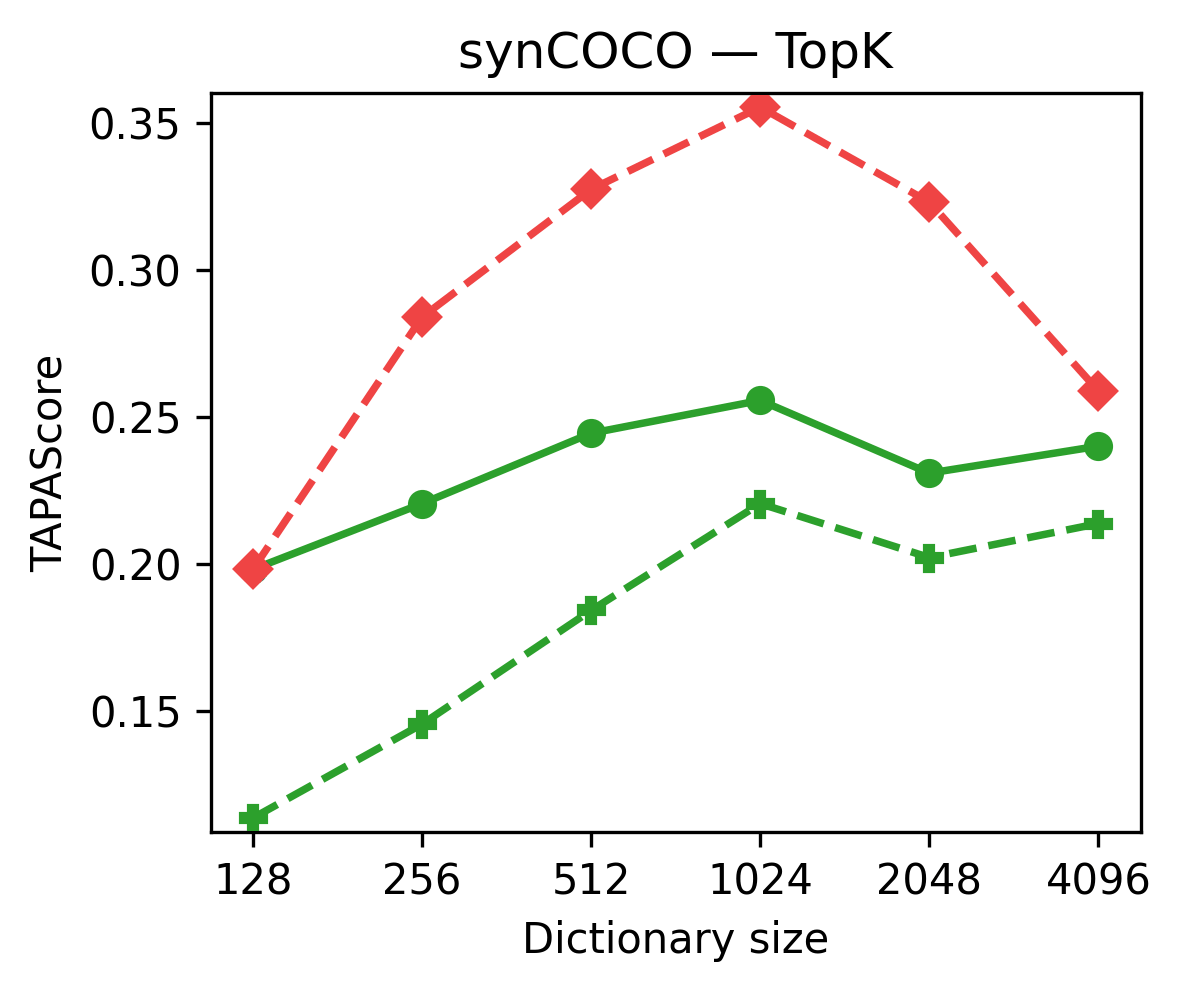}
    \end{subfigure}
    \hfill
    \begin{subfigure}[t]{0.24\textwidth}
      \centering
      \includegraphics[width=\linewidth]{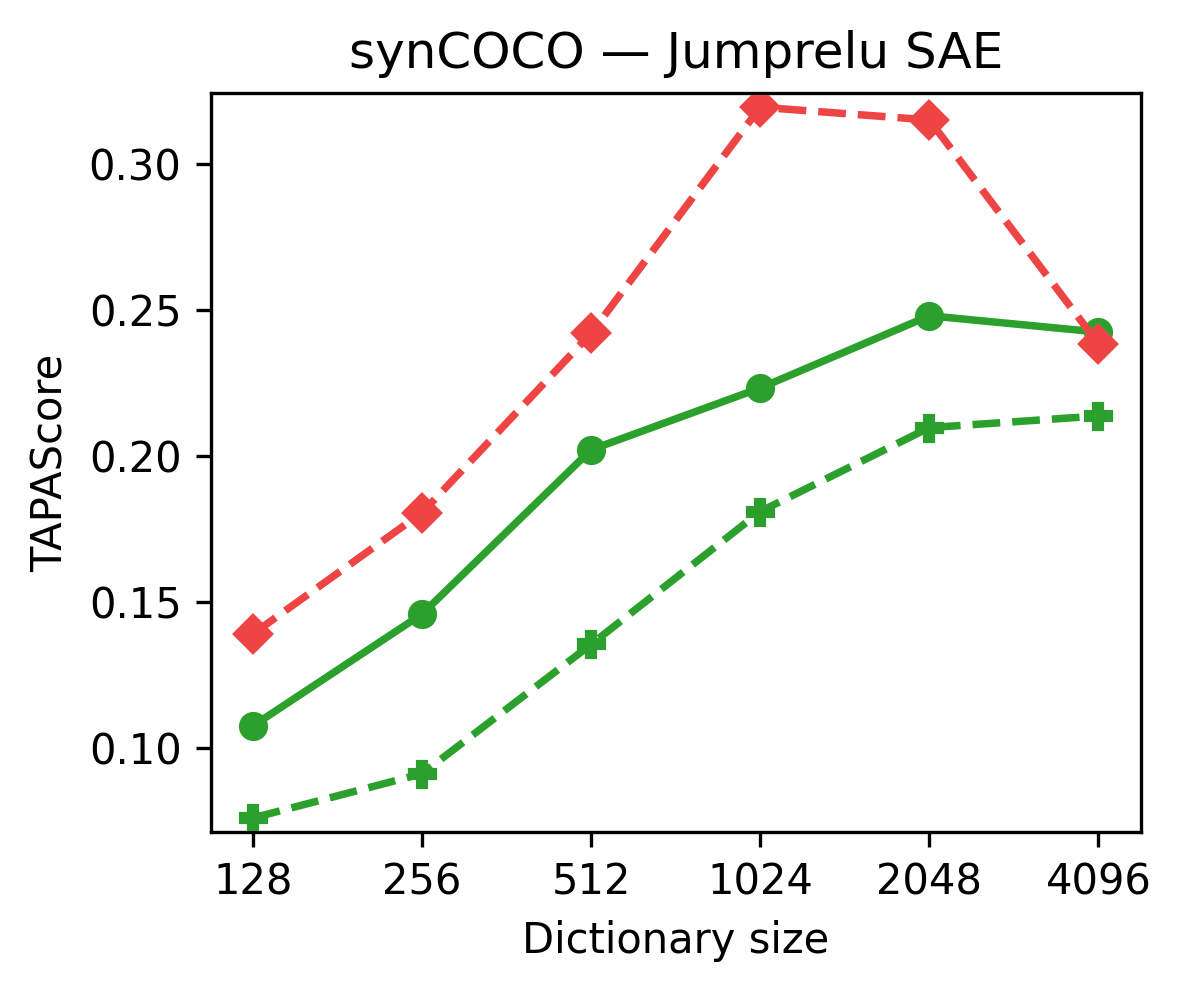}
    \end{subfigure}

    \vspace{1mm}
  
    \includegraphics[width=\linewidth]{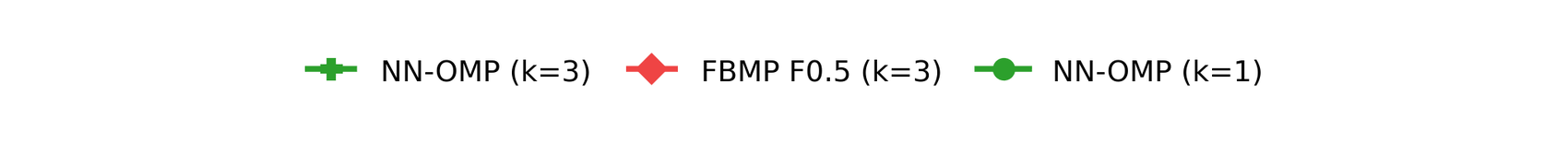}

  \vspace{-1.2em}
  \caption{$F_1$ $\Delta$MATCHScore on COCO (\textit{top row}) and TAPAScore on synCOCO (\textit{bottom row}) comparing NN-OMP and FBMP F0.5 matching criteria for SAEs trained on CLIP embeddings, across BatchTopK, Matryoshka, TopK and JumpReLU SAE variants (\textit{left to right}).}
  \label{fig:appendix_nnomp_coco}
  \vspace{-1em}
  \end{figure*}
\begin{figure*}[t]
  \centering
  \vspace{-2em}
    \begin{subfigure}[t]{0.24\textwidth}
      \centering
      \includegraphics[width=\linewidth]{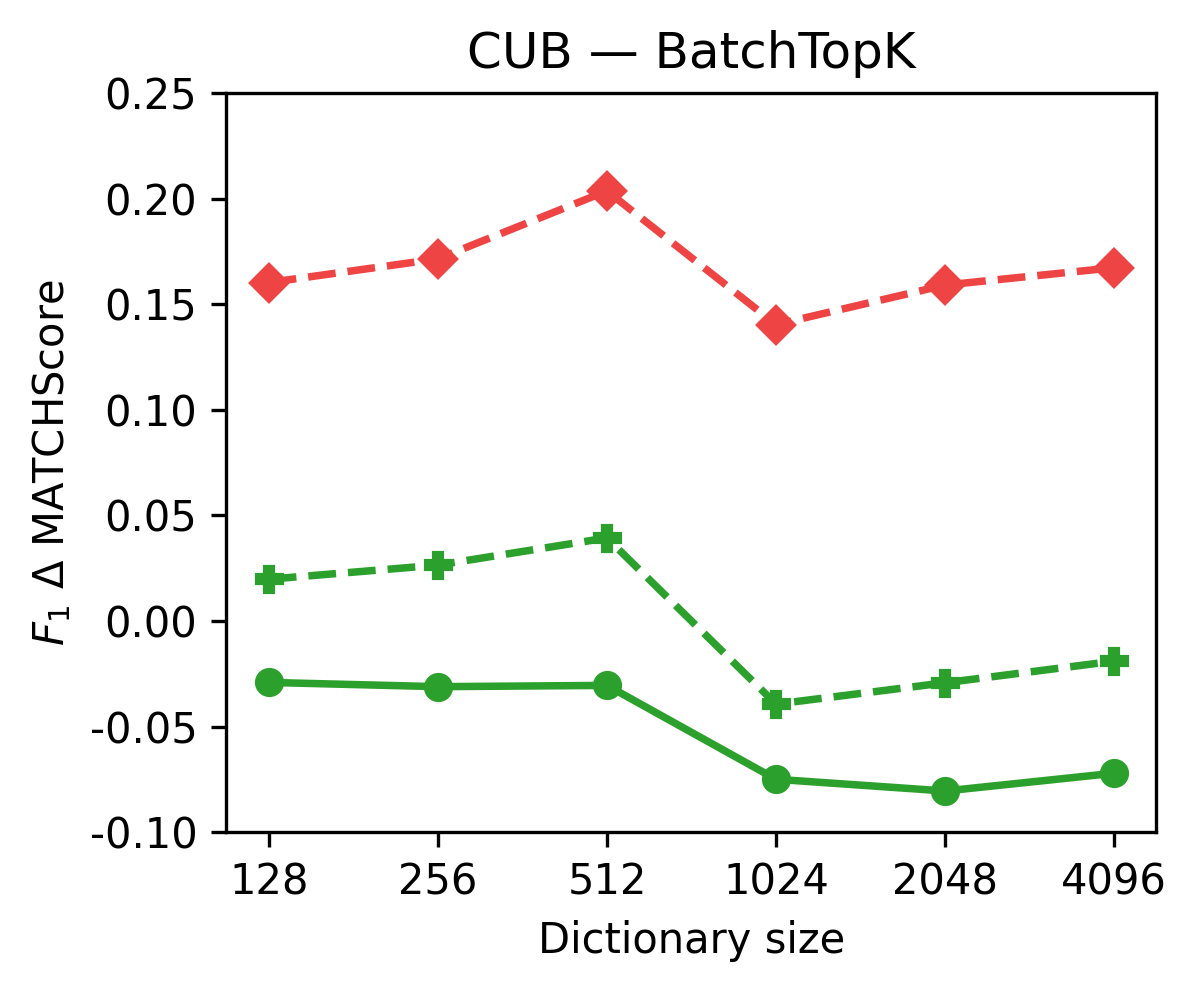}
    \end{subfigure}
    \hfill
    \begin{subfigure}[t]{0.24\textwidth}
      \centering
      \includegraphics[width=\linewidth]{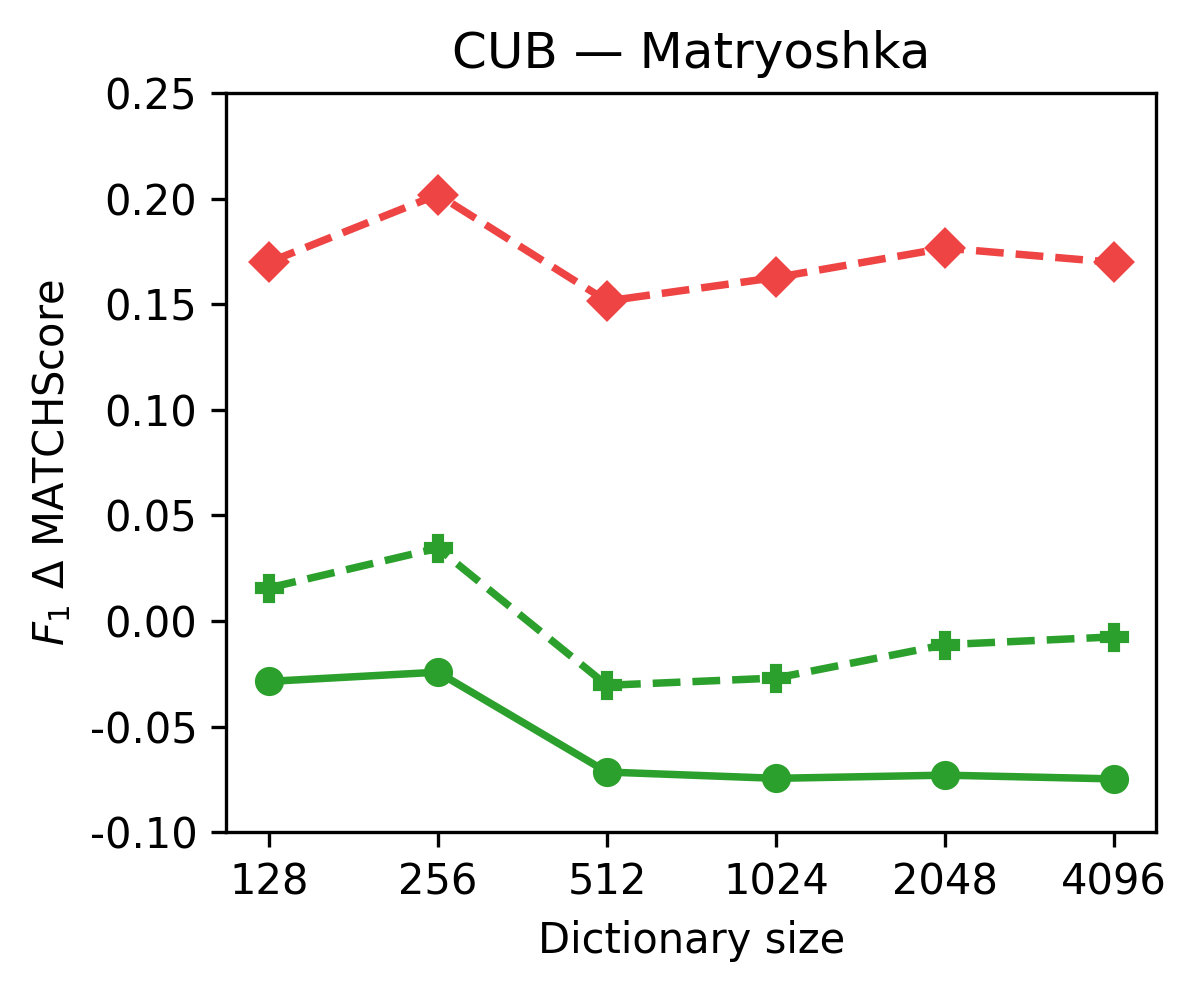}
    \end{subfigure}
    \hfill
    \begin{subfigure}[t]{0.24\textwidth}
      \centering
      \includegraphics[width=\linewidth]{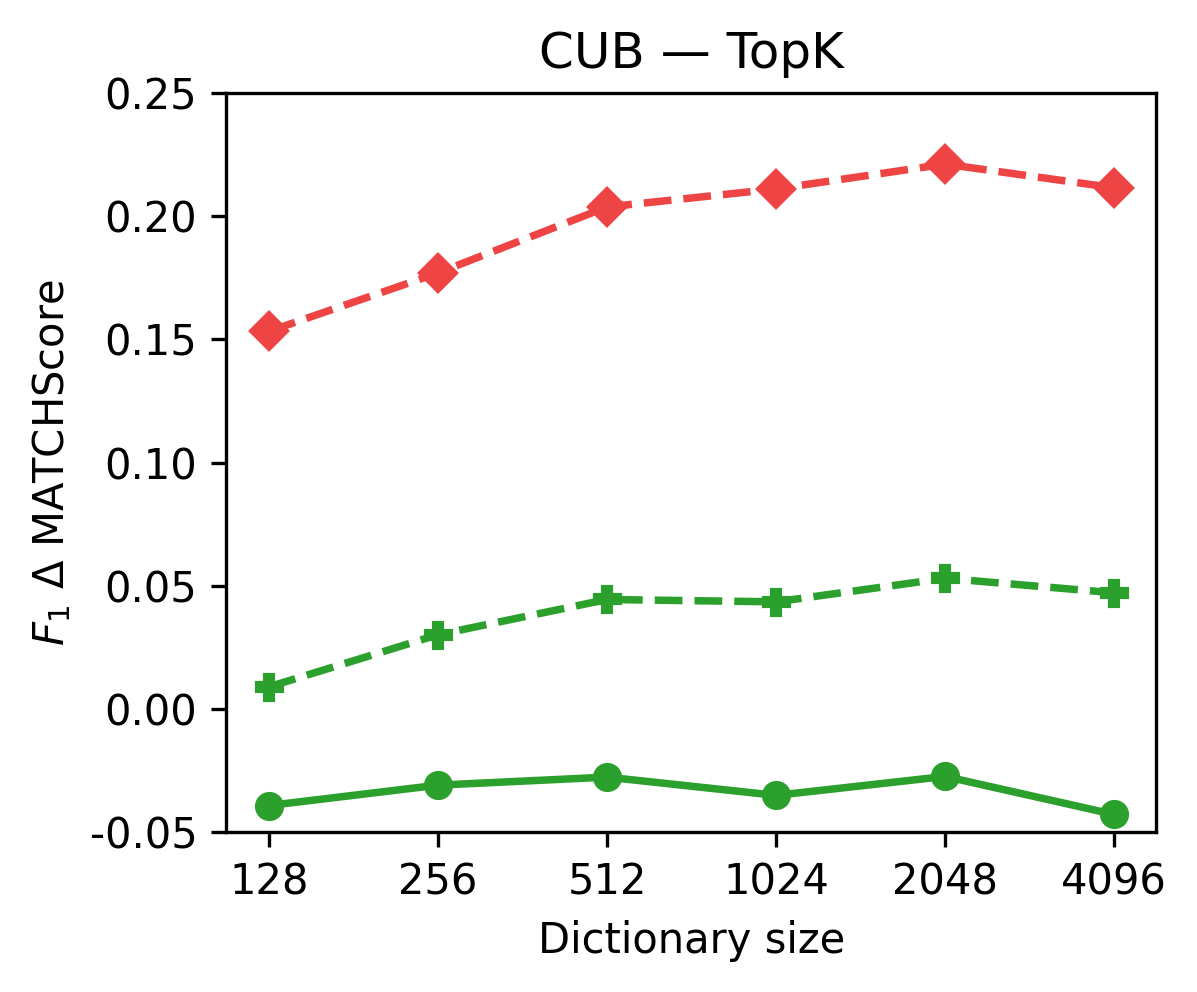}
    \end{subfigure}
    \hfill
    \begin{subfigure}[t]{0.24\textwidth}
      \centering
      \includegraphics[width=\linewidth]{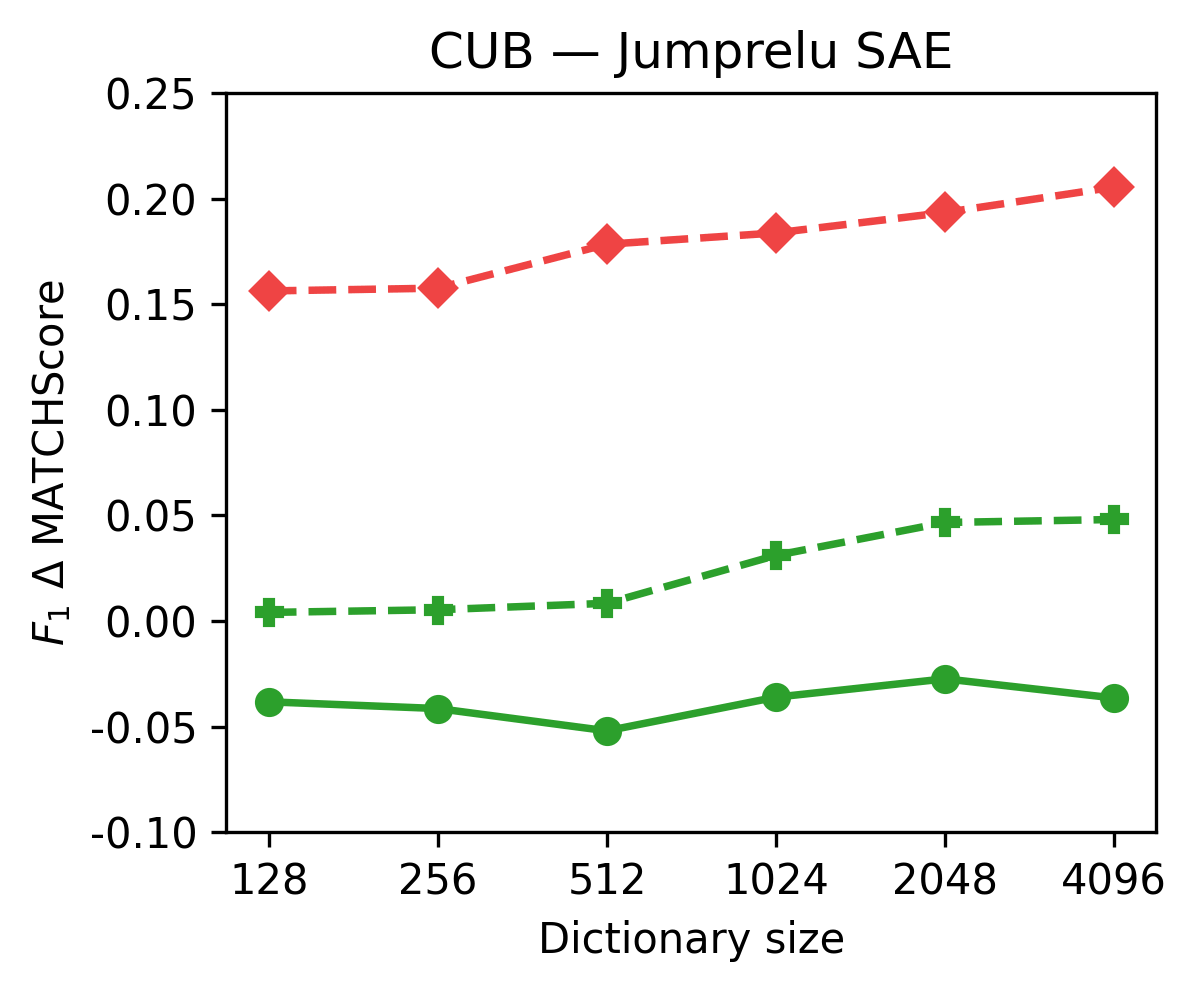}
    \end{subfigure}

    \vspace{2mm}
  
    \begin{subfigure}[t]{0.24\textwidth}
      \centering
      \includegraphics[width=\linewidth]{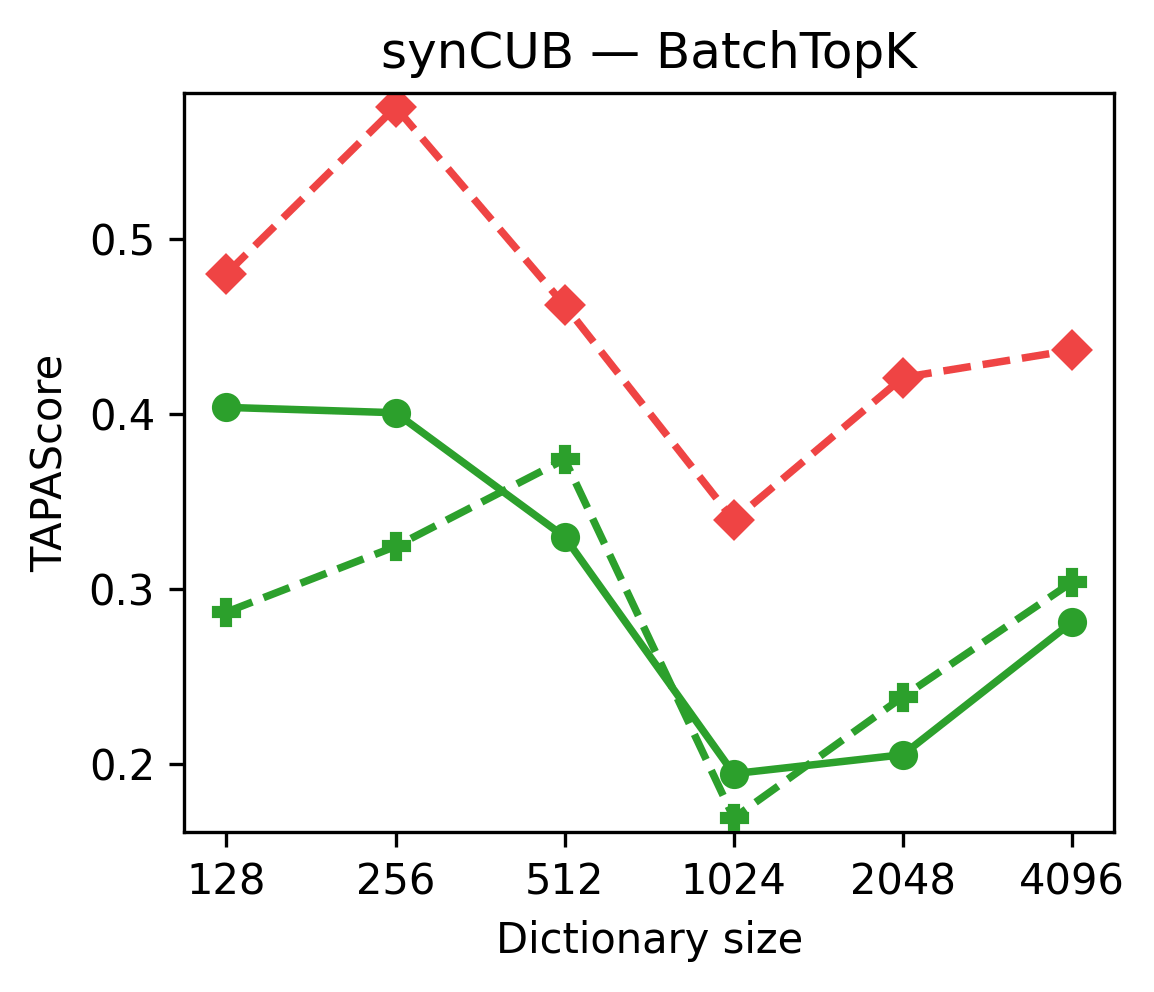}
    \end{subfigure}
    \hfill
    \begin{subfigure}[t]{0.24\textwidth}
      \centering
      \includegraphics[width=\linewidth]{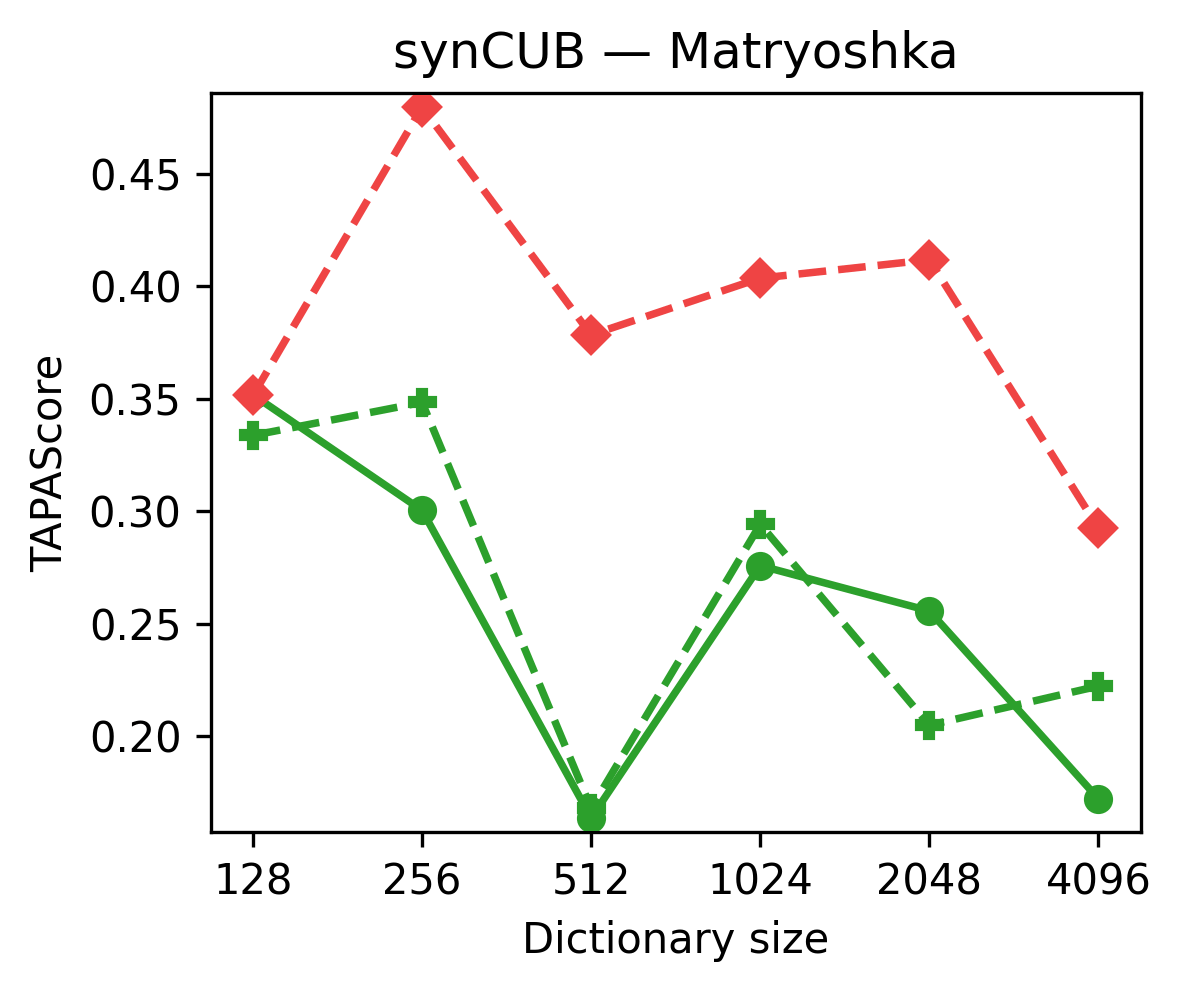}
    \end{subfigure}
    \hfill
    \begin{subfigure}[t]{0.24\textwidth}
      \centering
      \includegraphics[width=\linewidth]{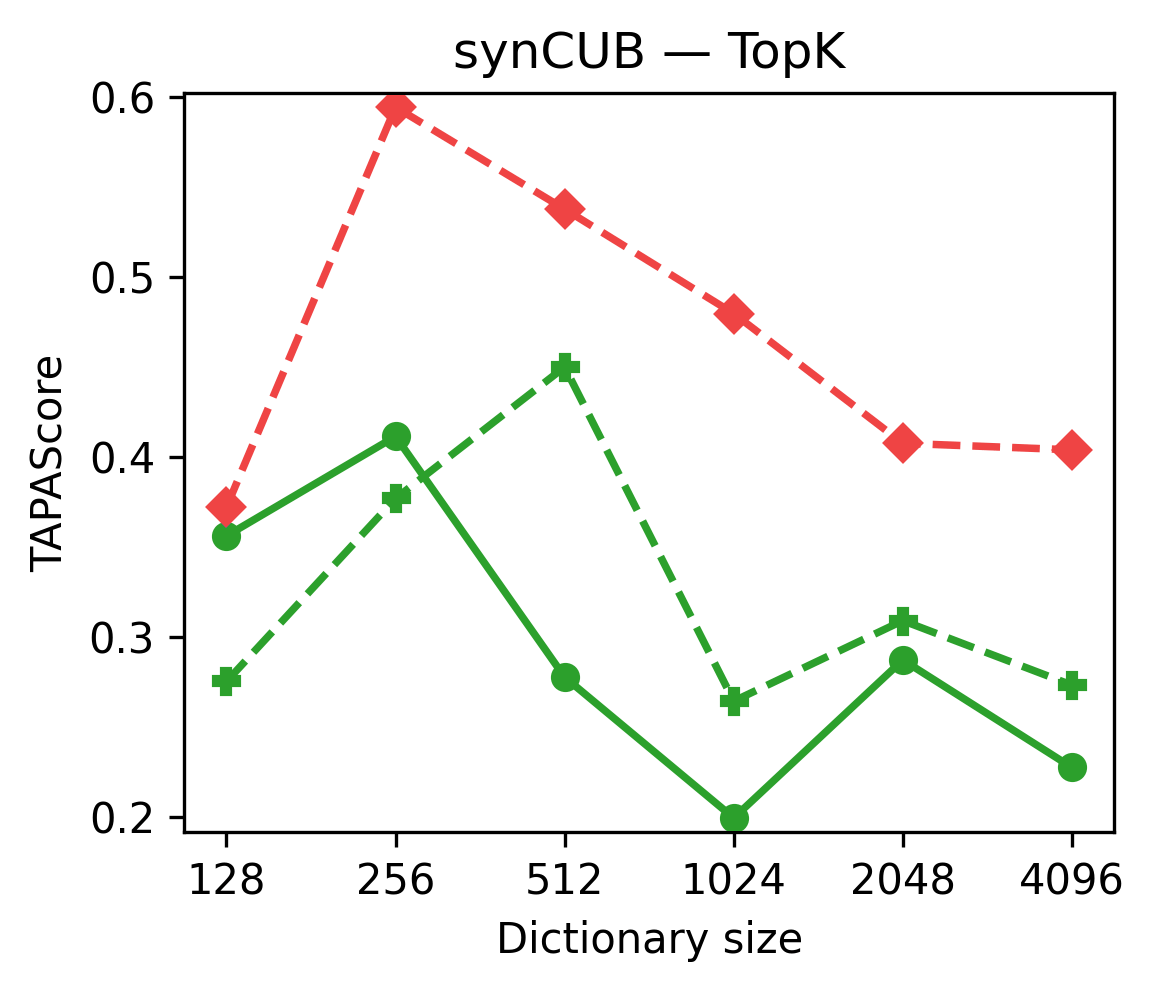}
    \end{subfigure}
    \hfill
    \begin{subfigure}[t]{0.24\textwidth}
      \centering
      \includegraphics[width=\linewidth]{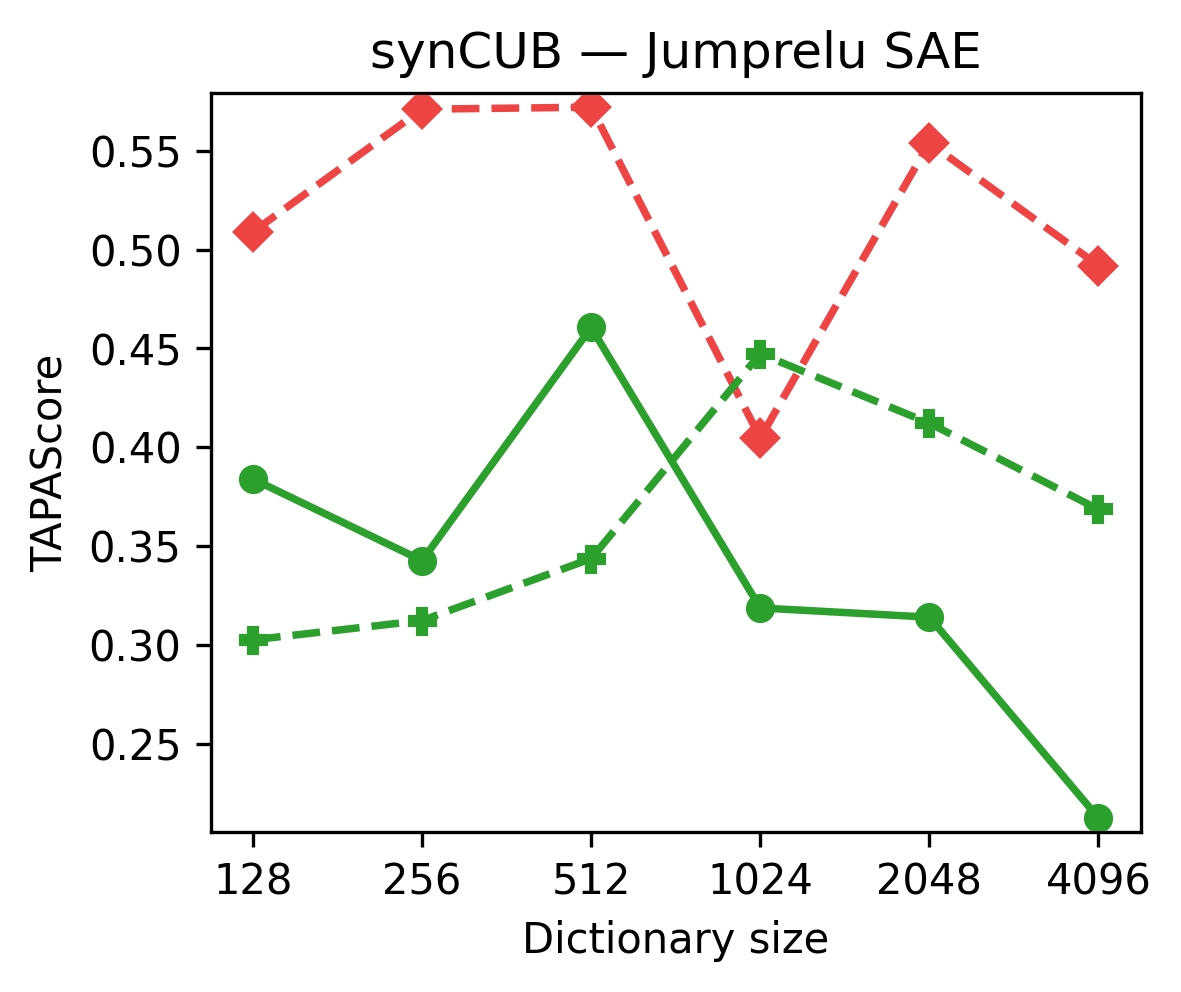}
    \end{subfigure}

    \vspace{1mm}

    \includegraphics[width=\linewidth]{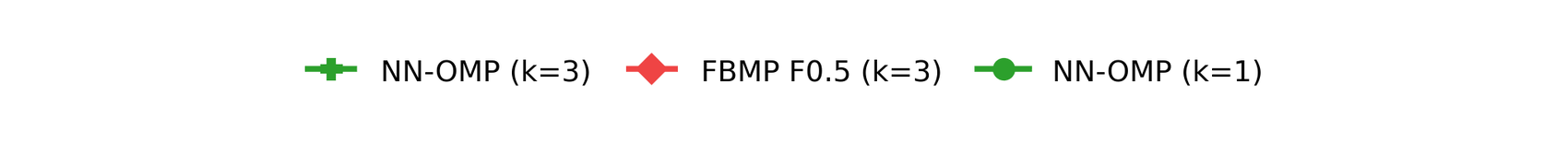}

  \vspace{-1.2em}
  \caption{$F_1$ $\Delta$MATCHScore on CUB (\textit{top row}) and TAPAScore on synCUB (\textit{bottom row}) comparing NN-OMP and FBMP F0.5 matching criteria for SAEs trained on CLIP embeddings, across BatchTopK, Matryoshka, TopK and JumpReLU SAE variants (\textit{left to right}).}
  \label{fig:appendix_nnomp_cub}
  \vspace{-2em}
  \end{figure*}


Binarizing the latent activations prior to matching discards their magnitudes,
which raises the question of whether a magnitude-aware matcher could exploit this
information to obtain better concept assignments. To investigate this, we replace
the binary $F_\beta$ selection criterion of FBMP with non-negative Orthogonal
Matching Pursuit (NN-OMP) \cite{bruckstein2008sparse}, a standard magnitude-aware
greedy algorithm that assigns each attribute to a coalition of latents using
inner-product correlations on the raw (non-binarized) activations. We evaluate
NN-OMP at coalition sizes $k{=}1$ (one-to-one) and $k{=}3$ (matched to the default
coalition size of FBMP), and compare it against FBMP F0.5 ($k{=}3$) across all
dictionary sizes ($128$--$4096$), both datasets, and the BatchTopK, Matryoshka,
TopK and JumpReLU variants. Figures~\ref{fig:appendix_nnomp_coco}
and~\ref{fig:appendix_nnomp_cub} report the $\Delta$MATCHScore (top rows) and the
TAPAScore (bottom rows) for COCO/synCOCO and CUB/synCUB, respectively.
On COCO, NN-OMP separates trained from untrained SAEs, as its $\Delta$ MATCHScore
(which subtracts the untrained baseline) is clearly positive across all
dictionary sizes; it is therefore a legitimate matcher rather than a degenerate
baseline, and at small to moderate dictionary sizes NN-OMP ($k{=}3$) is even
competitive with FBMP, which overtakes it at the largest dictionaries. On CUB,
however, the NN-OMP scores remain close to zero --- the one-to-one variant even
falls below the untrained baseline --- so that FBMP attains a markedly higher
$\Delta$MATCHScore at every dictionary size. The advantage is unambiguous under
the causal metric: on both synCOCO and synCUB, and across all four SAE variants,
FBMP F0.5 lies above both NN-OMP curves at essentially every dictionary size,
with the most consistent margins observed on synCOCO.

We attribute this behaviour to the binary structure of the matching task. The
ground-truth annotations are binary, and SAE sparsity already acts as a hard
concept selector, so that matching binary atoms under an $F_\beta$ criterion is
more discriminative than inner-product correlation over raw magnitudes. This
design choice does, however, place a particular concept of interpretability at
the center of the evaluation. A latent whose activation magnitude encodes several
distinct concepts is polysemantic in the sense studied by \cite{kopf2026capturing},
and a magnitude-aware matcher may reward such a latent for aligning with a target
attribute. We instead treat magnitude-encoded multiplicity as a form of
insufficient disentanglement: rather than crediting such latents, our binary
evaluation penalises them, which we regard as the intended behaviour for a metric
that aims to assess whether individual latents correspond to single, human-aligned
concepts.

\subsection{Effect of Activation Sparsity}
\label{app:sparsity_sweep}
\begin{figure}[p]
    \centering
    \begin{subfigure}[t]{0.24\textwidth}
      \includegraphics[width=\linewidth]{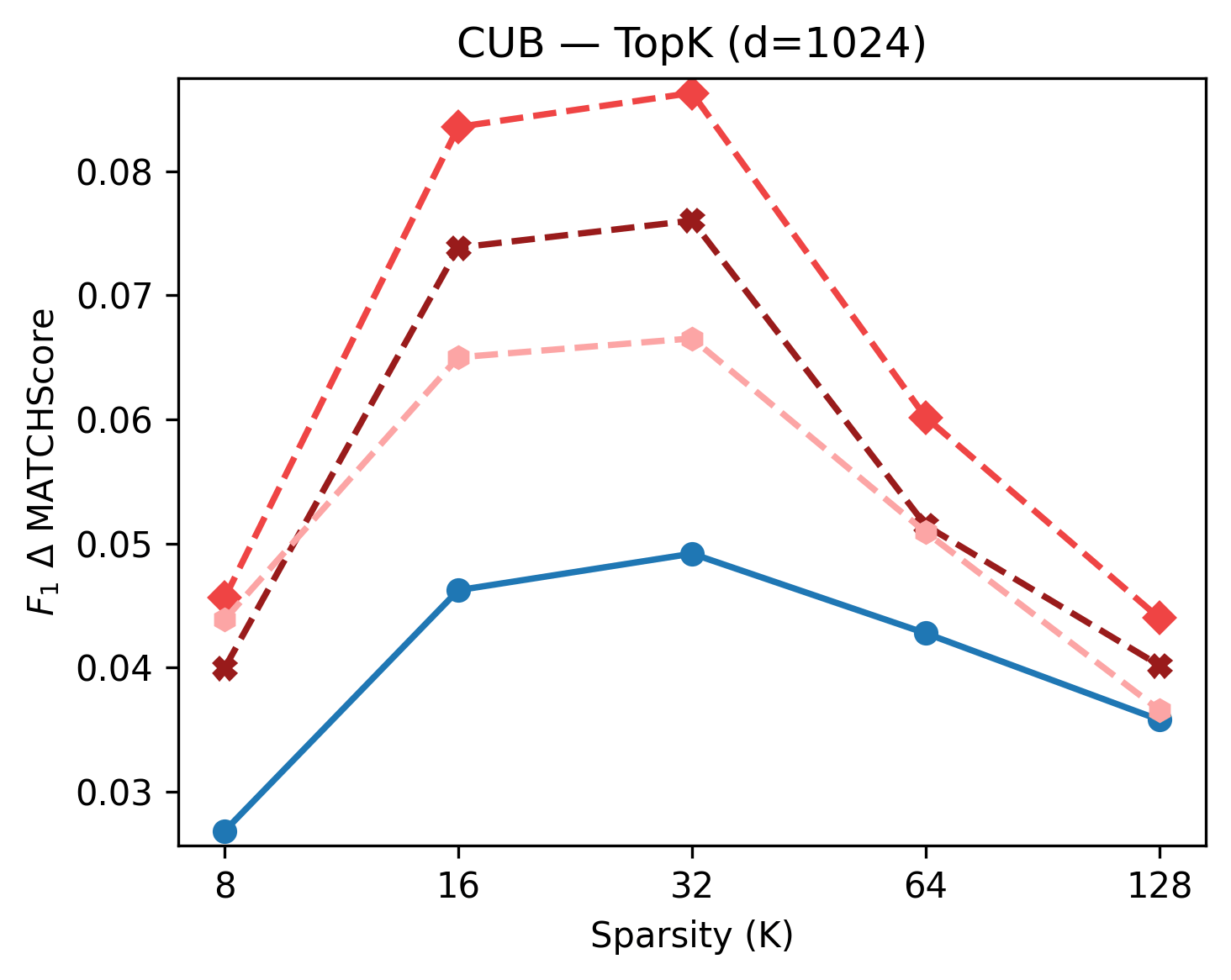}
    \end{subfigure}\hfill
    \begin{subfigure}[t]{0.24\textwidth}
      \includegraphics[width=\linewidth]{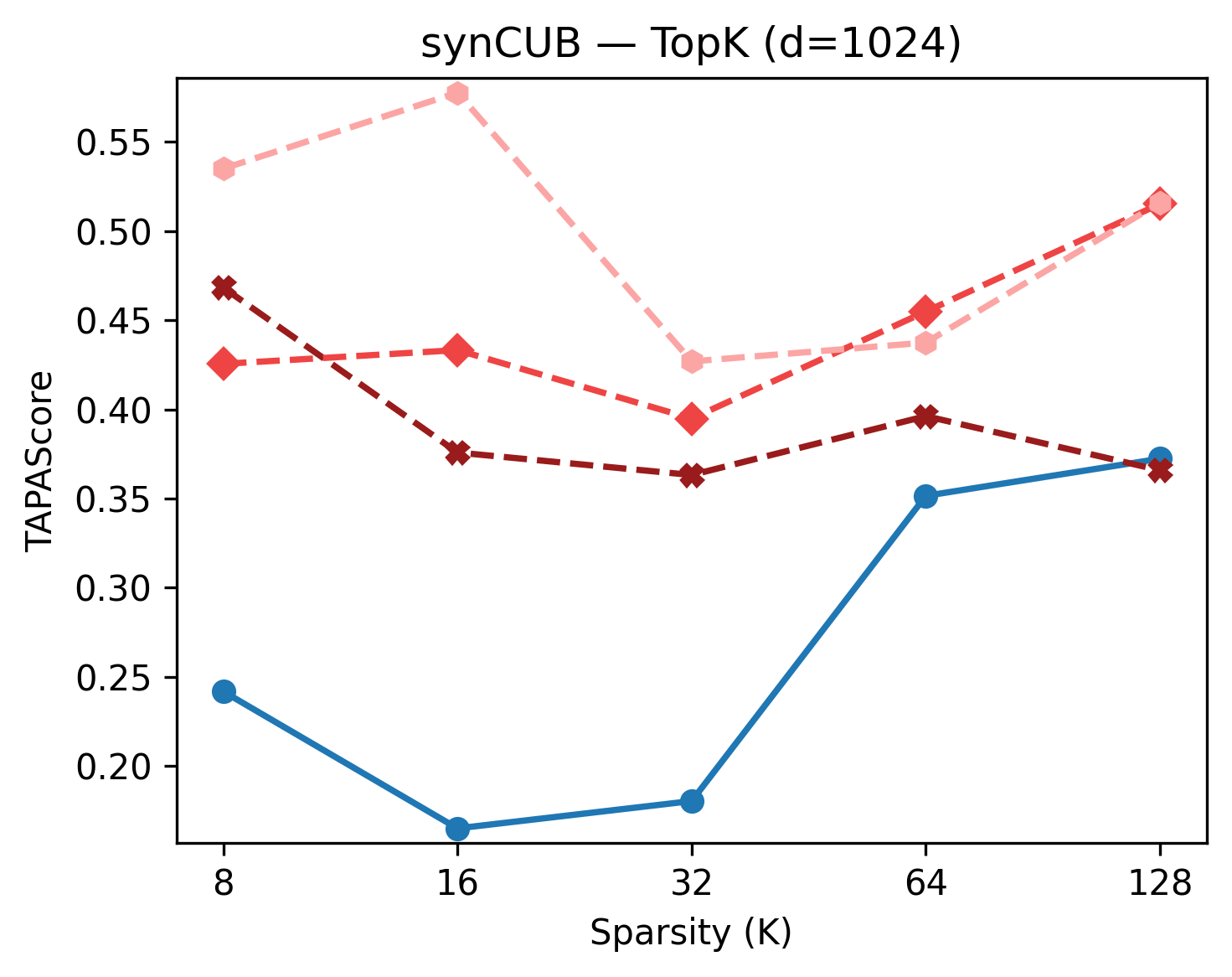}
    \end{subfigure}\hfill
    \begin{subfigure}[t]{0.24\textwidth}
      \includegraphics[width=\linewidth]{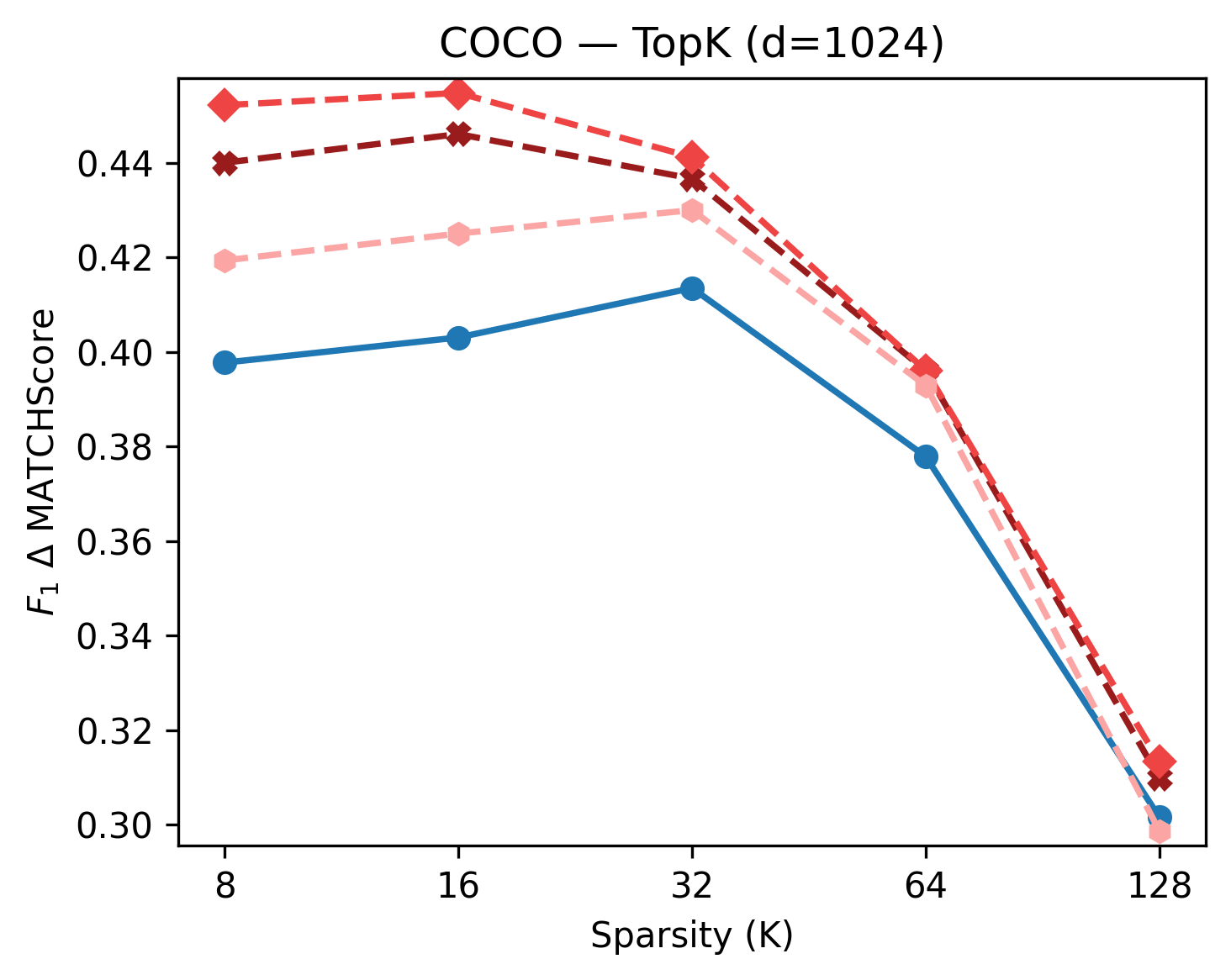}
    \end{subfigure}\hfill
    \begin{subfigure}[t]{0.24\textwidth}
      \includegraphics[width=\linewidth]{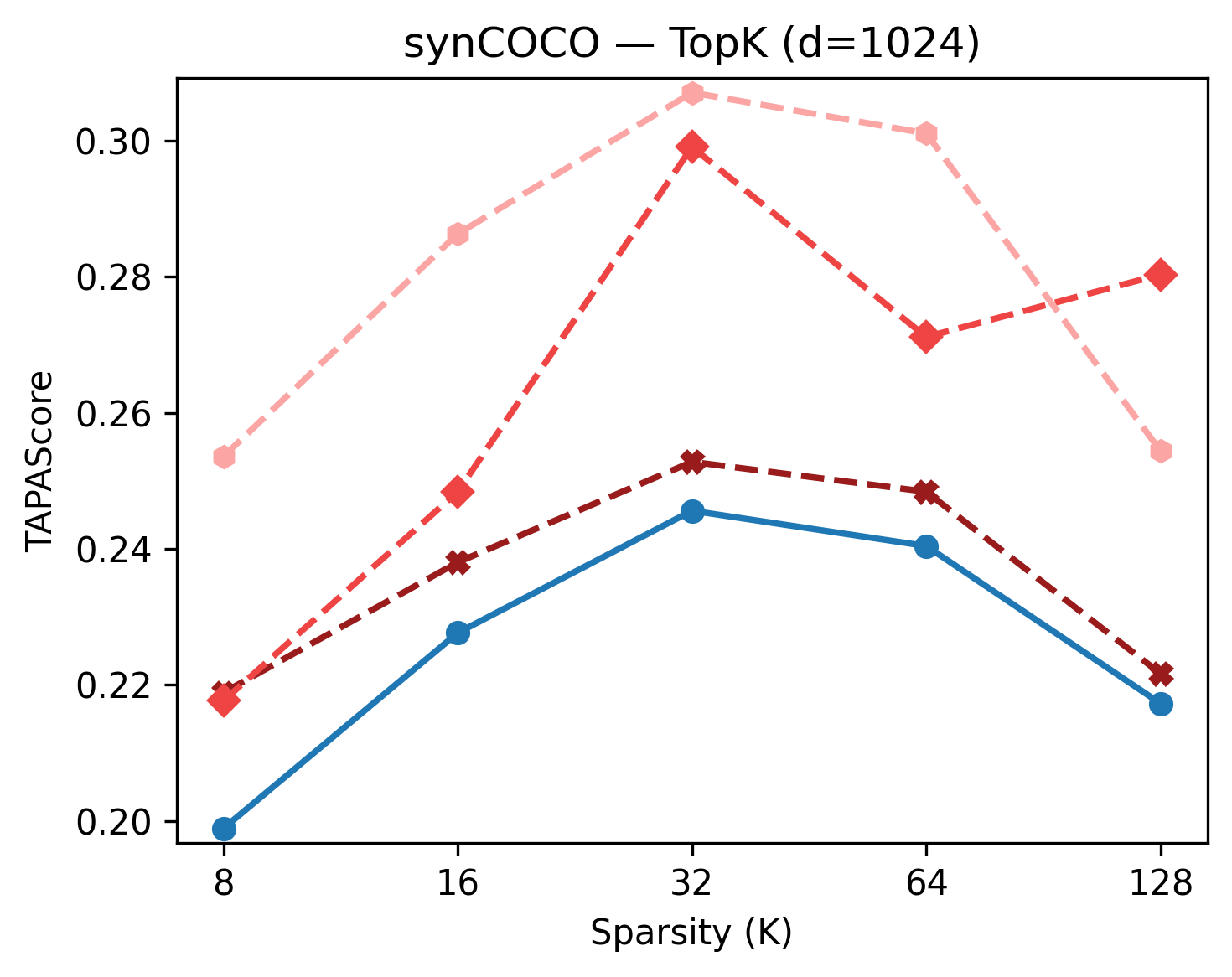}
    \end{subfigure}
    \\[4pt]
    \includegraphics[width=0.7\linewidth]{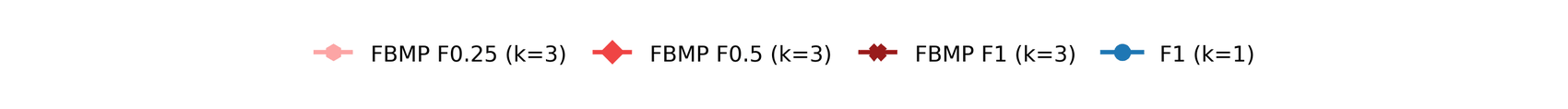}
    \caption{Effect of activation sparsity $K$ for TopK SAEs (CLIP) at dictionary size $d=1024$.
             From left to right: $\Delta$MATCHScore and TAPAScore on CUB/synCUB,
             followed by $\Delta$MATCHScore and TAPAScore on COCO/synCOCO.}
    \label{fig:appendix_sparsity_sweep}
  \end{figure}
Beyond dictionary size, we study the effect of the activation sparsity by training TopK SAEs with $K \in \{8, 16, 32, 64, 128\}$ at a fixed dictionary size of 1024 on CLIP embeddings of both datasets (Fig.~\ref{fig:appendix_sparsity_sweep}).
On CUB, the $\Delta$MATCHScore peaks at $K \in \{16, 32\}$ for all matching criteria and declines sharply for larger $K$. On COCO, matching scores remain stable for $K \leq 32$ before degrading in the same manner. The TAPAScore on synCOCO peaks at $K{=}32$ for all criteria. On synCUB, in contrast, TAPAScore is highest at small sparsity levels, peaking at $K{=}16$ for FBMP $F_{0.25}$, with the criteria diverging and partially recovering at larger $K$, while the one-to-one criterion behaves erratically.
Overall, moderate sparsity levels ($K \in \{16, 32\}$) provide the best trade-off between statistical and causal alignment. This extends our dictionary-size findings to the sparsity dimension, excessive capacity, whether through overcomplete dictionaries or large activation budgets, degrades concept alignment. This supports the default choice of $K{=}32$ used throughout the paper.

\subsection{Stability of Untouched Attributes ($\Delta$stay).}
\label{app:delta_stay}
TAPAScore measures whether matched latents respond to the \emph{perturbed} attribute, but a high score could in principle be inflated by leakage, i.e., latents drifting on attributes that were never touched by the intervention. To rule this out, we compute a complementary score, $\Delta$stay, on the same synthetic pairs. For each pair $(\mathbf{x}, \hat{\mathbf{x}})$ and each annotated attribute $a$ that is \emph{not} part of the intervention set, let $I_a$ denote its matched latent set. We define
\begin{equation}
\delta_{\text{stay}}^{a} = \max_{i \in I_a} \hat{\mathbf{z}}^i_\text{bin} - \max_{i \in I_a} \mathbf{z}^i_\text{bin},
\end{equation}
and report $\Delta$stay as the mean of $|\delta_{\text{stay}}^{a}|$ over all untouched-attribute instances. Small values indicate that latents matched to unrelated attributes remain stable under the perturbation, whereas large values indicate leakage. We keep $\Delta$stay separate from TAPAScore, rather than folding it into the score, so that the TAPAScore values reported throughout the paper remain directly comparable.

Table~\ref{tab:delta_stay} reports $\Delta$stay per SAE variant for FBMP $F_{0.5}$ matching with $k{=}3$, aggregated over dictionary sizes. We obtain $0.12 \pm 0.01$ on synCOCO and $0.21 \pm 0.04$ on synCUB, confirming low absolute drift on untouched attributes. The values are highly consistent across SAE variants, particularly on synCOCO, indicating that the high TAPAScores reported in the main paper cannot be explained by unselective latent drift.

\begin{table}[ht]
  \centering
  \caption{$\Delta$stay (mean $\pm$ std over dictionary sizes) for FBMP $F_{0.5}$ matching with $k{=}3$. Lower is better.}
  \label{tab:delta_stay}
  \begin{tabular}{l cc}
    \toprule
    SAE Variant & synCUB & synCOCO \\
    \midrule
    TopK & $0.18 \pm 0.05$ & $0.12 \pm 0.01$ \\
    BatchTopK & $0.24 \pm 0.03$ & $0.12 \pm 0.01$ \\
    Matryoshka & $0.24 \pm 0.02$ & $0.12 \pm 0.01$ \\
    JumpReLU & $0.20 \pm 0.02$ & $0.11 \pm 0.01$ \\
    \midrule
    All & $0.21 \pm 0.04$ & $0.12 \pm 0.01$ \\
    \bottomrule
  \end{tabular}
\end{table}

\end{document}